\documentclass[10pt,journal,compsoc]{IEEEtran}
\ifCLASSOPTIONcompsoc
\usepackage[nocompress]{cite}
\else
\usepackage{cite}
\fi

\usepackage{graphicx}
\usepackage{amsmath}
\usepackage{amssymb}
\usepackage{color}
\usepackage{epstopdf}
\usepackage{array}
\usepackage{multirow}
\usepackage{amsfonts}
\usepackage{booktabs}
\usepackage{float}
\usepackage{subfigure}
\usepackage{algorithm}
\usepackage{algpseudocode}
\usepackage[switch]{lineno}

\usepackage{url}  

\begin{document}
\bstctlcite{IEEEexample:BSTcontrol}
%
\title{Cross-Domain Facial Expression Recognition: \\ A Unified Evaluation Benchmark and \\ Adversarial Graph Learning}

\author{Tianshui Chen, Tao Pu, Hefeng Wu, Yuan Xie, Lingbo Liu, Liang Lin
\IEEEcompsocitemizethanks{\IEEEcompsocthanksitem Tianshui Chen is with The Guangdong University of Technology, Guangzhou, China. (E-mail: tianshuichen@gmail.com) \IEEEcompsocthanksitem Tao Pu, Hefeng Wu, Yuan Xie, and Liang Lin are with Sun Yat-Sen University, Guangzhou, China. (Email: putao3@mail2.sysu.edu.cn, wuhefeng@mail.sysu.edu.cn, phoenixsysu@gmail.com, linliang@ieee.org) \IEEEcompsocthanksitem Lingbo Liu is with The Hong Kong Polytechnic University. (Email: lingbo.liu@polyu.edu.hk) \IEEEcompsocthanksitem Tianshui Chen and Tao Pu contribute equally to this paper and share first authorship. Corresponding author: Hefeng Wu \IEEEcompsocthanksitem This work was supported in part by National Natural Science Foundation of China (NSFC) under Grant No. 61876045, 61836012, and 62002069, in part by the Natural Science Foundation of Guangdong Province under Grant No. 2017A030312006, and in part by Guangdong Provincial Basic Research Program under Grant No. 102020369.  \protect\\
}
}


\markboth{IEEE Transactions on Pattern Analysis and Machine Intelligence}%
{Chen \MakeLowercase{\textit{et al.}}: Cross-Domain Facial Expression Recognition: A Unified Evaluation Benchmark and Adversarial Graph Learning}

\IEEEtitleabstractindextext{
\begin{abstract}
Facial expression recognition (FER) has received significant attention in the past decade with witnessed progress, but data inconsistencies among different FER datasets greatly hinder the generalization ability of the models learned on one dataset to another. Recently, a series of cross-domain FER algorithms (CD-FERs) have been extensively developed to address this issue.

Although each declares to achieve superior performance, comprehensive and fair comparisons are lacking due to inconsistent choices of the source/target datasets and feature extractors. In this work, we first propose to construct a unified CD-FER evaluation benchmark, in which we re-implement the well-performing CD-FER and recently published general domain adaptation algorithms and ensure that all these algorithms adopt the same source/target datasets and feature extractors for fair CD-FER evaluations. Based on the analysis, we find that most of the current state-of-the-art algorithms use adversarial learning mechanisms that aim to learn holistic domain-invariant features to mitigate domain shifts. However, these algorithms ignore local features, which are more transferable across different datasets and carry more detailed content for fine-grained adaptation. Therefore, we develop a novel adversarial graph representation adaptation (AGRA) framework that integrates graph representation propagation with adversarial learning to realize effective cross-domain holistic-local feature co-adaptation. Specifically, our framework first builds two graphs to correlate holistic and local regions within each domain and across different domains, respectively. Then, it extracts holistic-local features from the input image and uses learnable per-class statistical distributions to initialize the corresponding graph nodes. Finally, two stacked graph convolution networks (GCNs) are adopted to propagate holistic-local features within each domain to explore their interaction and across different domains for holistic-local feature co-adaptation. In this way, the AGRA framework can adaptively learn fine-grained domain-invariant features and thus facilitate cross-domain expression recognition. We conduct extensive and fair comparisons on the unified evaluation benchmark and show that the proposed AGRA framework outperforms previous state-of-the-art methods.
\end{abstract}

}


\maketitle

\IEEEdisplaynontitleabstractindextext

%
\IEEEpeerreviewmaketitle

\IEEEraisesectionheading{\section{Introduction}
\label{sec:introduction}}
\IEEEPARstart{A}{utomatically} recognizing facial expressions helps understand human emotion states and behaviors, benefiting a wide range of applications such as human-computer interactions \cite{fragopanagos2005emotion}, medicine \cite{edwards2002emotion}, and security monitoring \cite{clavel2008fear,saste2017emotion}. Over the last decade, much effort has been dedicated to collecting various facial expression recognition (FER) datasets, namely, lab-controlled datasets (e.g., the Extended Cohn-Kanade (CK+) \cite{lucey2010extended}, Japanese Female Facial Expressions (JAFFE) \cite{lyons1998coding}, MMI	 \cite{valstar2010induced}, and Oulu-CASIA \cite{zhao2011facial} datasets) and in-the-wild datasets (e.g., the Real-world Affective Faces (RAF) \cite{li2017reliable,li2018reliable}, Static Facial Expressions in the Wild (SFEW2.0) \cite{dhall2011static}, Expression in-the-Wild (ExpW)  \cite{zhang2015learning}, SEWA DB \cite{kossaifi2019sewa}, and Facial Expression Recognition 2013 (FER2013) \cite{goodfellow2015challenges} datasets), which greatly facilitates FER performance. However, because humans’ understanding of facial expressions varies with their experiences and cultures, their annotations are inevitably subjective, leading to obvious domain shifts across different datasets \cite{Zeng_2018_ECCV,li2020deeper} (see Figure \ref{fig:dataset}). In addition, facial images of different datasets are usually collected in different environments (lab-controlled or in-the-wild environments) and from humans of different races, further enlarging the domain shift \cite{li2018deep}. Consequently, current best-performing methods may achieve satisfactory performance in intra-dataset protocols, but they suffer from dramatic performance deterioration in inter-dataset settings \cite{Zeng_2018_ECCV}.

\begin{table*}[!t]
\centering
\begin{tabular}{c|c|c|cccccc}
\hline
\centering  Method & Source set & Backbone & CK+ & JAFFE & SFEW2.0 &  FER2013 &  ExpW & Mean\\
\hline
\hline
Da et al. \cite{da2015effects} & BOSPHORUS & HOG \& Gabor filters & 57.60 & 36.2 & - & - & - & -\\
E3DCNN \cite{hasani2017facial} & MMI\&FERA\&DISFA & Inception-ResNet & 67.52 & - & - & - & -& -\\
STCNN \cite{hasani2017spatio} & MMI\&FERA & Inception-ResNet & 73.91 & - & - & - & -& -\\
GDFER \cite{mollahosseini2016going} &  Six datasets& Inception &  64.20 & - &39.80 & 34.00 & -& -\\
AIDN \cite{liu2015inspired} & CK+ & Manually designed network & - & - & 29.43 & - & - & -\\
DFA \cite{zhu2016discriminative} & CK+ & Manually designed network & - & 63.38 & - & - & - & -\\
DETN \cite{li2018deep} & RAF-DB & Manually designed network & 78.83 & 57.75 & 47.55 & 52.37  & -& -\\
ICID \cite{ji2019cross} & RAF-DB & DarkNet-19 & 84.50& - & - & - & -& -\\
ICID \cite{ji2019cross} & MMI & DarkNet-19 & 76.10& - & - & - & -& -\\
FTDNN \cite{zavarez2017cross} & Six datasets & VGGNet & 88.58 & 44.32 & - & - & -& -\\
ECAN \cite{li2020deeper} & RAF-DB 2.0 & VGGNet & 86.49 & 61.94 & 54.34 & 58.21 & -& -\\
\hline
\hline
DT & RAF-DB & ResNet-50 & 71.32 & 50.23 & 50.46 & 54.49 & 67.45 & 58.79 \\
PLFT & RAF-DB & ResNet-50 & 77.52 & 53.99 & 48.62 & 56.46 & 69.81 & 61.28 \\
DFA \cite{zhu2016discriminative} & RAF-DB &ResNet-50& 64.26 & 44.44 & 43.07 & 45.79 & 56.86 & 50.88 \\
DETN \cite{li2018deep} &RAF-DB &ResNet-50& 78.22 & 55.89 & 49.40 & 52.29 & 47.58 & 56.68 \\
ICID \cite{ji2019cross}  & RAF-DB &ResNet-50&  74.42 &  50.70 & 48.85 & 53.70 & 69.54 &  59.44\\ 
FTDNN \cite{zavarez2017cross} &RAF-DB &ResNet-50& 79.07 & 52.11 & 47.48 & 55.98 & 67.72 & 60.47 \\
ECAN \cite{li2020deeper} &RAF-DB &ResNet-50& 79.77 & 57.28 & 52.29 & 56.46 & 47.37 & 58.63 \\
LPL \cite{li2017reliable} & RAF-DB &ResNet-50& 74.42 & 53.05 & 48.85 & 55.89 & 66.90 & 59.82 \\
CADA \cite{long2018conditional} & RAF-DB &ResNet-50 & 72.09 & 52.11 & 53.44 & 57.61 & 63.15 & 59.68\\
SAFN \cite{xu2019larger} & RAF-DB &ResNet-50& 75.97 & 61.03 & 52.98 & 55.64 & 64.91 & 62.11 \\
SWD \cite{lee2019sliced} & RAF-DB &ResNet-50&75.19 & 54.93 & 52.06 & 55.84 &  68.35 & 61.27 \\
JUMBOT \cite{fatras2021unbalanced} & RAF-DB & ResNet-50 & 79.46 & 54.13 & 51.97 & 53.56 & 63.69 & 60.56 \\
ETD \cite{li2020enhanced} & RAF-DB & ResNet-50 & 75.16 & 51.19 & 52.77 & 50.41 & 67.82 & 59.47 \\
\hline
Ours & RAF-DB & ResNet-50 & \textbf{85.27} & \textbf{61.50} & \textbf{56.43} & \textbf{58.95}  & \textbf{68.50} & \textbf{66.13} \\
\hline
\end{tabular}
\vspace{2pt}
\caption{Evaluation settings and accuracies of current leading CD-FER methods on the CK+, JAFFE, SFEW2.0, FER2013, and ExpW datasets. The settings and results in the upper part are taken from the corresponding papers, and they generally use different source datasets and backbones for comparison. The results of the bottom part are generated by our implementation with ResNet-50 as the backbone and the RAF-DB dataset as the source dataset. Reference \cite{mollahosseini2016going} selects one dataset (i.e., CK+, SFEW2.0 or FER2013) from the seven datasets CK+, MultiPIE, MMI, DISFA, FERA, SFEW2.0, and FER2013 as the target domain and uses the rest six as the source domain; reference \cite{zavarez2017cross} selects one dataset (i.e., CK+ or JAFFE) from the seven datasets CK+, JAFFE, MMI, RaFD, KDEF, BU3DFE and ARFace as the target domain and uses the rest six as the source domain. ``-'' denotes the corresponding result is not provided.} 
\label{table:motivation}
\end{table*}

Recently, much effort has been dedicated to the cross-domain FER (CD-FER) task by learning transferable features. Although each newly proposed method claims to achieve superior  performance, it is difficult to assess the actual improvement of each method due to the inconsistent choice of the source/target datasets and feature extractors (see Table \ref{table:motivation}). On the one hand, the source datasets carry basic supervision for feature extractor and classifier learning, and the distribution similarity between the source and target datasets plays a key role in the final performance. Take the ICID algorithm \cite{ji2019cross} in Table \ref{table:motivation} for example, using different source datasets leads to a performance disparity of more than 8.0\% on the CK+ dataset. On the other hand, features extracted using different backbone networks inherently have different discrimination and generalization abilities. These inconsistent choices hinder fair comparisons among CD-FER algorithms.

\begin{figure}[!h]
   \centering
   \includegraphics[width=0.9\linewidth]{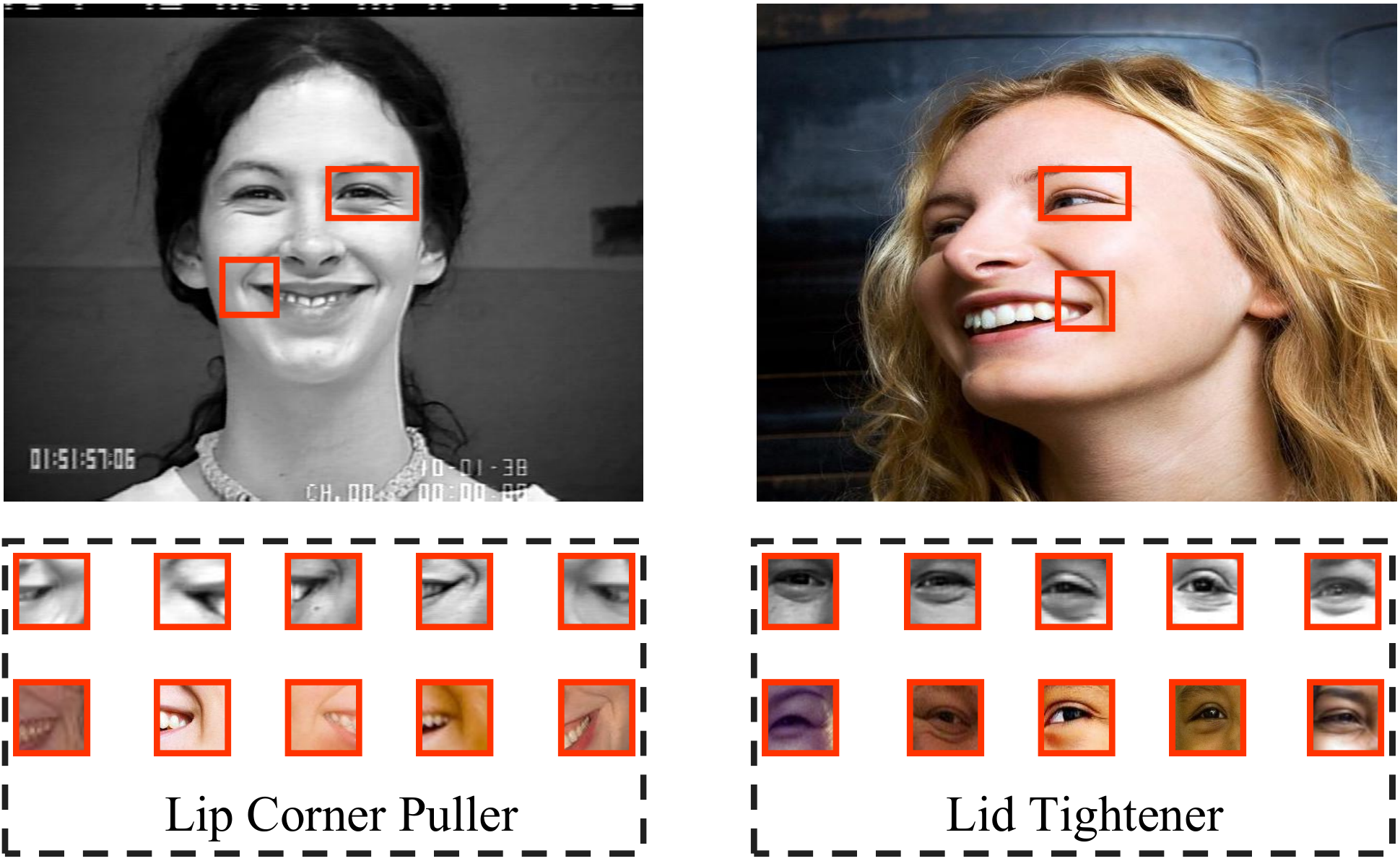}
   \caption{The upper part presents two examples from CK+ \cite{lucey2010extended} and RAF-DB \cite{li2018reliable}, and the lower part presents the local mouth-corner and eye regions of the images from CK+ and RAF-DB.}
   \label{fig:vis}
\end{figure}

Therefore, our first goal of this work is to construct a unified CD-FER evaluation benchmark for fair comparison and promote research in this field. To this end, we first analyze the performance gap caused by these inconsistent choices and re-implement several of the best-performing options \cite{ji2019cross,zhu2016discriminative,li2018deep,zavarez2017cross,li2020deeper,li2017reliable} according to the corresponding papers. In addition, many general domain adaptation methods exist, and we apply some of the best-performing ones \cite{long2018conditional,lee2019sliced,xu2019larger} to CD-FER. To ensure fair comparisons, we use the same backbone network for feature extraction and the same source/target datasets for all the algorithms. For example, the lower part of Table \ref{table:motivation} reports the results that are obtained by using the same RAF-DB \cite{li2018reliable} source dataset and ResNet-50 backbone, and it can better show the actual performance comparisons among different algorithms. On the other hand, current FER datasets mainly feature Western individuals. 
To facilitate more comprehensive evaluations, we further build a large-scale FER dataset (namely, the Asian Face Expression (AFE) dataset) that contains 54,901 well-labeled samples captured from Asian individuals. This dataset is used as the source dataset to evaluate the cross-culture FER performance.

Our second goal of this work is to propose a new adversarial learning framework that realizes effective cross-domain holistic-local feature co-adaptation for CD-FER, motivated by the following observations. To address the CD-FER task, most of the current state-of-the-art algorithms adopt adversarial learning mechanisms \cite{tzeng2017adversarial,long2018conditional, wang2018deepsurvey,goodfellow2014generative, wei2018unsupervised, wang2018unsupervised} to learn domain-invariant features, but they just focus on extracting holistic features for domain adaptation. We believe that incorporating local features can benefit CD-FER because they are more transferable across different datasets and carry more detailed content for fine-grained adaptation. As exhibited in Figure \ref{fig:vis}, the lip-corner-puller and lid tightener actions can be used to distinguish happy expressions, and they are similar for samples from different datasets. To give a more in-depth analysis, we conduct an experiment on the generalization abilities of using holistic, local, and holistic-local features (see Sec. \ref{sec:overview} for details). As shown in Figure \ref{fig:mmd}, using the local or holistic-local features can greatly reduce the distribution discrepancies between learned features of the source and target datasets compared with using the holistic features. Thus, it is worthy to model the correlation of holistic-local features for addressing CD-FER.

\begin{figure}[!t]
\centering
\subfigure{
\label{fig:subfig_vis_1} 
\includegraphics[width=0.48\linewidth]{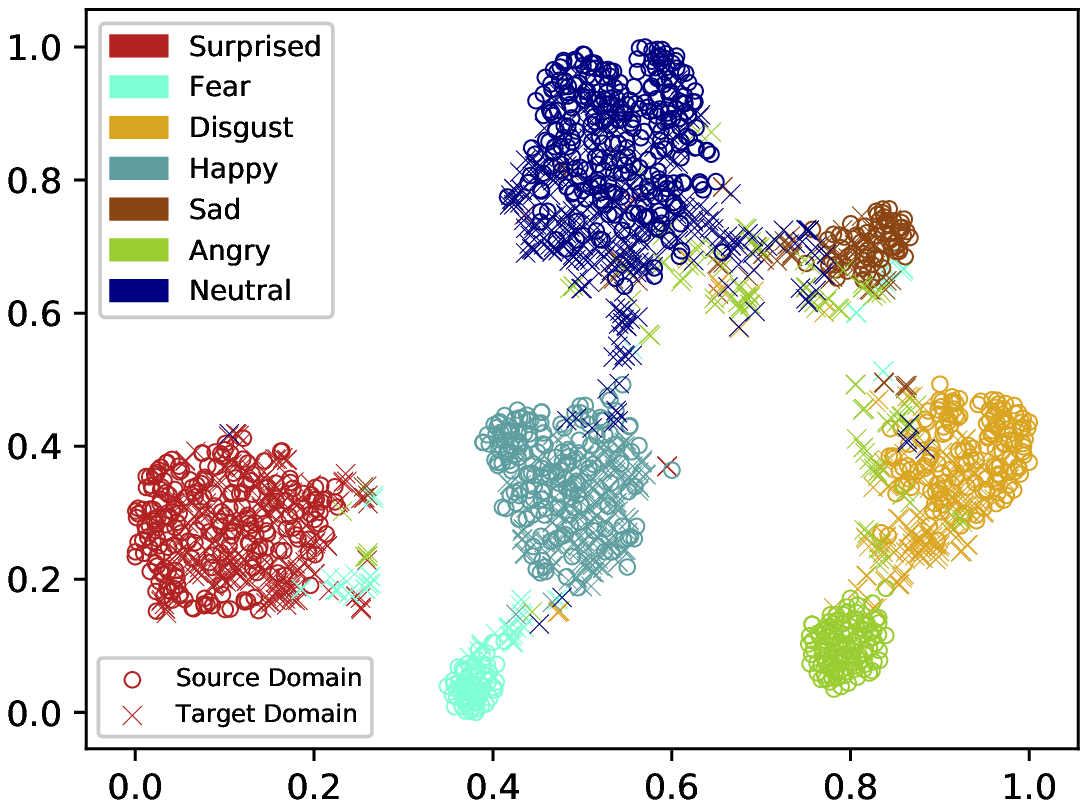}}
\subfigure{
\label{fig:subfig_vis_2} 
\includegraphics[width=0.48\linewidth]{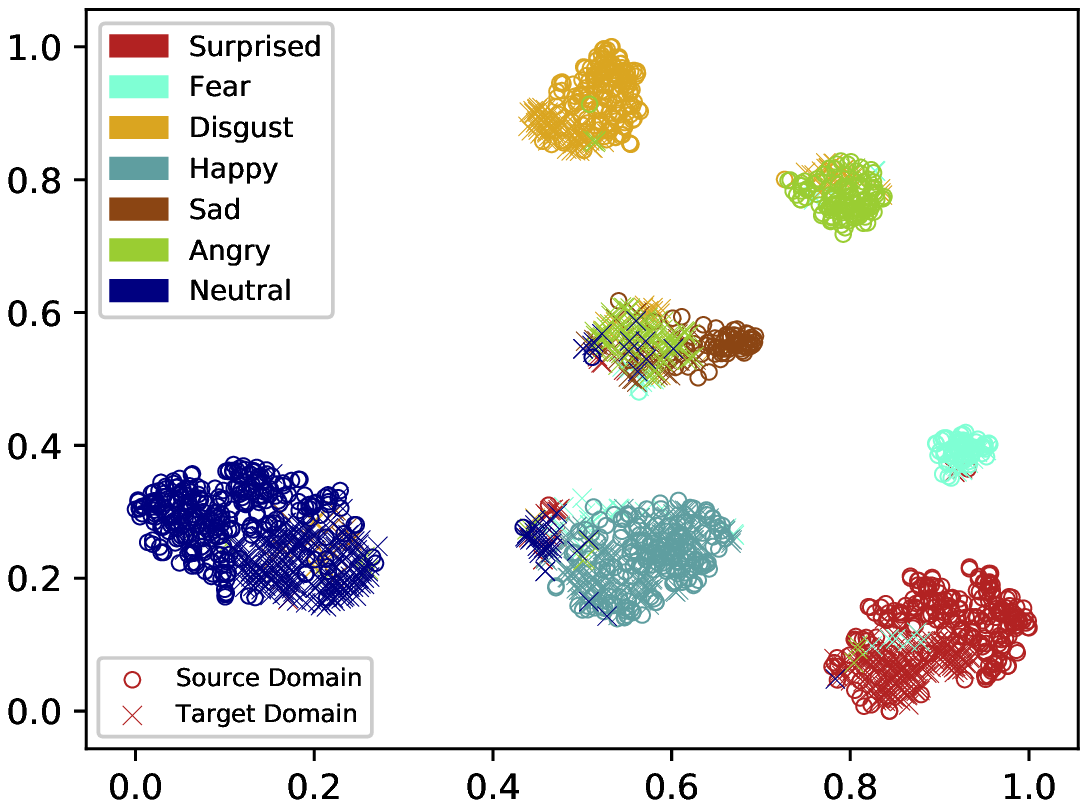}}
\caption{Illustration of a feature distribution learned by the baseline adversarial learning \cite{tzeng2017adversarial} method that merely uses holistic features (left) and our proposed AGRA framework (right). It is obvious that the AGRA framework can better gather the samples of the same category and from different domains together than the baseline method, suggesting that our framework can learn more discriminative domain-invariant features for CD-FER.}
\label{fig:motivation}
\end{figure}

In this work, we show that the correlation of holistic-local features within each domain and across the source and target domains can be explicitly represented by structured graphs, and their interplay and adaptation can be captured by adaptive message propagation through the graphs. To achieve the aforementioned goal, we develop a novel adversarial graph representation adaptation (AGRA) framework that integrates graph representation propagation with adversarial learning for cross-domain holistic-local feature interplay and co-adaptation. Specifically, we first extract several discriminative local regions based on facial landmarks (e.g., eyes, nose, and mouth corner) \cite{zhang2016joint,liu2019facial} and build two graphs to correlate holistic images and local regions within each domain and across different domains, respectively. Given an input image from one domain, we extract features of the holistic image and the local regions to initialize the corresponding nodes of this domain. The nodes of the other domain are initialized by the corresponding per-class learnable statistical feature distributions. Then, we introduce two stacked graph convolutional networks (GCNs) to propagate node messages within each domain to explore  holistic-local feature interactions and across the two different domains to enable holistic-local feature co-adaptation. In this way, our method can progressively mitigate the shift in  the holistic-local features between the source and target domains, enabling learning discriminative and domain-invariant features to facilitate CD-FER. Figure \ref{fig:motivation} shows the feature distributions learned by the baseline adversarial learning method\cite{tzeng2017adversarial} that merely uses holistic features for domain adaptation, and our proposed AGRA framework. The figure shows that our framework can better gather the features of samples that belong to the same category and are taken from different domains together than the baseline method. This phenomenon suggests that our framework can better learn domain-invariant features while improving their discriminative ability.

A preliminary version of this work was presented as a conference paper \cite{xie2020adversarial}. In this version, we strengthen the work from four aspects. First, we construct a unified and comprehensive CD-FER evaluation benchmark by re-implementing more best-performing algorithms and providing a variety of unified evaluation protocols (i.e., using the same source/target datasets and feature extractors) for comparing these algorithms fairly. We follow the evaluation benchmark to conduct extensive performance analysis and comparison under various evaluation protocols (i.e., different combination of source/target datasets and feature extractors). This is the first attempt to construct such a benchmark and it can doubtlessly facilitate the progress of CD-FER. Second, we build a new large-scale FER dataset that contains tens of thousands of samples captured mainly from Asian individuals. This dataset is incorporated into our benchmark and can be used to compare the cross-culture FER performance. Third, the motivation and details of the proposed AGRA framework are better described and enriched. Finally, substantially more experiments are conducted to demonstrate the effectiveness of the proposed framework and verify the contribution of each component.

The contributions of this work can be summarized as follows: 1) We construct a fair and comprehensive CD-FER evaluation benchmark by unifying the source/target datasets and feature extractors for different well-performing algorithms. To the best of our knowledge, this is the first attempt to construct such a unified evaluation benchmark. 2) We construct a large-scale FER dataset that contains 54,901 well-labeled samples mainly captured from Asian individuals. This dataset can be used to promote cross-culture FER task. 3) We propose to integrate graph representation propagation with the adversarial learning mechanism for holistic-local feature co-adaptation across different domains. This method can learn fine-grained and domain-invariant features to improve CD-FER performance. 4) We develop a class-aware two-stage updating mechanism to iteratively learn the statistical feature distribution of each domain for graph node initialization. This mechanism is a key factor in mitigating domain shifting to facilitate learning domain-invariant features. 5) We conduct comprehensive experiments using the unified evaluation benchmark to  compare the leading CD-FER algorithms fairly and verify the effectiveness of the proposed framework. When using RAF-DB as the source dataset and ResNet-50 as the backbone, the AGRA framework improves the accuracy averaging over the CK+ \cite{lucey2010extended}, JAFFE \cite{lyons1998coding}, FER2013 \cite{goodfellow2015challenges}, SFEW2.0 \cite{dhall2011static}, and ExpW \cite{zhang2018facial} datasets by 4.02\% compared with the previously identified best performers. The unified evaluation benchmark, including the implementation codes of our proposed AGRA framework and current leading methods, the trained models and the newly built AFE dataset, is available at \textbf{\url{https://github.com/HCPLab-SYSU/CD-FER-Benchmark}.}

The remainder of this work is organized as follows. We review the most related works in Sec. \ref{sec:related_works}. Sec. \ref{sec:fair_evaluation} presents the unified evaluation benchmark, and Sec. \ref{sec:AGRA} introduces the proposed AGRA framework in detail. We provide extensive experimental comparison and evaluation in Sec. \ref{sec:experiments} and conclude the work in Sec. \ref{sec:conclusion}.

\section{Related Works}
\label{sec:related_works}
In this section, we mainly review three streams of related works: cross-domain FER, adversarial domain adaptation, and graph representation learning.

\subsection{Cross-Domain Facial Expression Recognition}
Due to the subjective annotation process and inconsistent collection conditions, distribution divergences commonly exist among different FER datasets. To maintain the performance of cross-dataset validation, many CD-FER algorithms have been proposed \cite{miao2012cross,sangineto2014we,yan2016transfer,zhu2016discriminative,yan2016cross,zheng2016cross,chu2016selective,li2020deeper,zong2018domain,yan2019cross}. For instance, Yan et al. \cite{yan2016transfer} used subspace learning to transfer the knowledge extracted from the source dataset to the target dataset. However, this method still requires annotating some samples from the target dataset, which is unexpected in unsupervised CD-FER scenarios. In contrast, Zheng et al. \cite{zheng2016cross} proposed combining the labeled samples from the source domain and unlabeled auxiliary data from the target domain to jointly learn a discriminative subspace. This algorithm does not require any annotated samples from the target domain and thus facilitates CD-FER in an unsupervised manner. In contrast, Zong et al. \cite{zong2018domain} generated additional samples that share the same or similar feature distribution for both the source and target domains. Wang et al. \cite{wang2018unsupervised} further introduced generative adversarial networks \cite{goodfellow2014generative} to generate more subtle samples to facilitate CD-FER. More recently, Li et al. \cite{li2020deeper} observed that the conditional probability distributions between the source and target datasets are different. Based on this observation, they developed a deep emotion-conditional adaptation network (ECAN) that simultaneously considers conditional distribution bias and the expression class imbalance problem in CD-FER. 

Though each work claims to achieve superior performance to previous algorithms, the comparisons are somewhat unfair, as these works often select completely different feature extractors and are evaluated on different source/target datasets. Thus, it is difficult to assess the actual improvement of each algorithm. To promote a fair comparison, we build a unified and comprehensive CD-FER evaluation benchmark by unifying the choices of the source/target datasets and feature extractor for the competing algorithms. This benchmark is novel and crucial for the CD-FER community. In addition, we propose a novel adversarial graph representation adaptation framework, which integrates graph propagation networks with adversarial learning mechanisms for adaptive holistic-local feature co-adaptation to facilitate CD-FER.

\subsection{Adversarial Domain Adaptation}
Obvious domain discrepancies commonly exist among different datasets. Recently, variant domain adaptation methods \cite{tzeng2017adversarial,long2018conditional,xu2019larger} have been intensively proposed to learn domain-invariant features; thus, classifiers/predictors learned using source datasets can be generalized to target test datasets. Motivated by generative adversarial networks \cite{goodfellow2014generative} that aim to generate samples that are indistinguishable from real samples, recent domain adaptation methods  \cite{tzeng2017adversarial,long2018conditional} also resort to adversarial learning to mitigate domain shifts. Specifically, adversarial learning involves a two-player game in which a feature extractor aims to learn transferable domain-invariant features while a domain discriminator struggles to distinguish samples from the source domain from those from the target domain. As a pioneering work, Tzeng et al. \cite{tzeng2017adversarial} propose a generalized adversarial adaptation framework by combining discriminative modeling, untied weight sharing, and an adversarial loss. Long et al. further designed a conditional adversarial domain adaptation method that further introduces two strategies of multilinear conditioning and entropy conditioning to improve the discriminability and control the uncertainty of the classifier. Despite achieving impressive progress for cross-domain general image classification, these methods \cite{tzeng2017adversarial,long2018conditional} mainly focus on holistic features for adaptation and ignore local content, which carries more transferable and fine-grained features. Different from these works, we propose to represent the correlation of holistic and local features by structured graphs and integrate graph propagation networks with adversarial learning to learn domain-invariant holistic-local features.

\subsection{Graph Representation Learning}
Deep convolutional neural networks (CNNs) \cite{simonyan2015very,he2016deep} have achieved impressive performance in visual recognition tasks \cite{he2016deep,chen2016disc,wang2017multi,chen2018recurrent,chen2018learning}, but these networks deal with only gridded data and are difficult to adapt to graph-structured data. To solve this issue, recent efforts have been dedicated to devising a series of graph neural networks \cite{kipf2017semi,li2016gated} that can learn the representation of graph-structured data via iterative message propagation. Recently, these graph neural networks have also been adapted to model visual feature interactions to facilitate variant tasks \cite{DBLP:conf/cvpr/ChenWWG19,chen2018knowledge,chen2019learning,chen2020knowledge,chen2019knowledge,wang2018deep,jiang2018hybrid}, ranging from object classification and detection \cite{DBLP:conf/cvpr/ChenWWG19,chen2019learning,jiang2018hybrid,chen2020knowledge1} to visual relationship reasoning \cite{wang2018deep,chen2019knowledge} and visual navigation \cite{yang2019visual} to traffic forecasting \cite{liu2020physical,liu2021online}. For example, references \cite{chen2019learning,DBLP:conf/cvpr/ChenWWG19} modeled label dependencies with structured graphs and used graphs to guide feature and classifier learning to facilitate multilabel image recognition. Jiang et al. \cite{jiang2018hybrid} introduced semantic label/attribute relationships and spatial relationships to help learn contextualized features and applied them to boost the large-scale object detection performance. Chen et al. \cite{chen2019knowledge} further extended graph neural networks to capture the interactions between candidate objects and their relationships, which can improve the visual relationship detection performance and alleviate the performance degradation caused by the long-tail distribution issue. Wang et al. \cite{wang2018deep} introduced a graph to model the correlations between social relationships and related semantic objects and applied graph neural networks to perform message propagation through the graph to capture their interactions. Inspired by these works, we introduce graph neural networks to capture the interactions among holistic-local features within each domain and across different domains and integrate them with adversarial learning to facilitate learning fine-grained domain-invariant features.

\section{Unified Evaluation Benchmark}
\label{sec:fair_evaluation}
In this section, we present the datasets, feature extractors and competing algorithms involved in constructing the unified CD-FER evaluation benchmark, along with the analysis of performance gap caused by the inconsistent choices of the source/target datasets and feature extractors. Then, we present the unified evaluation protocols of the benchmark for fair comparison.

\subsection{Datasets for CD-FER}\label{sec:datasets}
There exist enormous FER datasets, and they also serve as the source/target datasets for CD-FER. Here, we mainly discuss several publicly available datasets that are widely used as source/target datasets in CD-FER works, including lab-controlled (e.g., the CK+ \cite{lucey2010extended} and JAFFE \cite{lyons1998coding}) and in-the-wild (e.g., the FER2013 \cite{goodfellow2015challenges}, SFEW2.0 \cite{dhall2011static}, ExpW \cite{zhang2018facial}, RAF-DB \cite{li2018reliable}) datasets. These datasets cover only the seven basic expressions. Additionally, we introduce the newly built AFE dataset here.

\begin{figure}[!t]
   \centering
   \includegraphics[width=0.99\linewidth]{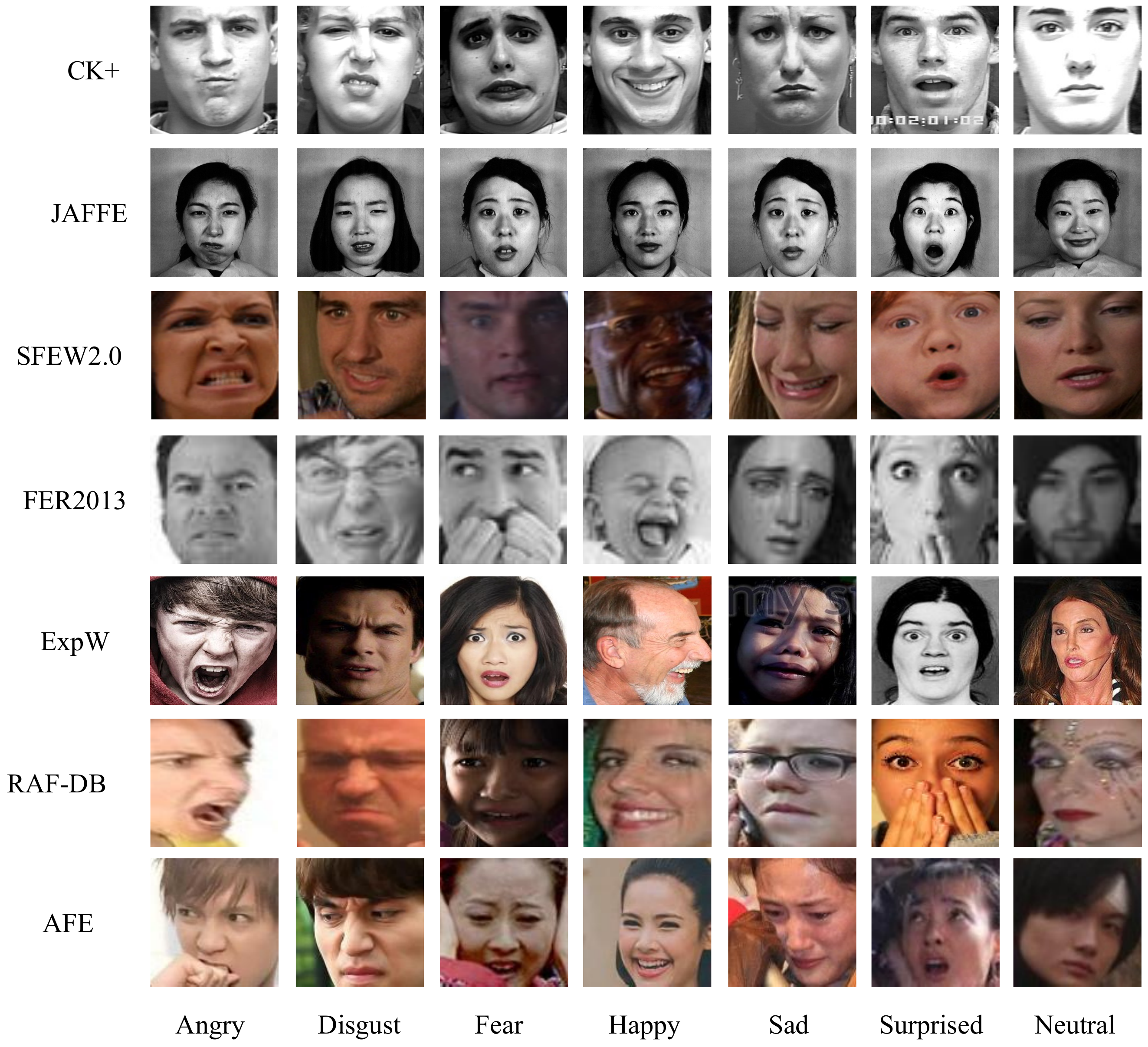}
\caption{Visualization of samples from the CK+, JAFFE, SFEW2.0, FER2013, ExpW, and RAF-DB datasets. The examples from different datasets differ in appearance, color, and view point.}
   \label{fig:dataset}
\end{figure}

\noindent\textbf{CK+ \cite{lucey2010extended} }is a lab-controlled dataset that is mostly used for FER. It contains 593 videos from 123 subjects, among which 309 sequences are labeled with 
six basic expressions 
based on the Facial Action Coding System (FACS). We follow previous work \cite{li2018deep} to select the three frames with peak formation from each sequence and the first frame (neutral expression) of each sequence, resulting in 1,236 images for evaluation. The dataset is divided into a training set of 1,125 and a test set of 129 images.

\noindent\textbf{JAFFE \cite{lyons1998coding}} is another lab-controlled dataset that contains 213 images from 10 Japanese females. Approximately 3-4 images of each person are annotated with one of the six basic expressions and 1 image annotated with a neutral expression. This dataset covers only Asian people and could be used for cross-culture evaluation. As this dataset merely contains 213 images, we follow previous works \cite{li2018reliable,li2018deep} to use the whole dataset for training and test sets.

\noindent\textbf{SFEW2.0 \cite{dhall2011static} }is an in-the-wild dataset collected from different films with spontaneous expressions, various head poses, age ranges, occlusions and illuminations. This dataset is divided into training, validation, and test sets, with 958, 436, and 372 samples, respectively.

\noindent\textbf{FER2013 \cite{goodfellow2015challenges} }is a large-scale uncontrolled dataset that was automatically collected by the Google Image Search application programming interface (API). It contains 35,887 images of size 48$\times$48 pixels, and each image is annotated with the seven basic expressions. The dataset is further divided into a training set of 28,709 images, a validation set of 3,589 images, and a test set of 3,589 images.

\noindent\textbf{ExpW \cite{zhang2018facial} }is an in-the-wild dataset with images that have been downloaded from Google Image searches. This dataset contains 91,793 face images, and each image is manually annotated with one of the seven expressions. In the experiments, we divide the dataset into a training set of 28,848 images, a validation set of 28,848 images, and a test set of 34,097 images.

\noindent\textbf{RAF-DB \cite{li2018reliable} }contains 29,672 highly diverse facial images from thousands of individuals that were also collected from the Internet. Among these images, 15,339 images are annotated with the seven basic expressions, which are divided into 12,271 training samples and 3,068 testing samples for evaluation.

\noindent\textbf{AFE } is a new dataset constructed in this work that covers thousands of Asian individuals. To collect this dataset, we first downloaded approximately 500,000 images of faces from the film \textit{DouBan}\footnote{https://movie.douban.com/}. Then, each image was annotated by 3-4 annotators, and only the image that all annotators set to the same expression are kept, leading to 54,901 well-labeled samples. The dataset is further divided into 32,757 images for training, 16,380 images for validation, and 5,464 images for testing. As most images from existing datasets are of Americans and Europeans, this dataset can be used for cross-culture domain adaptation. As described above, JAFFE \cite{lyons1998coding} also covers Asian people, but it contains merely 213 images. In contrast, AFE contains much more images (i.e., more than 50,000), and it can better facilitate cross-culture FER.

\begin{figure}[!t]
\centering
\subfigure[]{
\label{fig:dataset-val} 
\includegraphics[width=0.55\linewidth]{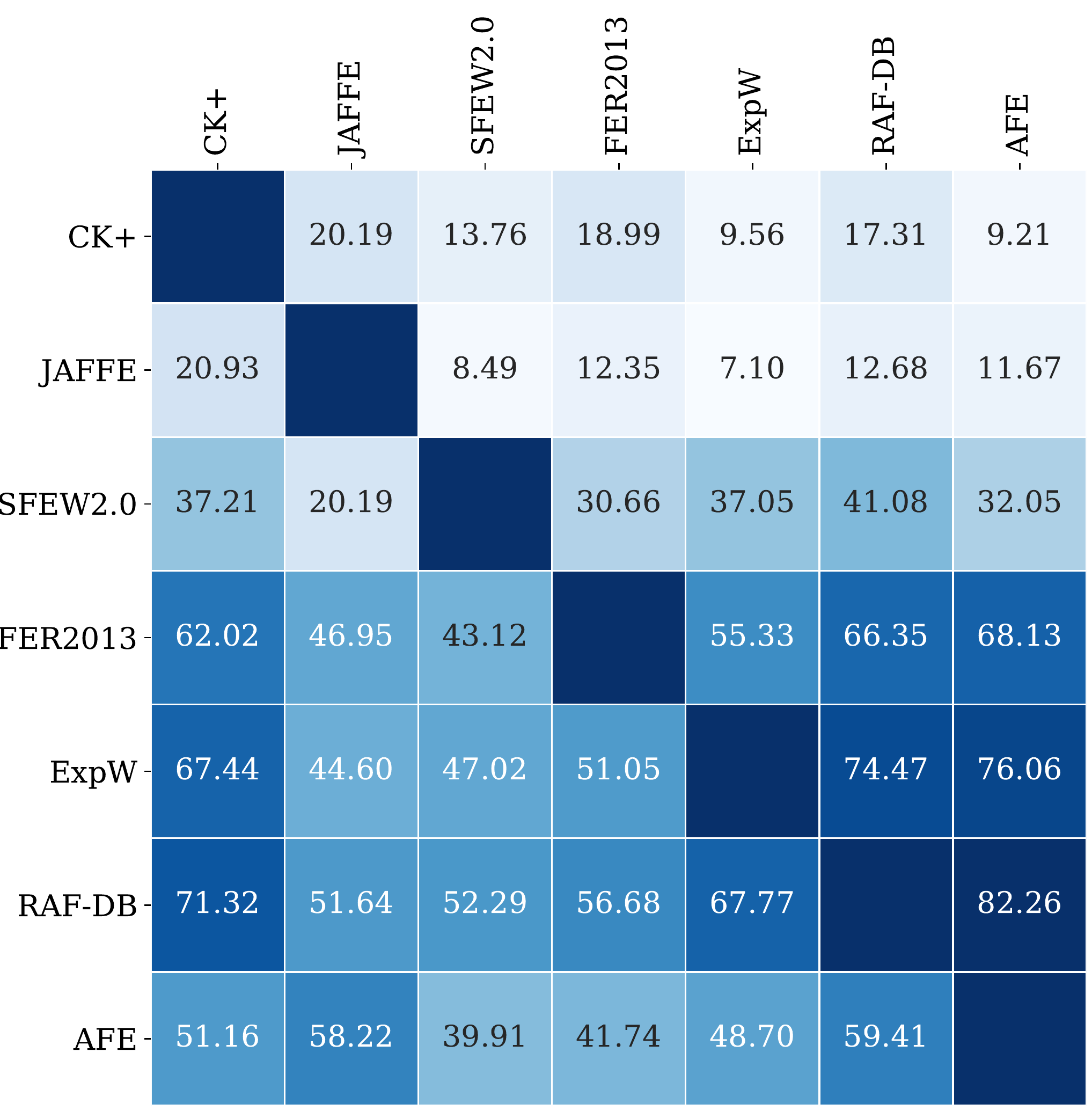}}
\subfigure[]{
\label{fig:backbone-val} 
\includegraphics[width=0.38\linewidth]{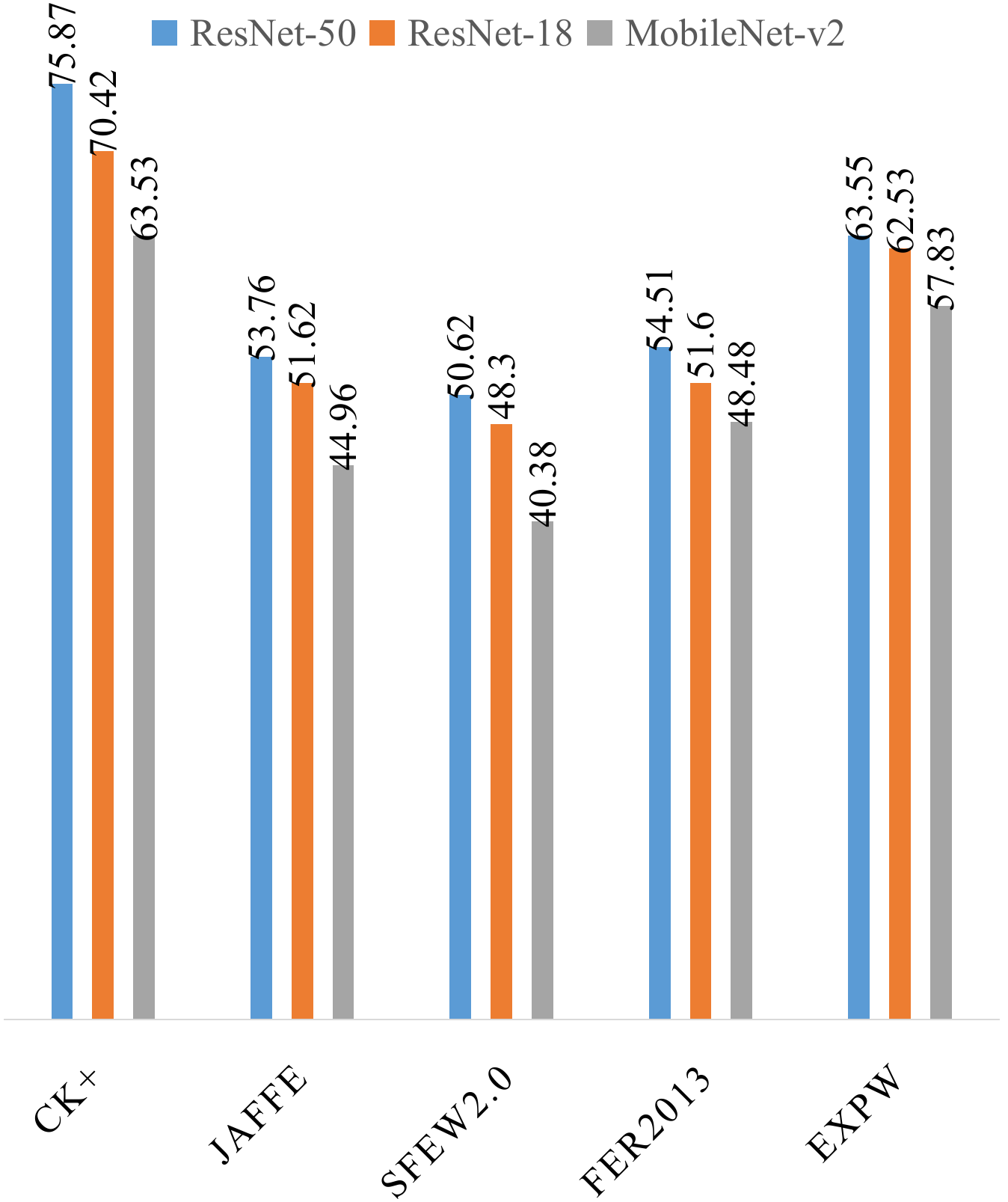}}
\caption{(a) Accuracies of the cross-dataset evaluation using the ResNet-50 baseline and (b) the accuracies of cross-dataset evaluation using the ResNet-50, ResNet-18, and MobileNet-v2 baselines.}
\label{fig:cross-dataset}
\end{figure}

As mentioned above, these datasets are different from each other (see Figure \ref{fig:dataset}). Thus, the different choices of either the source or the target datasets may inevitably lead to a performance gap. As exhibited in Table \ref{table:motivation}, current algorithms select different datasets or even combine multiple datasets as the source, and these algorithms are also evaluated on different target datasets, leading to extremely unfair comparisons. To quantitatively analyze this point, we conduct an experiment that trains a ResNet-50 baseline \cite{he2016deep} on each dataset and tests the network on all the datasets without fine-tuning. As shown in Figure \ref{fig:dataset-val}, we find that testing on the same target dataset but selecting different source datasets may result in a more than 70\% accuracy gap, while using the same source dataset but testing on different target datasets leads to a more than 30\% accuracy gap.

\subsection{Feature Extractors for CD-FER}
Features play a key role in visual recognition tasks, and current algorithms mainly use deep neural networks to extract learnable visual features. However, the features extracted by different deep networks inherently have different discrimination and generalization abilities. Current methods use different feature extractors, ranging from classical Visual Geometry Group (VGG), Inception and ResNet models  \cite{simonyan2015very,szegedy2015going,he2016deep} to different manually designed networks (see Table \ref{table:motivation}), further aggravating the unfairness of  comparisons. 

To clearly show the effect of the feature extractor choice, we conduct an experiment that uses the widely used ResNet-50, ResNet-18, and lightweight MobileNet-v2 \cite{sandler2018mobilenetv2} networks as backbone networks for feature extraction and perform cross-dataset validation by training on the RAF-DB dataset and testing on the CK+, JAFFE, SFEW2.0, FER2013, and ExpW datasets. As shown in Figure \ref{fig:backbone-val}, adopting different backbone networks leads to an accuracy gap of  approximately 10\% when using the same source and target datasets.

\subsection{Algorithms for CD-FER}
Another issue that prevents fair evaluation is that most CD-FER algorithms do not release the codes, and thus, it is difficult to evaluate different algorithms with the same source/target datasets and feature extractors. To address these issues, we re-implement several fruitful CD-FER methods according to the detailed descriptions in each paper, including the ICID algorithm \cite{ji2019cross}, discriminative feature adaptation (DFA) \cite{zhu2016discriminative}, the locality-preserving loss (LPL) \cite{li2017reliable}, a deep emotion transfer network (DETN) \cite{li2018deep}, a fine-tuned deep convolutional network (FTDNN) \cite{zavarez2017cross}, and an ECAN \cite{li2020deeper}. In addition, many general domain adaptation algorithms exist, and we adapt some algorithms to address the CD-FER task. To this end, we select several recently published and advanced algorithms, i.e., conditional adversarial domain adaptation (CADA) \cite{long2018conditional}, the stepwise adaptive feature norm (SAFN) \cite{xu2019larger}, the sliced Wasserstein discrepancy (SWD) \cite{lee2019sliced}, Joint Unbalanced MiniBatch OT (JUMBOT) \cite{fatras2021unbalanced}, Enhanced Transport Distance (ETD) \cite{li2020enhanced}, and use the codes released by the authors for implementation. We also implement two more baselines that can be easily adapted to address the CD-FEE task, namely, direct transferring (DT) that directly transfers the model trained on the source domain to test samples of the target domain, and pseudo-label fine-tuning (PLFT) that uses the model to annotate the target domain (pseudo labels) and then conducts fine-tuning on the target domain.

\subsection{Unified Evaluation Protocols}
The evaluation benchmark currently contains fourteen competing algorithms, including our AGRA method. For fair comparisons, we provide unified evaluation protocols to ensure that all the algorithms can be evaluated with the same source/target datasets and the same feature extractor. Through the provided protocols, the competing algorithms can feasibly select different options of source/target datasets and feature extractors. More details are described as follows.

\noindent\textbf{Dataset choice. } In the evaluation benchmark, all the seven datasets presented in Sec. \ref{sec:datasets} can be used as source or target datasets. Due to limited space, we present in the paper the comparison results of all the methods using two representative datasets (i.e.,  RAF-DB and AFE) as the source datasets. The reason is that RAF-DB can achieve the overall best cross-dataset testing performance (see Figure \ref{fig:dataset-val}) and the AFE dataset can be used to evaluate the cross-culture FER ability of the methods. The corresponding evaluation results are exhibited in Table \ref{table:fair-evaluation-results}. For more comprehensive evaluation, we have conducted experiments using each of seven datasets as the source domain and the remaining ones as the target domain. The comparison results are reported in the supplementary materials (Tables \ref{table:fair-evaluation-results-appendix-resnet50}, \ref{table:fair-evaluation-results-appendix-resnet18}, and \ref{table:fair-evaluation-results-appendix-mobilenetv2}).

\noindent\textbf{Feature extractor choice. } We unify the feature extractors for all competing methods to eliminate the effect of feature inconsistency. Here, we adopt ResNet-50, ResNet-18, and MobileNet-v2 as the backbone networks for the feature extractors. The reason is that ResNet-50 and ResNet-18 are most widely used for visual recognition and MobileNet-v2 is lightweight and can be adapted for mobile device applications.

In addition, our proposed AGRA method aggregates both holistic and local features to facilitate CD-FER, while the above-mentioned competing methods just use holistic features. To ensure fair comparison between our method and the competing methods, we follow the same process as described in Sec. \ref{Network architecture} to extract holistic and local features, and concatenate them as the input for all these competing methods. The results that use ResNet-50 as the backbone network and RAF-DB as the source dataset are presented in Table \ref{table:motivation}, and more results are shown in Sec. \ref{sec:experiments} and in the supplemental materials.

\begin{figure}[!t]
   \centering
   \includegraphics[width=0.8\linewidth]{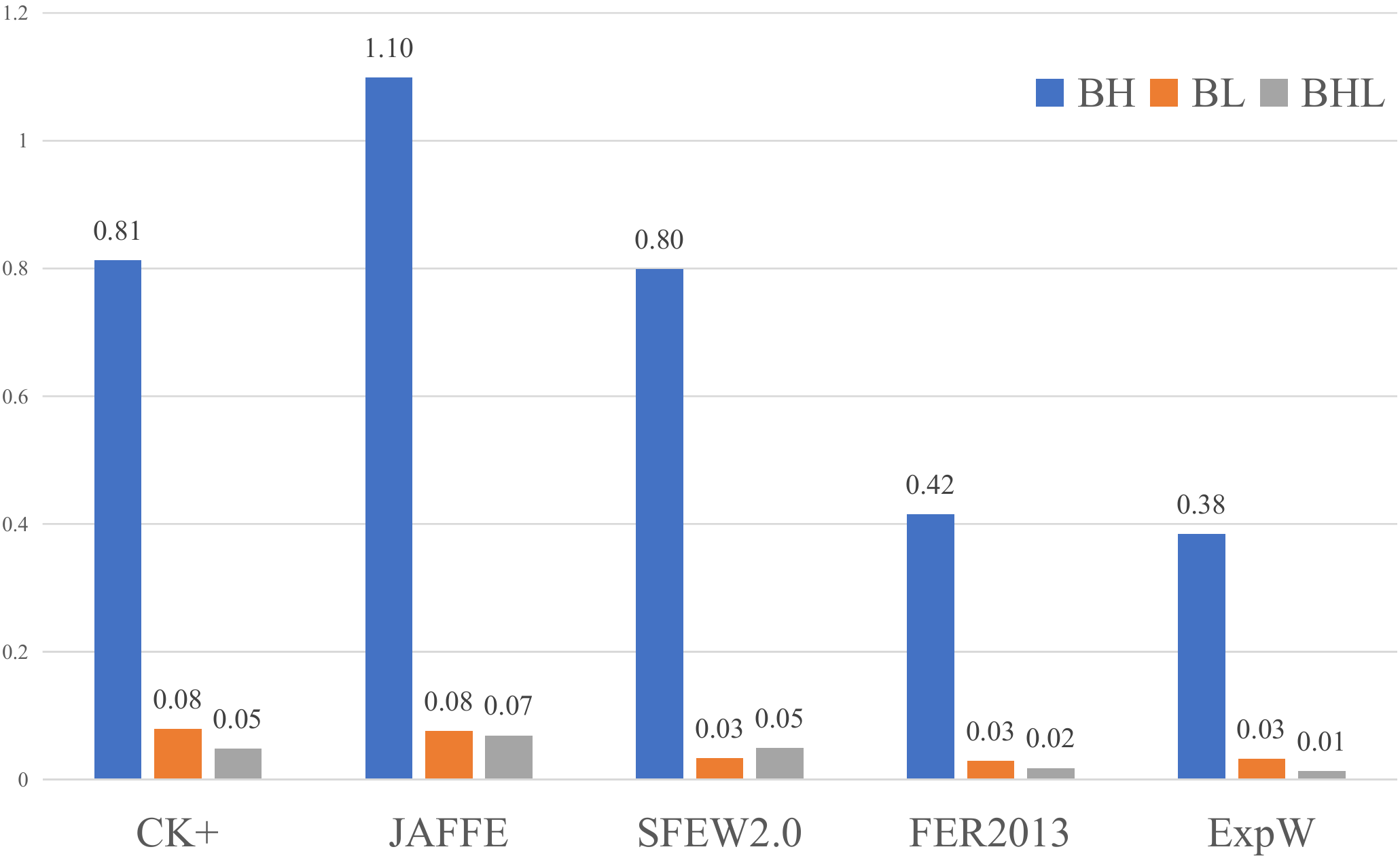}
\caption{The MMD comparisons between features of the source RAF-DB and those of each target CK+, JAFFE , FER2013, SFEW2.0, ExpW for the baseline with holistic features (BH), baseline with local features (BL), and baseline with holistic-local features (BHL).}
   \label{fig:mmd}
\end{figure}

\begin{figure*}[!t]
   \centering
   \includegraphics[width=0.9\linewidth]{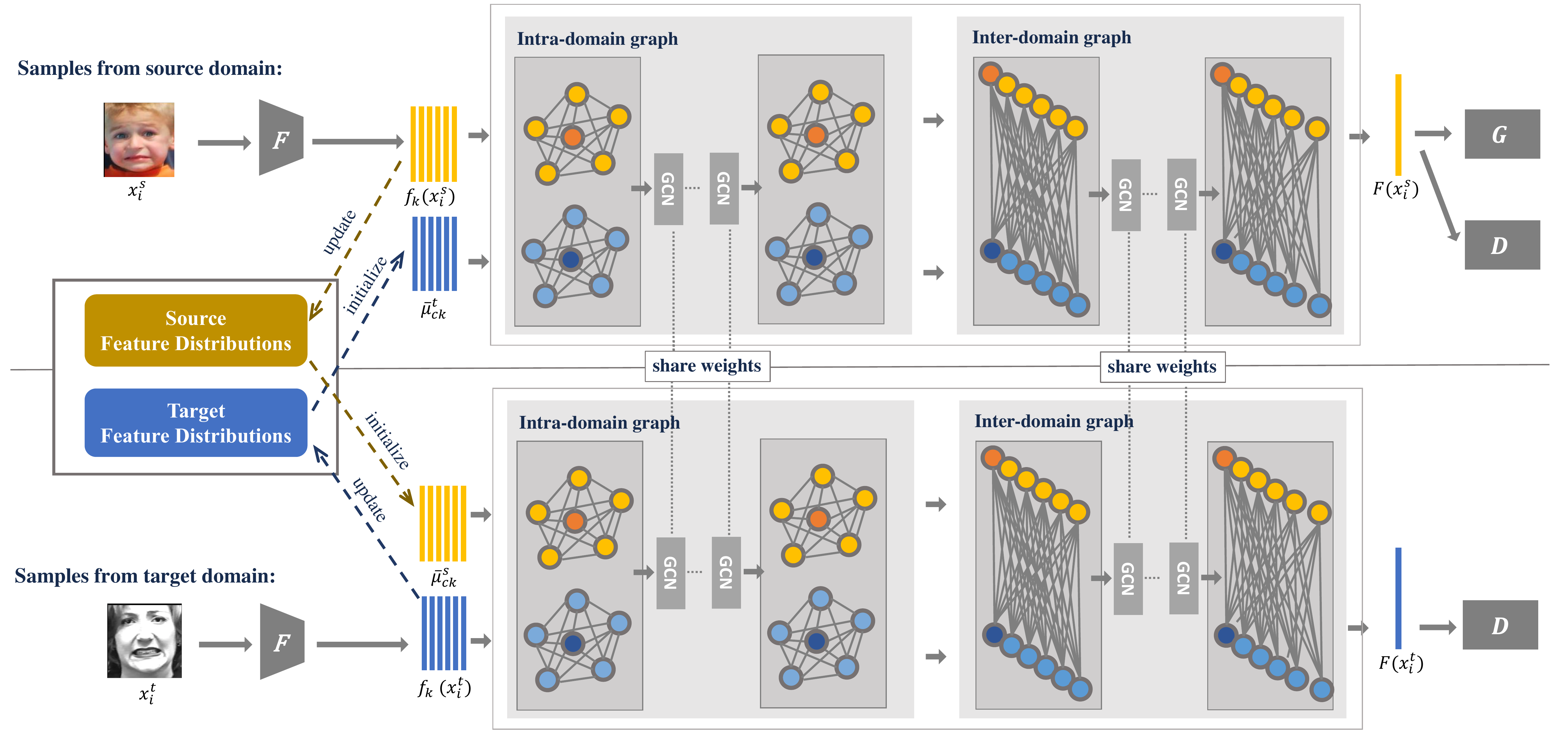}
\vspace{-10pt}
\caption{An illustration of the proposed AGRA framework. The framework builds two graphs to correlate holistic-local features within each domain and across different domains, initializes the graph nodes with input image features of a certain domain and the learnable statistical distribution of the other domain, and introduces two stacked GCNs to propagate node information with each domain and transfer node messages across different domains for holistic-local feature co-adaptation. Note that the nodes in the intra-domain and inter-domain graphs are the same, and we arrange them in different layouts for a clearer illustration of the connections. The feature extractor $F$ and domain discriminator $D$ are also the same for the source and target domains, and we present two weight-sharing branches merely for simple illustration.}
   \label{fig:framework}
\end{figure*}

\section{AGRA Framework}
\label{sec:AGRA}

\subsection{Overview}
\label{sec:overview}
In the CD-FER task, a source domain FER dataset $\mathcal{D}_s=\{(x^s_i, y^s_i)\}_{i=1}^{n_s}$ and a target domain FER dataset $\mathcal{D}_t=\{(x^t_i)\}_{j=1}^{n_t}$ are provided. The two datasets are sampled from two different distributions: $p_s(X, Y)$ and $p_t(X, Y)$. Each sample from the source data $x^s_i$ has a label $y^s_i$, while the samples from the target dataset do not have labels. A learned model is required to perform well on the target dataset.

To address this task, the proposed AGRA framework builds on the adversarial cross-domain mechanism that learns domain-invariant features via two-player games
\begin{eqnarray}
&&\mathop{\min}\limits_{D}  \mathcal{L}(F, G, D) \label{eqn:al1}\\
&&\mathop{\min}\limits_{G, F} \mathcal{L}(F, G) - \mathcal{L}(F, G, D) \label{eqn:al2}
\end{eqnarray}
where
\begin{equation}
   \begin{split}
    \mathcal{L}(F, G)=&-\mathbb{E}_{(x^s, y^s)\sim \mathcal{D}_s}\ell(G(F(x^s)), y^s)\\
    \mathcal{L}(F, G, D)=&-\mathbb{E}_{(x^s, y^s)\sim \mathcal{D}_s}\log\left[D(F(x^s))\right]\\
    & - \mathbb{E}_{x^t\sim \mathcal{D}_t}\log\left[1-D(F(x^t))\right] \\
   \end{split}
   \label{eq:al}
\end{equation}
Here, $F$ is the feature extractor, $G$ is the classifier, and $D$ is the domain discriminator. As suggested in the above two objectives, the feature extractor targets on generating transferable features that can foolish the domain discriminator, while the domain discriminator aims to distinguish the samples of the source domain from those of the target domain. In this way, it can gradually reduce the domain shift and learn domain-invariant image features that are transferable across both the source and target domains. Thus, the classifier trained with only the labeled samples from the source domain can be used to classify samples from both domains.

Many works have applied the above adversarial mechanism to domain adaptation tasks, but they mainly extract holistic features for domain adaptation and usually ignore local patterns that are more transferable. With regard to the CD-FER task, these local features are also valuable, as this task requires a fine-grained and detailed understanding of the face images. For a more direct and in-depth analysis, we conduct experiments on the generalization abilities of different types of features. Specifically, we design three baselines with the same ResNet-50 backbone and use three different heads to extract holistic, local, and holistic-local features, namely BH, BL, BHL. We use the same process as described in Sec. \ref{Network architecture} to extract the holistic and local feature vectors. For the BH baseline, we directly use the 64-dimensional holistic feature vector. For the BL baseline, we concatenate the five local feature vectors and use a fully-connected layer followed by an rectified linear unit (ReLU) function to map the concatenated feature vectors to a 64-dimensional vector. For the BHL baseline, we concatenate the holistic and five local feature vectors and use a fully-connected layer followed by an ReLU function to map the concatenated feature vector to a 64-dimensional vector. Then, we train the three baselines on RAF-DB and use the trained models to extract features for the samples of both the source RAF-DB and target CK+ \cite{lucey2010extended}, JAFFE \cite{lyons1998coding}, FER2013 \cite{goodfellow2015challenges}, SFEW2.0 \cite{dhall2011static}, and ExpW \cite{zhang2018facial}. Finally, we compute the maximum mean discrepancy (MMD) \cite{dalib} between features of the source RAF-DB dataset and those of each target dataset. As shown in Figure \ref{fig:mmd}, using the local features can greatly decrease the MMDs compared with using the holistic features, and integrating holistic-local features can generally further decrease the MMDs. These comparisons suggest that local features have better generalization abilities than holistic features and that integrating holistic-local features can further strengthen these abilities. The way to effectively integrating the two for CD-FER is worthy investigating.

In this work, we propose to represent correlation of holistic-local features in structured graphs and integrate graph propagation networks with adversarial learning mechanisms to learn domain-invariant holistic-local features for CD-FER. To this end, we extract several discriminative regions based on facial landmarks and build an intra-domain graph to correlate holistic-local regions within each domain and an inter-domain graph to correlate these regions across different domains. We develop a class-aware two-stage updating mechanism to iteratively learn the per-class statistical feature distributions for both the holistic and local regions from both domains. Given an input image from one domain, we extract the holistic-local features from corresponding regions to initialize the graph nodes of this domain and apply the statistical feature distribution to initialize the graph nodes of the other domain. Finally, we use two stacked GCNs to propagate messages through the intra-domain graph to explore holistic-local feature interactions and transfer information across the inter-domain graph to enable holistic-local co-adaptation. An overall pipeline of the proposed AGRA framework is illustrated in Figure \ref{fig:framework}.

\subsection{Graph Construction}
In this section, we introduce the constructions of the intra-domain and inter-domain graphs. According to the FACS \cite{friesen1978facial,tian2001recognizing}, facial expression can be interpreted as facial action units, and most of these action units are defined at the regions centered on the left eye (\textit{le}), right eye (\textit{re}), nose (\textit{no}), left mouth corner (\textit{lm}), and right mouth corner (\textit{rm}). Thus, these regions contain the most detailed information for FER, and we extract the holistic face and further crop the five local regions. We then build the two graphs $\mathcal{G}_{intra}=(\mathbf{V}, \mathbf{A}_{intra})$ and $\mathcal{G}_{inter}=(\mathbf{V}, \mathbf{A}_{inter})$. $\mathbf{V}=\{v^s_h, v^s_{le}, v^s_{re}, v^s_{no}, v^s_{lm}, v^s_{rm}, v^t_h, v^t_{le}, v^t_{re}, v^t_{no}, v^t_{lm}, v^t_{rm}\}$ is the node set denoting the holistic image and five local regions of the source and target domains, and it is the same for both graphs. $\mathbf{A}_{intra}$ is the prior intra-domain adjacency matrix denoting the connections among nodes within each domain. It contains two types of connections;  the first type is holistic-to-local connections, and the second type is local-to-local connections. $\mathbf{A}_{inter}$ is the prior inter-domain adjacency matrix denoting the connections between nodes from the different domains. Similarly, it contains three types of connections: holistic-to-holistic connections, holistic-to-local connections, and local-to-local connections. We use different values to denote different connections.

\subsection{Graph Representation Adaptation}
Once the two graphs are constructed, message propagations are performed through the intra-domain graph to explore holistic-local feature interactions with each domain and through the inter-domain graph to enable holistic-local feature co-adaptation. As suggested in previous works \cite{kipf2017semi}, GCNs \cite{kipf2017semi} can effectively update node features of graph-structured data by iteratively propagating node massages to the neighborhood nodes. In this work, we apply two stacked GCNs to propagate messages through the two graphs.

As discussed above, the graphs contain nodes from two domains. Given an input sample of one domain $d$ $(d\in\{s, t\})$, we extract the features of the corresponding regions to initialize the nodes of domain $d$. It is expected that these features can interact with the feature distributions of the other domain, and thus, the model can gradually reduce  the domain shift. In addition, motivated by a previous work \cite{wang2018stratified}, it is essential to integrate class information to enable finer-grained intraclass interaction and adaptation. To this end, we estimate the per-class statistical feature distributions of each domain, i.e., $\bar\mu^s_{ck}$ and $\bar\mu^t_{ck}$, where $c\in \{0, 1, \dots, C-1\}$ is the class label and $k\in \{h, le, re, no, lm, rm\}$ is the node type. This estimation is implemented by a class-aware two-stage updating mechanism as follows.

\subsubsection{Class-aware two-stage updating mechanism}
Here, we update the statistical distribution by epoch-level clustering that reclusters the samples to obtain the distribution every $E$ epochs and iteration-level updating that updates the distribution every iteration. Specifically, we first extract features for all the  samples from both the source and target datasets using the backbone network pretrained using the labeled source samples. For each domain, we divide the samples into $C$ clusters using the K-means clustering algorithm and compute the means for each cluster to obtain the initial statistical distribution, which is formulated as
\begin{equation}
   \begin{split}
    \bar{\mu}^s_{ck}&=\frac{1}{n^s_c}\sum_{i=1}^{n^s_c}f_k(x^s_{ci})\\
    \bar{\mu}^t_{ck}&=\frac{1}{n^t_c}\sum_{j=1}^{n^t_c}f_k(x^t_{ci})
   \end{split}
   \label{eq:average1}
\end{equation}
where $f_k(\cdot)$ is the feature extractor for region $k$; $n^{s/t}_c$ is the number of samples in cluster $c$ of domain $s/t$; and $x^{s/t}_{ci}$ is the $i$-th sample of cluster $c$. During training, we further use the moving average to iteratively update these statistical distributions in a progressive manner. For each batch iteration, we compute the distances between each sample and the distributions of each cluster. These samples are grouped into the cluster with the smallest distance. Then, we compute the mean features (i.e., ${\mu}^s_{ck}$ and ${\mu}^t_{ck}$) over the samples in the same cluster and update the statistical distribution by
\begin{equation}
   \begin{split}
    \bar{\mu}^s_{ck}&=(1-\alpha)\bar{\mu}^s_{ck}+\alpha {\mu}^s_{ck}\\
    \bar{\mu}^t_{ck}&=(1-\alpha)\bar{\mu}^t_{ck}+\alpha {\mu}^t_{ck}
   \end{split}
   \label{eq:average2}
\end{equation}
where $\alpha$ is a balance parameter, which is set to 0.1 in our experiments. To avoid distribution shifting, this process is repeated every $E$ epochs. Then, we recluster the samples to obtain new distributions for each cluster according to equation \ref{eq:average1}. Epoch-level reclustering and iteration-level updating are iteratively performed along with the training process to obtain the final statistical distributions.

\subsubsection{Stacked graph convolutional networks}
As discussed above, we use two stacked GCNs: one GCN propagates messages through the intra-domain graph to explore holistic-local feature interactions within each domain,  and the GCN transfers messages through the inter-domain GCN to enable holistic-local feature co-adaptation. In this section, we describe the two GCNs in detail.

Given an input sample $x^s_i$ from the source domain, we can extract features of the holistic image and the corresponding local regions to initialize the corresponding node of the source domain
\begin{equation}
   h_{intra, k}^{s,0}=f_{k}(x^s_i).
 \end{equation}
 
Then, we compute the distance between this sample and the feature distributions of all clusters of the target domain and obtain the cluster $c$ with the smallest distance. Then, each node of the target domain is initialized by the corresponding feature distribution
\begin{equation}
  h_{intra, k}^{t,0}=\bar{\mu}^t_{ck}.
\end{equation}
The initial features are then rearranged to obtain feature matrix $\mathbf{H}^0_{intra}\in \mathcal{R}^{n\times d^0_{intra}}$, where $n=12$ is the number of nodes. Then, we perform a graph convolution operation on the input feature matrix to iteratively propagate and update the node features, which is formulated as
\begin{equation}
   \mathbf{H}^l_{intra}=\sigma(\widehat{\mathbf{A}}_{intra}\mathbf{H}^{l-1}_{intra}\mathbf{W}^{l-1}_{intra}),
\end{equation}
By stacking $L_{intra}$ graph convolutional layers, the node messages are fully explored within the intra-domain graph, and the feature matrix $\mathbf{H}_{intra}$ is obtained. This feature matrix is then used to initialize the nodes of the inter-domain graph
\begin{equation}
  \mathbf{H}^0_{inter}=\mathbf{H}_{intra}.
\end{equation}
The graph convolution operation is performed to iteratively update the node features
\begin{equation}
   \mathbf{H}^l_{inter}=\sigma(\widehat{\mathbf{A}}_{inter}\mathbf{H}^{l-1}_{inter}\mathbf{W}^{l-1}_{inter}),
\end{equation}
Similarly, the graph convolution operation is repeated $L_{inter}$ times, and the final feature matrix $\mathbf{H}$ is generated. We concatenate the features of nodes from the source domain as the final feature $F(x^s_i)$, which is fed into the classifier to predict the expression label and domain discriminator to estimate its domain. The two matrices $\widehat{\mathbf{A}}_{intra}$ and $\widehat{\mathbf{A}}_{inter}$ are initialized by the prior matrices $\mathbf{A}_{intra}$ and $\mathbf{A}_{inter}$ and jointly fine-tuned to learn better relationships during the training process.

Similarly, given a sample from the target domain, the nodes of the source domain are initialized by the corresponding extracted feature, and those of the target domain are initialized by the corresponding statistical feature distributions. Then, the same process is conducted to obtain the final feature $F(x^t_i)$. As the final feature does not have expression label annotation, it is merely fed into the domain discriminator for domain estimation.

\subsection{Implementation Details}

\subsubsection{Network architecture}
\label{Network architecture}
As stated in previous works, we use ResNet-50, ResNet-18, and MobileNet-v2 \cite{he2016deep,zhao2019multi} as backbone networks for feature extraction. All three networks consist of four block layers. Given an input image with a size of $112 \times 112 \times 3$, we can obtain feature maps with the size of  $28 \times 28 \times 128$ from the second layer and feature maps with the size of $7\times 7\times 512$ from the fourth layer. For the holistic features, we perform a convolution operation to obtain feature maps with the size of $7 \times 7 \times 64$, which is followed by an average pooling layer to obtain a 64-dimensional vector. For the local features, we use a multi-task CNN (MT-CNN) \cite{zhang2016joint} to locate the landmarks and use the feature maps from the second layer as they have a higher resolution. Specifically, we crop $7 \times 7 \times 128$ feature maps centered at the corresponding landmark and use similar convolution operations and average pooling to obtain a 64-dimensional vector for each region.

The intra-domain GCN consists of two graph convolutional layers with 128 and 64 output channels. Thus, the sizes of the parameter matrices $\mathbf{W}^{0}_{intra}$ and $\mathbf{W}^{1}_{intra}$ are $64\times 128$ and $128\times 64$, respectively. The inter-domain GCN contains only one graph convolutional layer, and the number of output channels is also set to 64. The parameter matrix $\mathbf{W}^{0}_{inter}$ has a size of $64\times 64$. We perform ablation studies to analyze the effect of the number of layers of the two GCNs and find setting them to 2 and 1 obtains the best results.

The classifier is simply implemented by a fully connected layer that maps the 384-dimensional (i.e., $64\times 6$) feature vector to seven scores that indicate the confidence of each expression label. The domain discriminator is implemented by two fully connected layers with an ReLU nonlinear function, followed by another fully connected layer to one score to indicate its domain.

\begin{table*}[htp]
\scriptsize
\centering
\begin{tabular}{c|c|c|cccccc}
\hline
\centering  Method & Source set & Backbone & CK+ & JAFFE & SFEW2.0 & FER2013 & ExpW & Mean\\
\hline
\hline
DT & RAF-DB & ResNet-50 & 71.32 & 50.23 & 50.46 & 54.49 & 67.45 & 58.79 \\
PLFT & RAF-DB & ResNet-50 & 77.52 & 53.99 & 48.62 & 56.46 & 69.81 & 61.28 \\
ICID \cite{ji2019cross} & RAF-DB & ResNet-50 & 74.42 & 50.70 & 48.85 & 53.70 & \textbf{69.54} & 59.44 \\ 
DFA \cite{zhu2016discriminative} & RAF-DB & ResNet-50 & 64.26 & 44.44 & 43.07 & 45.79 & 56.86 & 50.88 \\
LPL \cite{li2017reliable} & RAF-DB & ResNet-50 & 74.42 & 53.05 & 48.85 & 55.89 & 66.90 & 59.82 \\
DETN \cite{li2018deep} & RAF-DB & ResNet-50 & 78.22 & 55.89 & 49.40 & 52.29 & 47.58 & 56.68 \\
FTDNN \cite{zavarez2017cross} & RAF-DB & ResNet-50 & 79.07 & 52.11 & 47.48 & 55.98 & 67.72 & 60.47 \\
ECAN \cite{li2020deeper} & RAF-DB & ResNet-50 & 79.77 & 57.28 & 52.29 & 56.46 & 47.37 & 58.63 \\
CADA \cite{long2018conditional} & RAF-DB & ResNet-50 & 72.09 & 52.11 & 53.44 & 57.61 & 63.15 & 59.68\\
SAFN \cite{xu2019larger} & RAF-DB & ResNet-50 & 75.97 & 61.03 & 52.98 & 55.64 & 64.91 & 62.11 \\
SWD \cite{lee2019sliced} & RAF-DB & ResNet-50 & 75.19 & 54.93 & 52.06 & 55.84 & 68.35 & 61.27 \\
JUMBOT \cite{fatras2021unbalanced} & RAF-DB & ResNet-50 & 79.46 & 54.13 & 51.97 & 53.56 & 63.69 & 60.56 \\
ETD \cite{li2020enhanced} & RAF-DB & ResNet-50 & 75.16 & 51.19 & 52.77 & 50.41 & 67.82 & 59.47 \\
\hline
\textbf{Ours} & RAF-DB & ResNet-50 & \textbf{85.27} & \textbf{61.50} & \textbf{56.43} & \textbf{58.95}  & 68.50 & \textbf{66.13} \\
\hline
\hline
DT & RAF-DB & ResNet-18 & 68.22 & 49.30 & 49.31 & 52.71 & 67.63 & 57.34 \\
PLFT & RAF-DB & ResNet-18 & 74.42 & 53.05 & 48.62 & 54.21 & 69.06 & 59.67 \\
ICID \cite{ji2019cross} & RAF-DB & ResNet-18 & 67.44 & 48.83 & 47.02 & 53.00 & 68.52 & 56.96 \\ 
DFA \cite{zhu2016discriminative} & RAF-DB & ResNet-18 & 54.26 & 42.25 & 38.30 & 47.88 & 47.42 & 46.02 \\
LPL \cite{li2017reliable} & RAF-DB & ResNet-18 & 72.87 & 53.99 & 49.31 & 53.61 & 68.35 & 59.63 \\
DETN \cite{li2018deep} & RAF-DB & ResNet-18 & 64.19 & 52.11 & 42.25 & 42.01 & 43.92 & 48.90 \\
FTDNN \cite{zavarez2017cross} & RAF-DB & ResNet-18 & 76.74 & 50.23 & 49.54 & 53.28 & 68.08 & 59.57 \\
ECAN \cite{li2020deeper} & RAF-DB & ResNet-18 & 66.51 & 52.11 & 48.21 & 50.76 & 48.73 & 53.26 \\
CADA \cite{long2018conditional} & RAF-DB & ResNet-18 & 73.64 & 55.40 & 52.29 & 54.71 & 63.74 & 59.96 \\
SAFN \cite{xu2019larger} & RAF-DB & ResNet-18 & 68.99 & 49.30 & 50.46 & 53.31 & 68.32 & 58.08 \\
SWD \cite{lee2019sliced} & RAF-DB & ResNet-18 & 72.09 & 53.52 & 49.31 & 53.70 & 65.85 & 58.89 \\
JUMBOT \cite{fatras2021unbalanced} & RAF-DB & ResNet-18 & 76.67 & 52.10 & 49.19 & 50.58 & 61.45 & 58.00 \\
ETD \cite{li2020enhanced} & RAF-DB & ResNet-18 & 72.34 & 49.44 & 49.67 & 47.66 & 64.62 & 56.75 \\
\hline
\textbf{Ours} & RAF-DB & ResNet-18 & \textbf{77.52} & \textbf{61.03} & \textbf{52.75} & \textbf{54.94}  & \textbf{69.70} & \textbf{63.19} \\
\hline
\hline
DT  & RAF-DB & MobileNet-V2 & 66.67 & 38.97 & 41.74 & 49.99 & 63.08 & 52.09 \\
PLFT & RAF-DB & MobileNet-V2 & 72.09 & 38.97 & 41.97 & \textbf{51.11} & \textbf{64.12} & 53.65 \\
ICID \cite{ji2019cross} & RAF-DB & MobileNet-v2 & 57.36 & 37.56 & 38.30 & 44.47 & 60.64 & 47.67 \\ 
DFA \cite{zhu2016discriminative} & RAF-DB & MobileNet-v2 & 41.86& 35.21 & 29.36 & 42.36&43.66 & 38.49 \\
LPL \cite{li2017reliable} & RAF-DB & MobileNet-v2 & 59.69 & 40.38 & 40.14 & 50.13 & 62.26 & 50.52 \\
DETN \cite{li2018deep} & RAF-DB & MobileNet-v2 & 53.49 & 40.38 & 35.09 & 45.88 & 45.26 & 44.02 \\
FTDNN \cite{zavarez2017cross} & RAF-DB & MobileNet-v2 & 71.32 & 46.01 & 45.41 & 49.96 & 62.87 & 55.11 \\
ECAN \cite{li2020deeper} & RAF-DB &MobileNet-v2 & 53.49 & 43.08 & 35.09 & 45.77 & 45.09 & 44.50 \\
CADA \cite{long2018conditional} & RAF-DB & MobileNet-v2 &62.79 & 53.05& 43.12 & 49.34  & 59.40 & 53.54 \\
SAFN \cite{xu2019larger} & RAF-DB & MobileNet-v2 & 66.67 & 45.07 & 40.14 & 49.90 & 61.40 & 52.64 \\
SWD \cite{lee2019sliced} & RAF-DB & MobileNet-v2 & 68.22 & 55.40 & 43.58 & 50.30 & 60.04 & 55.51 \\
JUMBOT \cite{fatras2021unbalanced} & RAF-DB & MobileNet-v2 & \textbf{73.64} & 51.35 & 44.41 & 49.05 & 60.84 & 55.86 \\
ETD \cite{li2020enhanced} & RAF-DB & MobileNet-v2 & 69.27 & 48.57 & 41.34 & 49.43 & 57.05 & 53.13 \\
\hline
\textbf{Ours} & RAF-DB & MobileNet-v2 & 72.87 & \textbf{55.40} & \textbf{45.64} & 51.05 & 63.94 & \textbf{57.78} \\
\hline
\hline
DT & AFE & ResNet-50 & 61.24 & 57.28 & 46.79 & 47.79 & 52.03 & 53.03 \\
PLFT & AFE & ResNet-50 & 68.22 & 58.22 & 45.41 & 48.97 & 53.72 & 54.91 \\
ICID \cite{ji2019cross} & AFE & ResNet-50 & 56.59 & 57.28 & 44.27 & 46.92 & 52.91 & 51.59\\ 
DFA \cite{zhu2016discriminative} & AFE & ResNet-50 & 51.86 & 52.70 & 38.03 & 41.93 & 60.12 & 48.93\\
LPL \cite{li2017reliable} & AFE & ResNet-50 & 73.64 & 61.03 & 49.77 & 49.54 & 55.26 & 57.85 \\
DETN \cite{li2018deep} & AFE & ResNet-50 & 56.27 & 52.11 & 44.72 & 42.17 & 59.80 & 51.01 \\
FTDNN \cite{zavarez2017cross} & AFE & ResNet-50 & 61.24 & 57.75 & 47.25 & 46.36 & 52.89 & 53.10 \\
ECAN \cite{li2020deeper} & AFE & ResNet-50 & 58.14 & 56.91 & 46.33 & 46.30 & 61.44 & 53.82 \\
CADA \cite{long2018conditional} & AFE & ResNet-50 & 72.09 & 49.77 & 50.92 & 50.32 & 61.70 & 56.96 \\
SAFN \cite{xu2019larger} & AFE & ResNet-50 & 73.64 & 64.79 & 49.08 & 48.89 & 55.69 & 58.42 \\
SWD \cite{lee2019sliced} & AFE & ResNet-50 & 72.09 & 61.50 & 48.85 & 48.83 & 56.22 & 57.50 \\
JUMBOT \cite{fatras2021unbalanced} & AFE & ResNet-50 & 75.36 & 54.38 & 48.38 & 50.75 & 60.74 & 57.92 \\
ETD \cite{li2020enhanced} & AFE & ResNet-50 & 73.12 & 51.43 & 49.71 & 50.34 & 62.37 & 57.39 \\
\hline
\textbf{Ours} & AFE & ResNet-50 & \textbf{78.57} & \textbf{65.43} & \textbf{51.18} & \textbf{51.31}  & \textbf{62.71} & \textbf{61.84} \\
\hline
\hline
DT & AFE & ResNet-18 & 71.32 & 63.38 & 52.52 & 51.03 & 54.65 & 58.58 \\
PLFT & AFE & ResNet-18 & 78.29 & 59.62 & 51.15 & \textbf{51.98} & 57.85 & 59.78 \\
ICID \cite{ji2019cross} & AFE & ResNet-18 & 54.26 & 51.17 & 47.48 & 46.44 & 54.85 & 50.84 \\ 
DFA \cite{zhu2016discriminative} & AFE & ResNet-18 & 35.66 & 45.82 & 34.63 & 36.88 & 62.53 & 43.10 \\
LPL \cite{li2017reliable} & AFE & ResNet-18 & 67.44 & \textbf{62.91} & 48.39 & 49.82 & 54.51 & 56.61 \\
DETN \cite{li2018deep} & AFE & ResNet-18 & 44.19 & 47.23 & 45.46 & 45.39 & 58.41 & 48.14 \\
FTDNN \cite{zavarez2017cross} & AFE & ResNet-18 & 58.91 & 59.15 & 47.02 & 48.58 & 55.29 & 53.79 \\
ECAN \cite{li2020deeper} & AFE & ResNet-18 & 44.19 & 60.56 & 43.26 & 46.15 & 62.52 & 51.34 \\
CADA \cite{long2018conditional} & AFE & ResNet-18 & 72.09 & 53.99 & 48.39 & 48.61 & 58.50 & 56.32 \\
SAFN \cite{xu2019larger} & AFE & ResNet-18 & 68.22 & 61.50 & 50.46 & 50.07 & 55.17 & 57.08 \\
SWD \cite{lee2019sliced} & AFE & ResNet-18 & 77.52 & 59.15 & 50.69 & 51.84 & 56.56 & 59.15 \\
JUMBOT \cite{fatras2021unbalanced} & AFE & ResNet-18 & 68.06 & 53.62 & 47.81 & 48.89 & 60.37 & 55.75 \\
ETD \cite{li2020enhanced} & AFE & ResNet-18 & 72.08 & 50.79 & 48.46 & 49.38 & 61.79 & 56.50 \\
\hline
\textbf{Ours} & AFE & ResNet-18 & \textbf{79.84} & 61.03 & \textbf{51.15} & 51.95  & \textbf{65.03} & \textbf{61.80} \\
\hline
\hline
DT & AFE & MobileNet-V2 & 68.22 & 49.30 & 45.41 & 46.64 & 53.71 & 52.66 \\
PLFT & AFE & MobileNet-V2 & 69.77 & 52.58 & 40.37 & 46.36 & 53.19 & 52.45 \\
ICID \cite{ji2019cross} & AFE & MobileNet-v2 & 55.04 & 42.72 & 34.86 & 39.94 & 44.34 & 43.38 \\ 
DFA \cite{zhu2016discriminative} & AFE & MobileNet-v2 & 44.19 & 27.70 & 31.88 & 35.95 & 61.55 & 40.25 \\
LPL \cite{li2017reliable} & AFE & MobileNet-v2 & 69.77 & 50.23 & 43.35 & 45.57 & 51.63 & 52.11 \\
DETN \cite{li2018deep} & AFE & MobileNet-v2 & 57.36 & 54.46 & 32.80 & 44.11 & \textbf{64.36} & 50.62 \\
FTDNN \cite{zavarez2017cross} & AFE & MobileNet-v2 & 65.12 & 46.01 & 46.10 & 46.69 & 53.02 & 51.39 \\
ECAN \cite{li2020deeper} & AFE & MobileNet-v2 & 71.32 & \textbf{56.40} & 37.61 & 45.34 & 64.00 & 54.93 \\
CADA \cite{long2018conditional} & AFE & MobileNet-v2 & 70.54 & 45.07 & 40.14 & 46.72 & 54.93 & 51.48 \\
SAFN \cite{xu2019larger} & AFE & MobileNet-v2 & 62.79 & 53.99 & 42.66 & 46.61 & 52.65 & 51.74 \\
SWD \cite{lee2019sliced} & AFE & MobileNet-v2 & 64.34 & 53.52 & 44.72 & \textbf{50.24} & 55.85 & 53.73 \\
JUMBOT \cite{fatras2021unbalanced} & AFE & MobileNet-v2 & 67.29 & 54.07 & 46.88 & 47.06 & 54.26 & 53.91 \\
ETD \cite{li2020enhanced} & AFE & MobileNet-v2 & 66.67 & 49.42 & 46.11 & 46.37 & 52.50 & 52.21 \\
\hline
\textbf{Ours} & AFE & MobileNet-v2 & \textbf{75.19} & 54.46  & \textbf{47.25} & 47.88  & 61.10 & \textbf{57.18} \\
\hline
\end{tabular}
\vspace{2pt}
\caption{Accuracies of our proposed framework with current leading methods on the CK+, JAFFE, SFEW2.0, FER2013, and ExpW datasets. The results are generated by our implementation with exactly the same source dataset and backbone network.}
\label{table:fair-evaluation-results}
\end{table*}

\subsubsection{Training details}
\label{Sec:TrainingDetail}
The AGRA framework is trained with the objectives of equations \ref{eqn:al1} and \ref{eqn:al2} to optimize the feature extractor, classifier, and domain discriminator. Here, we follow previous domain adaptation works \cite{wen2019exploiting} to adopt a two-stage training process. We initialize the parameters of the backbone networks with those pretrained on the MS-Celeb-1M \cite{guo2016ms} dataset and the parameters of the newly added layers with the Xavier algorithm \cite{glorot2010understanding}. In the first stage, we train the feature extractor and classifier with the cross-entropy loss using stochastic gradient descent (SGD) with an initial learning rate of 0.0001, a momentum of 0.9, and a weight decay of 0.0005. It is trained for approximately 15 epochs. In the second stage, we use the objective loss in equation \ref{eqn:al1} to train the domain discriminator and the objective loss in equation \ref{eqn:al2} to fine-tune the feature extractor and the classifier. It is also trained using SGD with the same momentum and weight decay as the first stage. The learning rate for the feature extractor and the source classifier is initialized at 0.0001, and it is divided by 10 after approximately 10 epochs. As the domain discriminator is trained from scratch, we initialize it at 0.001 and divide it by 10 when the error saturates.

\subsubsection{Inference details}
Given an input image, we extract holistic and local images to initialize the corresponding nodes of the target domain. Then, we compute the distances between the given image and all the per-class feature distributions of the source domain. We select the feature distributions with the smallest distance to initialize the nodes of the source domain. After GCN message propagation, we can obtain its feature and feed it into the classifier to predict the final score vector.

\section{Experiments}
\label{sec:experiments}
In this section, we present the results of all the re-implemented algorithms and the proposed AGRA framework under the fair evaluation setting and perform an ablation  study to analyze the actual contribution of each component.

\subsection{Performance Evaluation and Analysis}

\subsubsection{Effect of the inconsistent choice}
In this part, we use the unified evaluation benchmark to analyze the performance impact of the inconsistent choices of the source/target datasets and backbone networks. The performances of all the methods are presented in Table \ref{table:fair-evaluation-results}.

\begin{figure}[!t]
   \centering
   \includegraphics[width=0.99\linewidth]{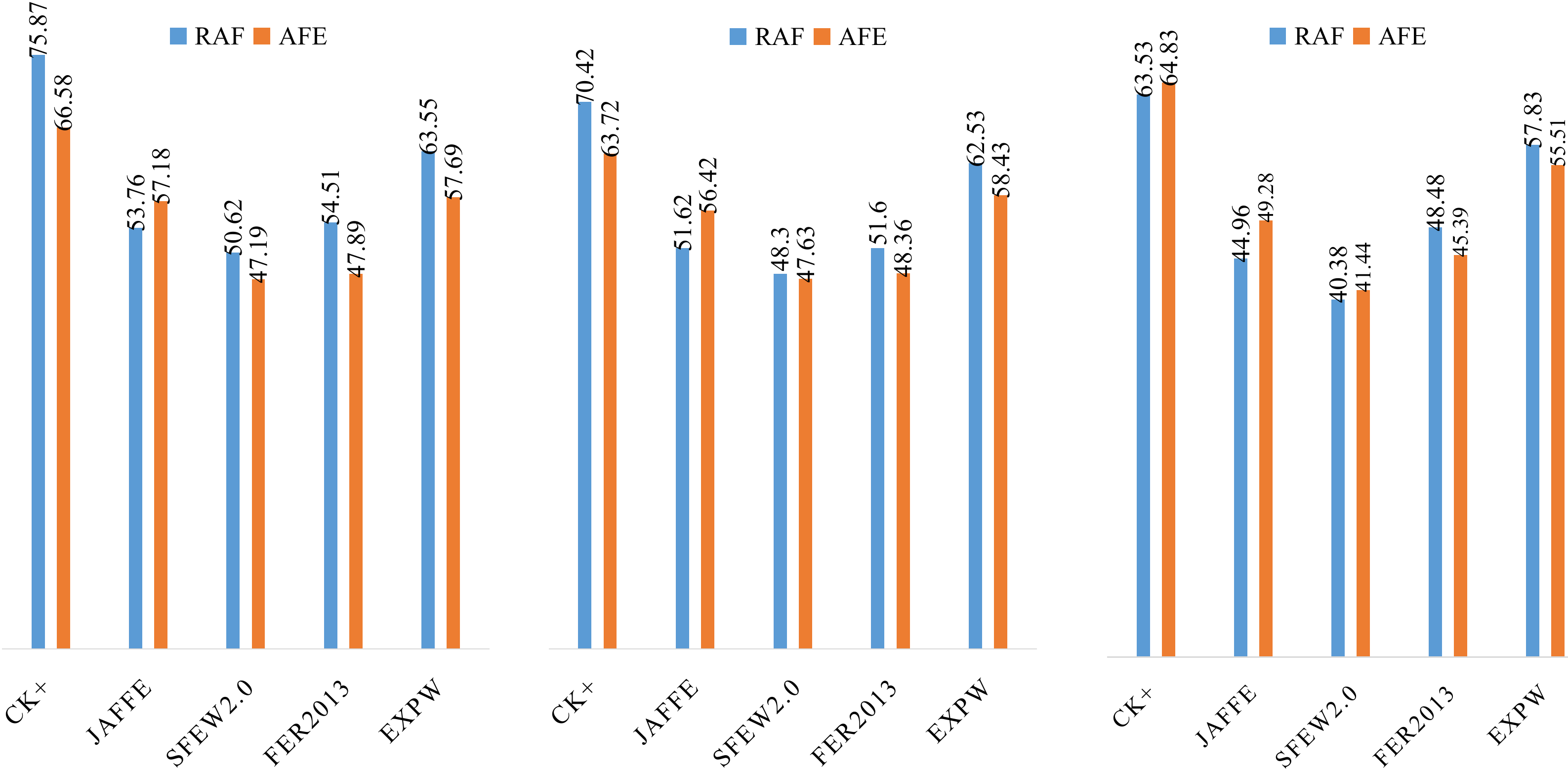}
\caption{The average accuracy comparisons using the RAF-DB and AFE datasets as the source dataset. We average the accuracies of all the methods using ResNet-50 (left), ResNet-18 (middle), and MobileNet-v2 (right) as the backbone. }
   \label{fig:source-comparison}
\end{figure}

\begin{figure}[!t]
   \centering
   \includegraphics[width=0.95\linewidth]{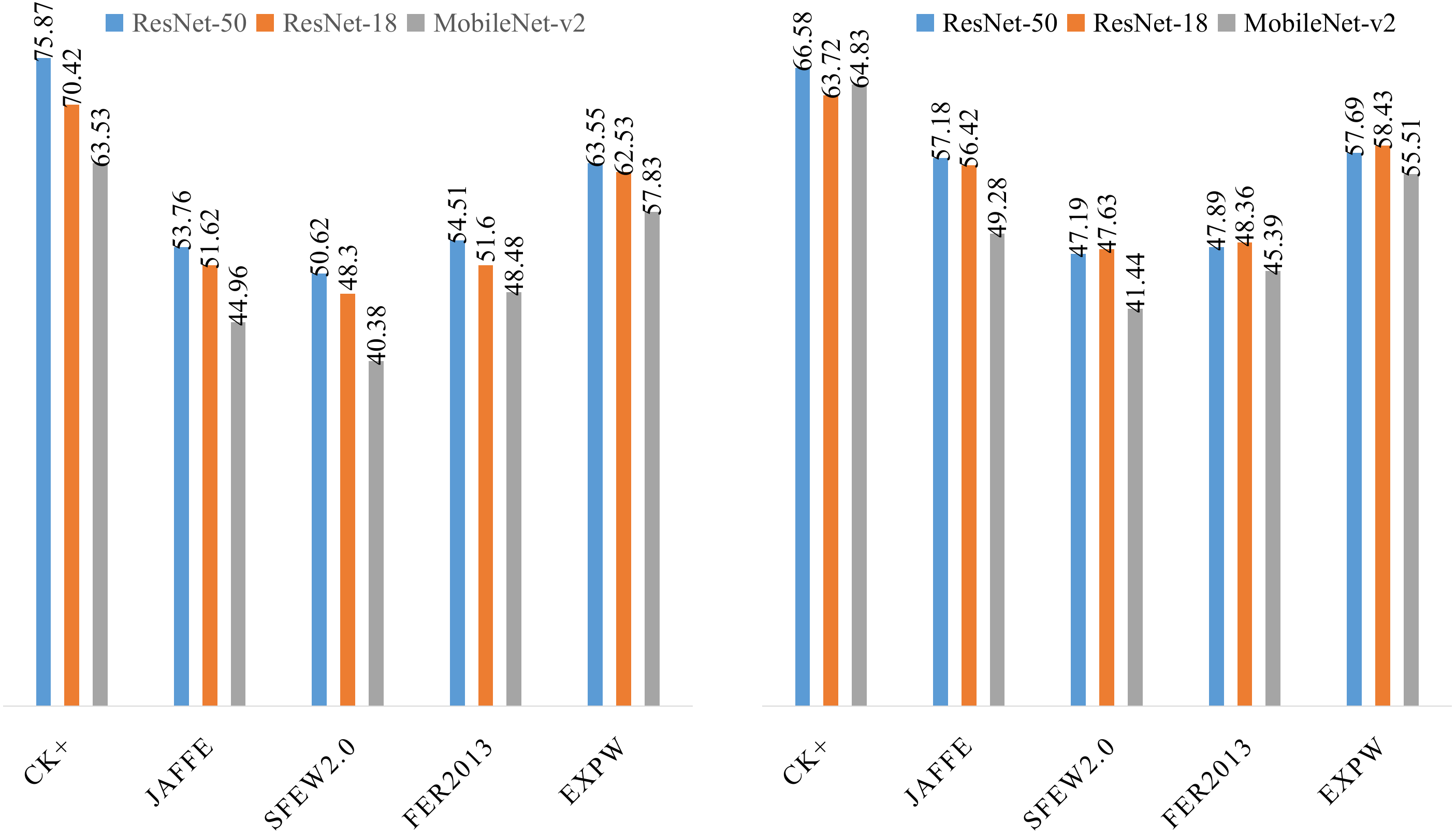}
\caption{The average accuracy comparisons using ResNet-50, ResNet-18, and MobileNet-v2 as the backbone. We average the accuracies of all the methods using the RAF-DB (left) and AFE (right) datasets as the source datasets.}
   \label{fig:backbone-comparison}
\end{figure}


\noindent\textbf{Effect of dataset inconsistency: }The source dataset provides basic supervision for recognition, and the distribution similarity with the target dataset is key for the final performance. To compare the performance using different source datasets, we compute the average value over all the methods that use the same backbone and test on the same target dataset, as shown in Figure \ref{fig:source-comparison}. We find that using the AFE dataset as the source dataset performs better on the JAFFE dataset, while using the RAF dataset performs well on the remaining CK+, SFEW2.0, FER2013, and ExpW datasets. One possible reason for this phenomenon is that the JAFFE dataset is collected from Japanese people, and its distribution is more similar to that of the AFE dataset. In contrast, the remaining datasets are mainly captured from Western areas, and their distributions are more similar to those of the RAF dataset. On the other hand, when using the same source dataset and the same backbone, the performances of the  different target datasets are also different. For example, the accuracies of the SWD vary from 52.06\% to 75.19\% if using the RAF-DB source dataset and the ResNet-50 backbone. This phenomenon is natural because different target datasets have different difficulties and different similarities with the source dataset.

\noindent\textbf{Effect of feature extractor inconsistency: }Feature extractors with different backbones can learn features that have inherently different discrimination and generalization abilities. Similarly, to compare the performance using different backbones, we compute the average value over all the methods that use the same source and target datasets. The results are presented in Figure \ref{fig:backbone-comparison}. We find that the performances decrease with the backbone choice from ResNet-50 to ResNet-18 to MobileNet-v2 because their discrimination and generalization abilities successively weaken.

\subsubsection{Comparison of the AGRA with current state-of-the-art algorithms}
In this part, we use the unified evaluation benchmark to compare the proposed AGRA approach with the current competing methods. As shown in Table \ref{table:fair-evaluation-results}, the proposed AGRA method consistently outperforms all the current methods for almost all the source/target datasets and backbones. Specifically, when using the RAF source dataset and ResNet-18 backbone, our AGRA approach obtains accuracies of 77.52\%, 61.03\%, 52.75\%, 54.94\%, 69.70\% on the CK+, JAFFE, SFEW2.0, FER2013, ExpW datasets, outperforming all of the current best-performing methods. For a comprehensive comparison, we average the accuracies of all the target datasets to obtain the mean accuracy for each source dataset and backbone network choice. As shown, our AGRA approach achieves the best mean accuracies for all the RAF-DB/ResNet-50, RAF-DB/ResNet-18, RAF-DB/MobileNet-v2, AFE/ResNet-50, AFE/ResNet-18, and AFE/MobileNet-v2 choices.

\subsection{Ablation Study}
In this subsection, we conduct ablation studies to discuss and analyze the actual contribution of each component and obtain a more thorough understanding of the framework. To ensure a fair comparison and evaluation, the experiments are conducted with the same ResNet-50 model as the backbone and RAF-DB dataset as the source domain. We eliminate the backbone and source dataset information for convenient illustration.

\begin{table}[htp]
\centering
\footnotesize
\begin{tabular}{p{1.4cm}|p{0.5cm}p{0.5cm}p{0.75cm}p{0.75cm}p{0.6cm}p{0.6cm}}
\hline
\centering  Method & CK+ & JAFFE & SFEW2.0 &  FER2013 &  ExpW & Mean\\
\hline
\hline
\centering Ours HFs  & 72.09 & 52.11 & 53.44 & 57.61 & 63.15 & 59.68\\
\centering Ours HLFs & 72.09 & 56.34 & 50.23 & 57.30 & 64.00 & 59.99\\
\centering Ours  & \textbf{85.27} & \textbf{61.50} & \textbf{56.43} & \textbf{58.95}  & \textbf{68.50} & \textbf{66.13} \\
\hline
\end{tabular}
\vspace{2pt}
\caption{Accuracies of our approach using holistic features (HFs), concatenating holistic-local features (HLFs) and ours for adaptation on the CK+, JAFFE, SFEW2.0, FER2013, and ExpW datasets.}
\label{table:result-hlf}
\end{table}

\begin{figure*}[!t]
\centering
{
\includegraphics[width=0.95\linewidth]{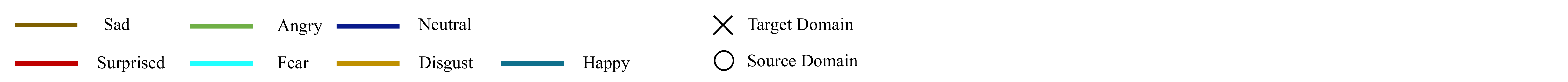}}
{
\label{fig:subfig1_1} 
\includegraphics[width=0.18\linewidth]{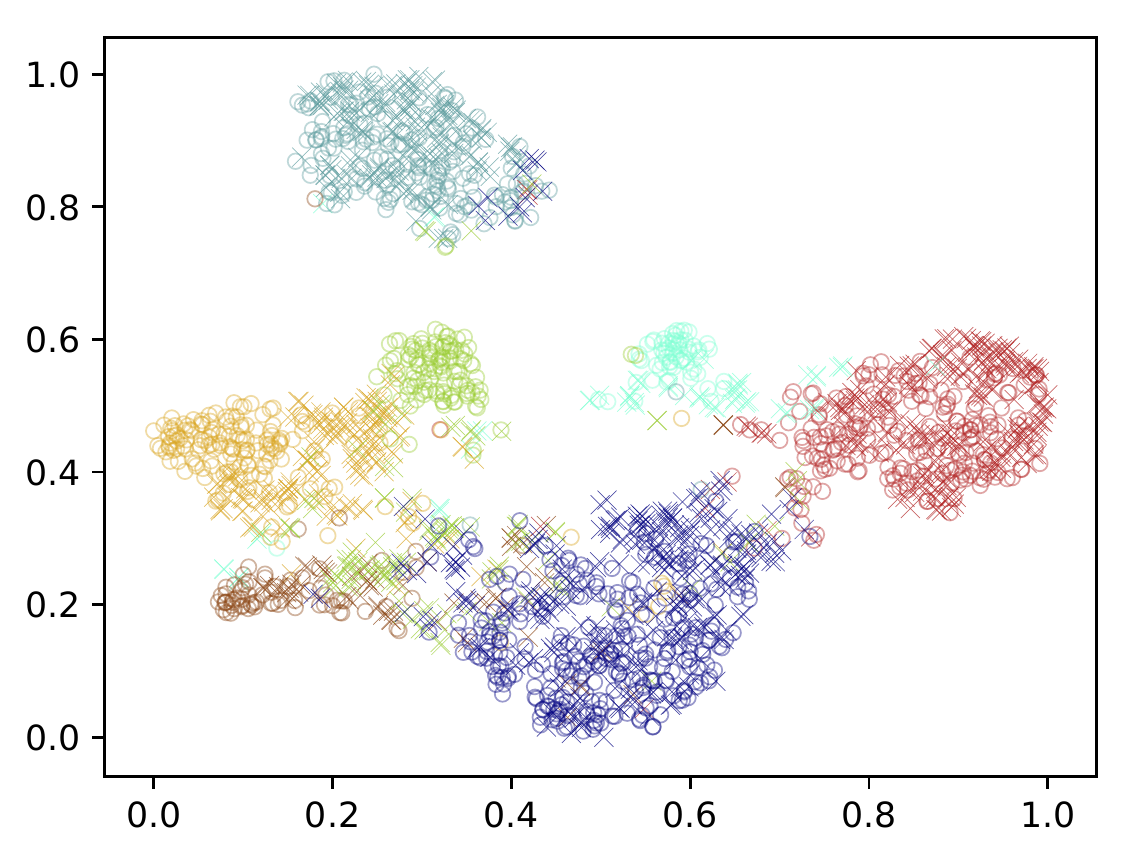}}
{
\label{fig:subfig1_2} 
\includegraphics[width=0.18\linewidth]{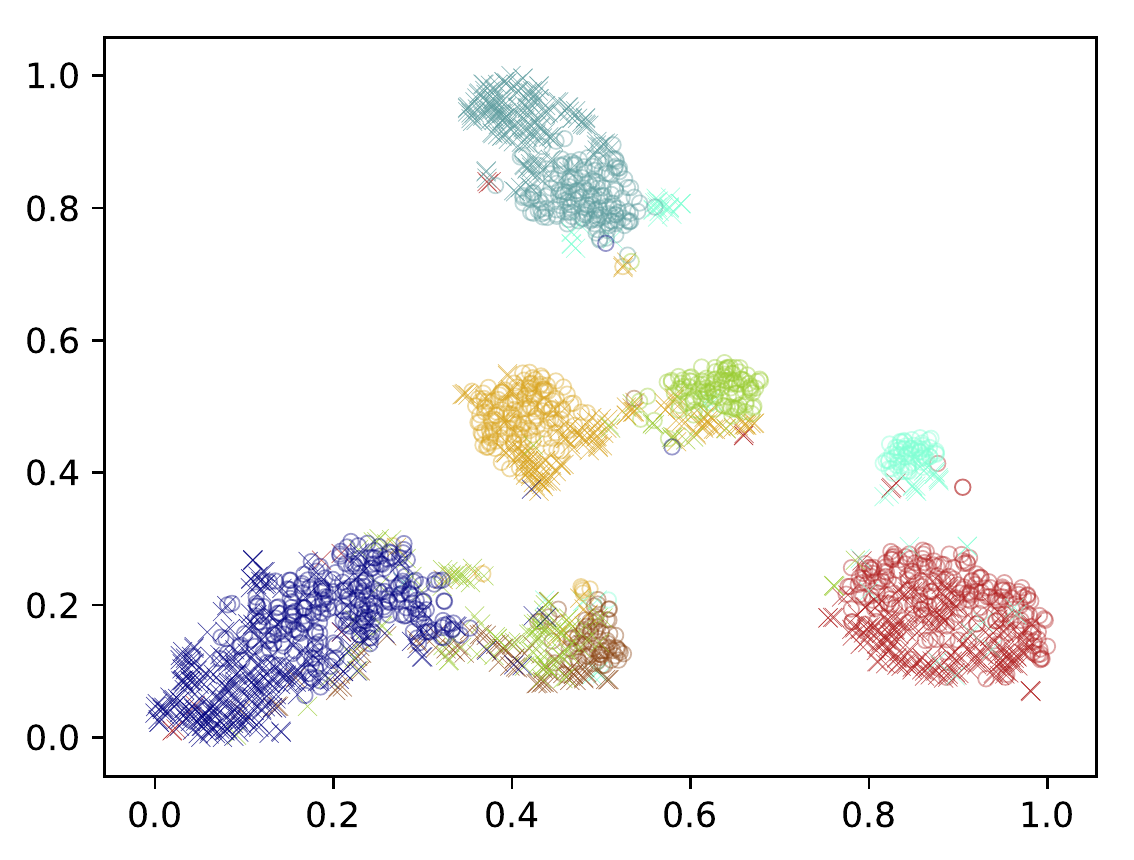}}
{
\label{fig:subfig1_3} 
\includegraphics[width=0.18\linewidth]{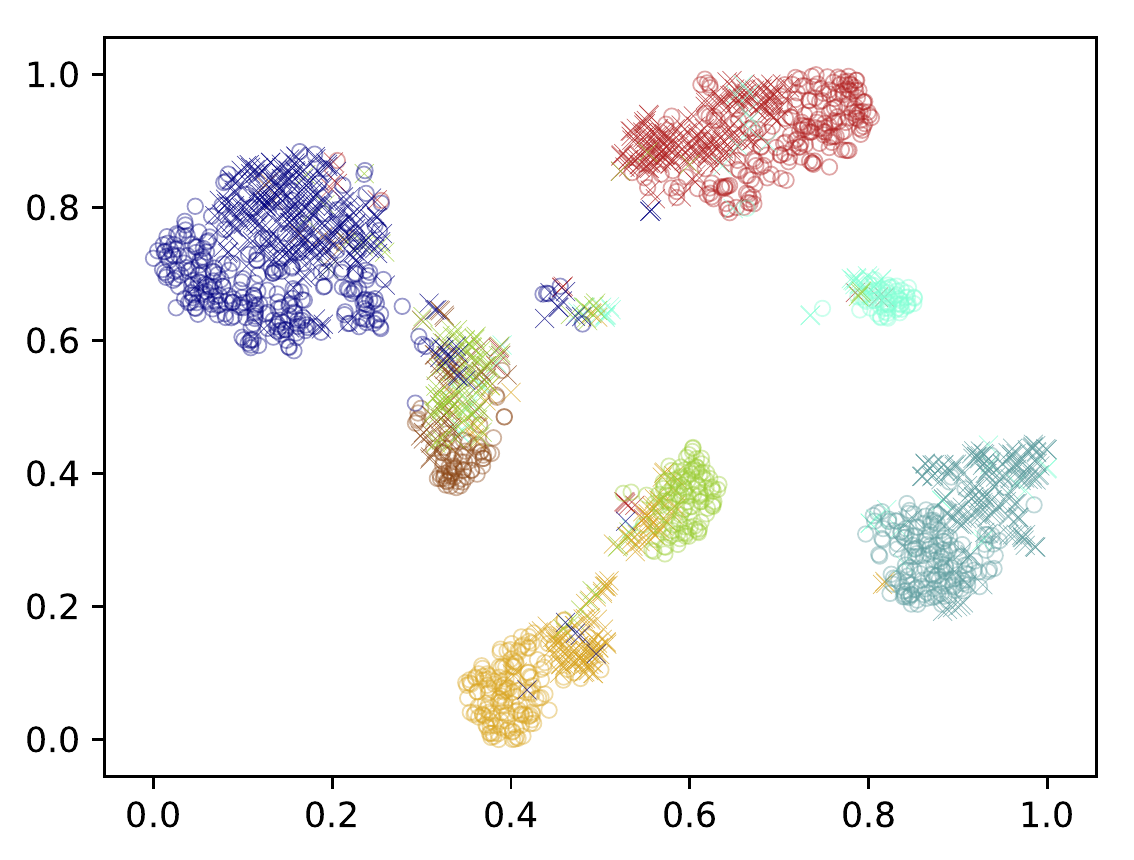}}
{
\label{fig:subfig1_4} 
\includegraphics[width=0.18\linewidth]{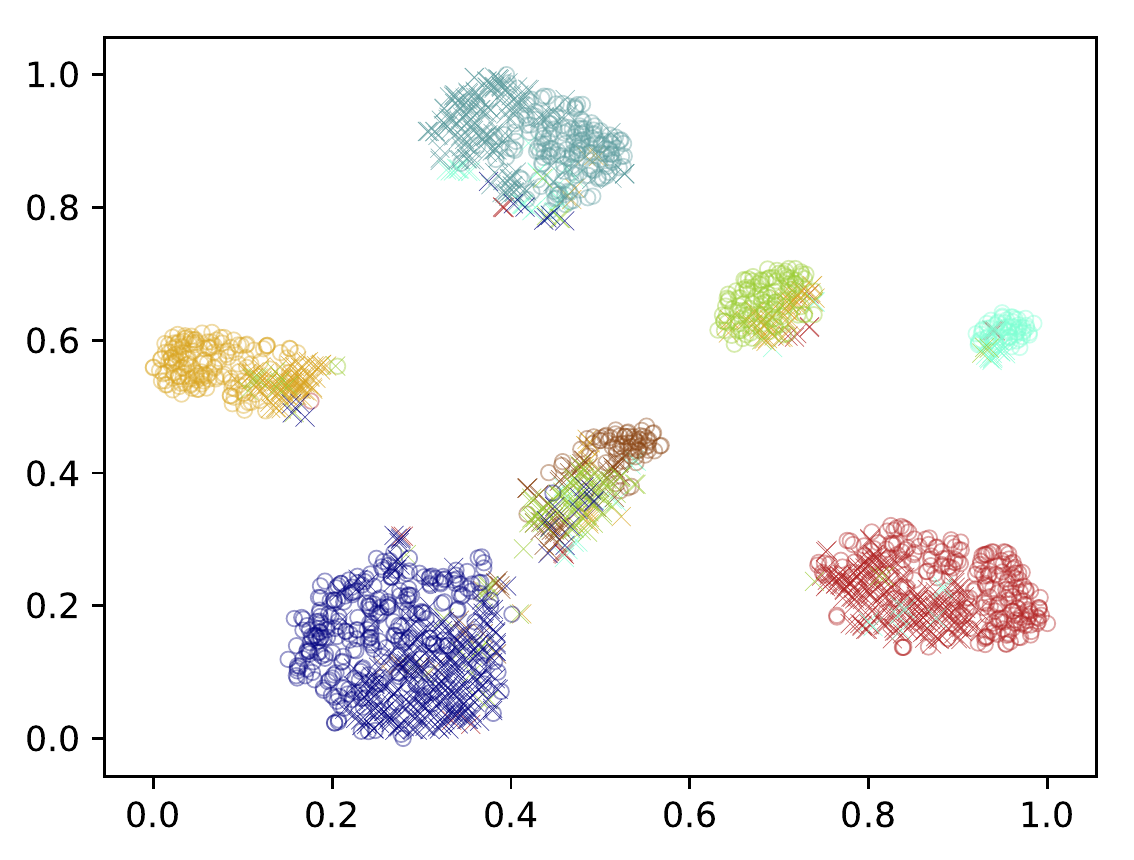}}
{
\label{fig:subfig1_5} 
\includegraphics[width=0.18\linewidth]{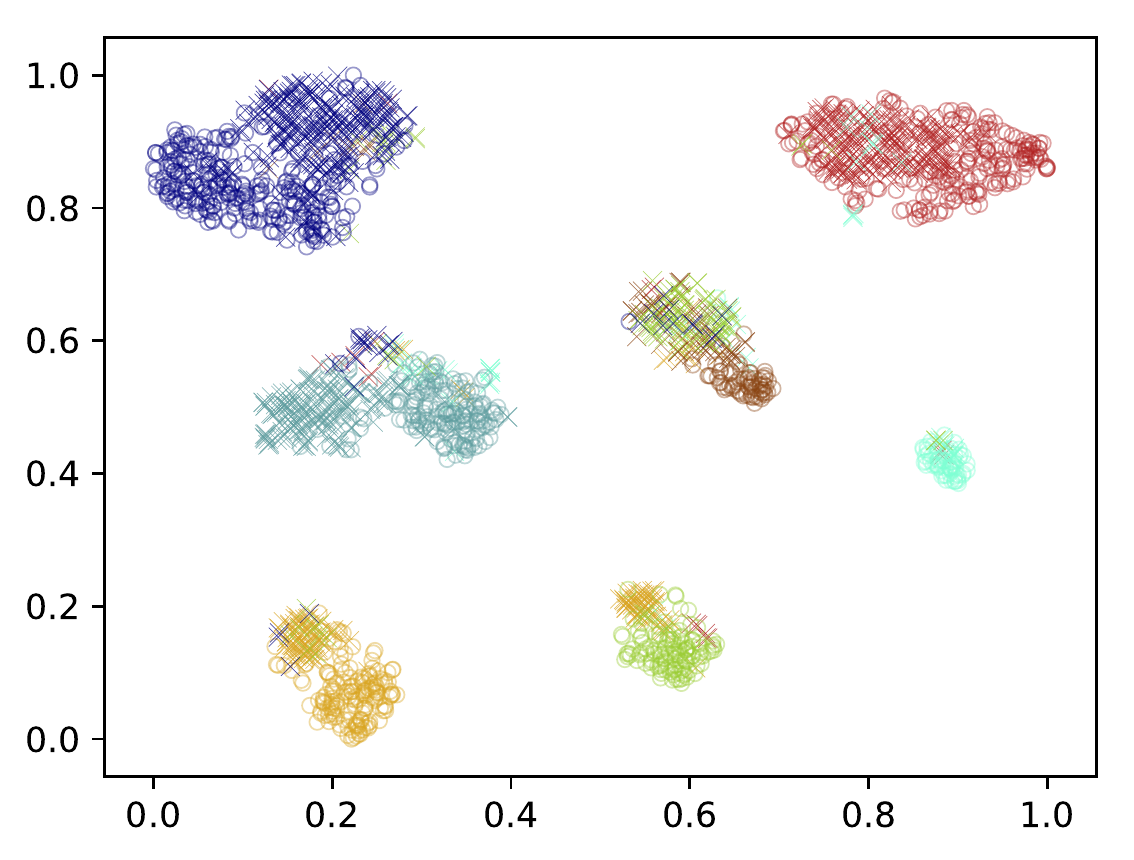}}
{
\label{fig:subfig2_1} 
\includegraphics[width=0.18\linewidth]{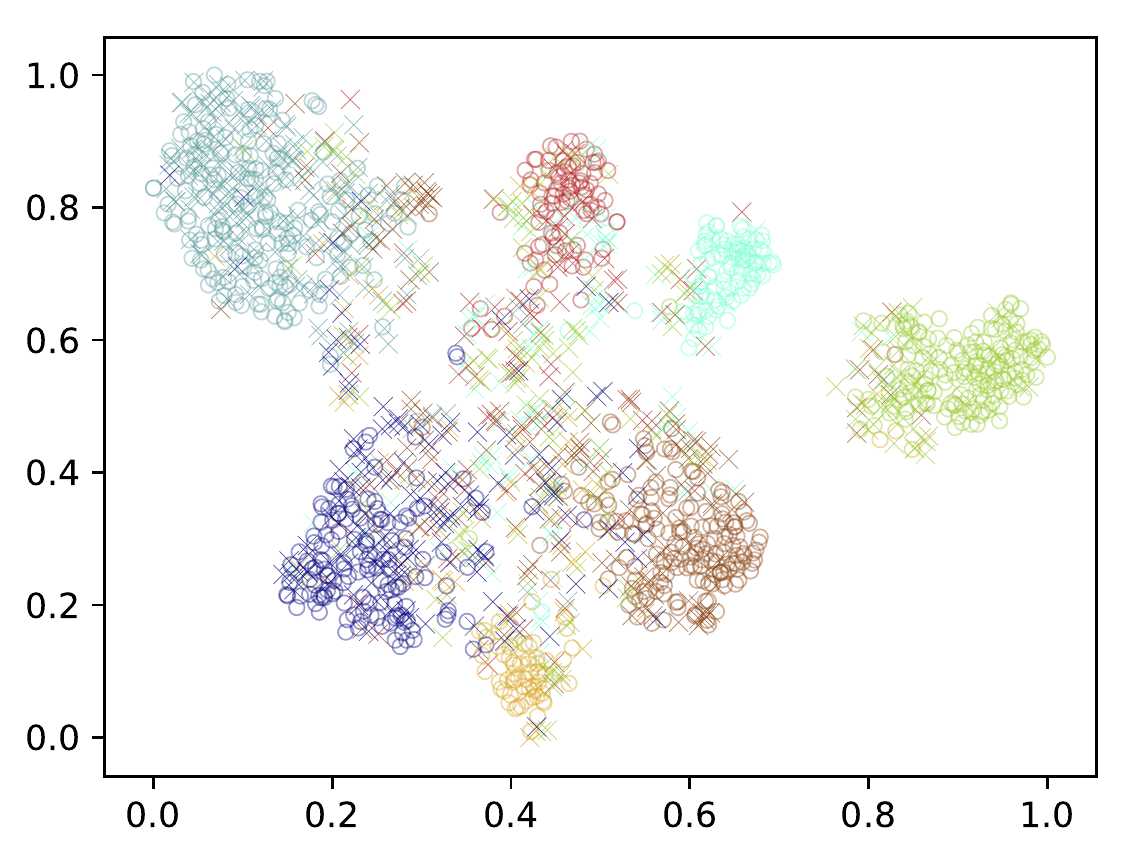}}
{
\label{fig:subfig2_2} 
\includegraphics[width=0.18\linewidth]{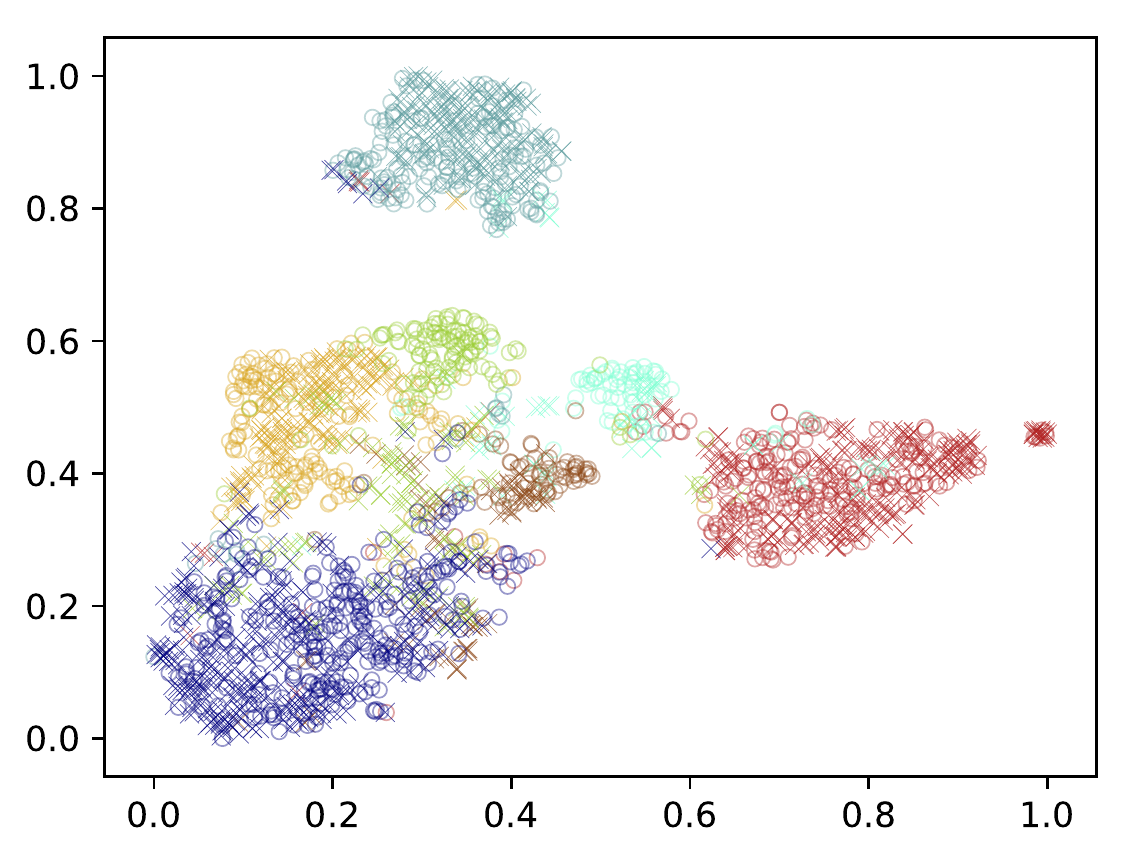}}
{
\label{fig:subfig2_3} 
\includegraphics[width=0.18\linewidth]{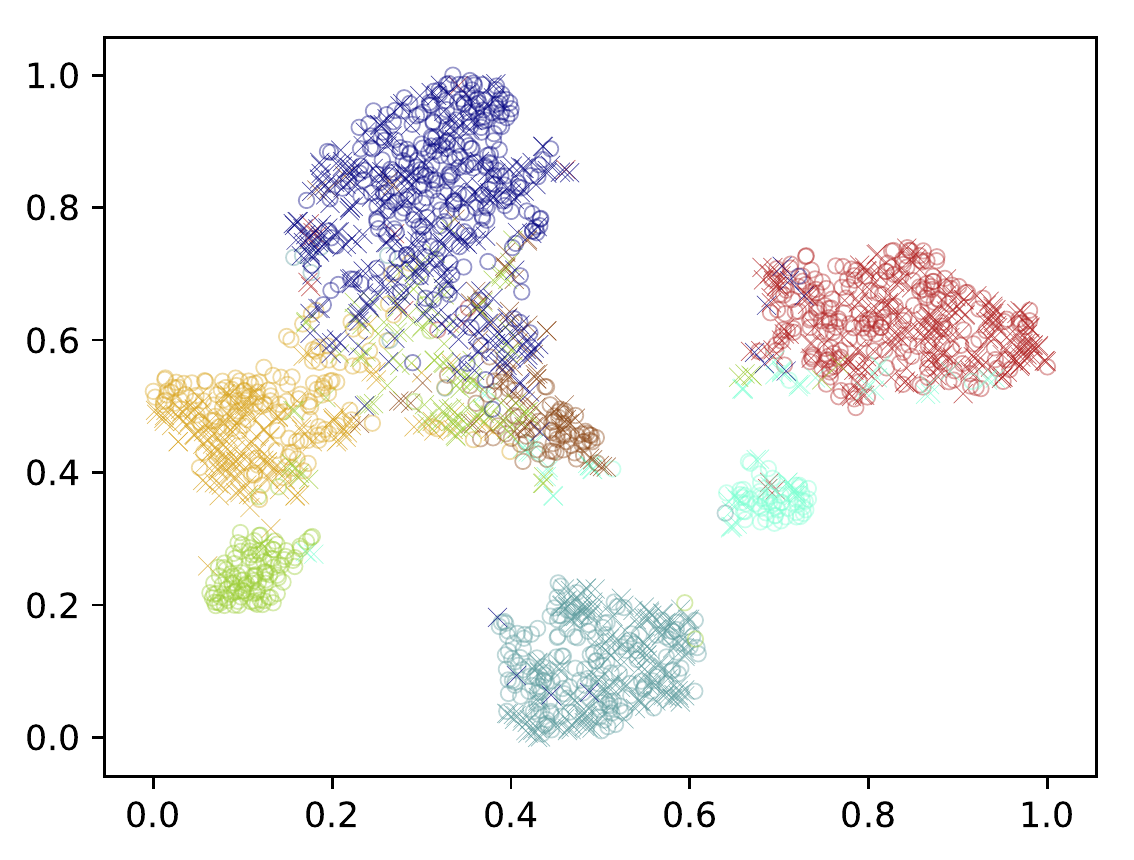}}
{
\label{fig:subfig2_4} 
\includegraphics[width=0.18\linewidth]{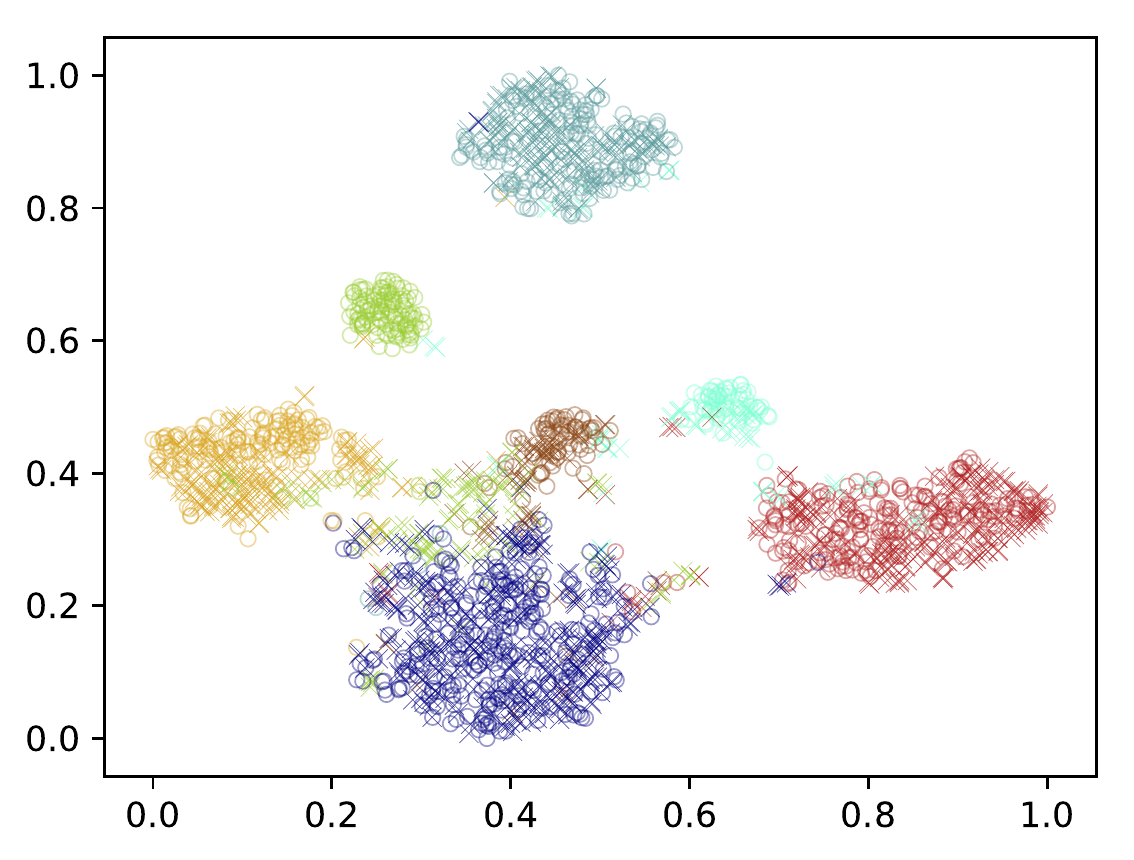}}
{
\label{fig:subfig2_5} 
\includegraphics[width=0.18\linewidth]{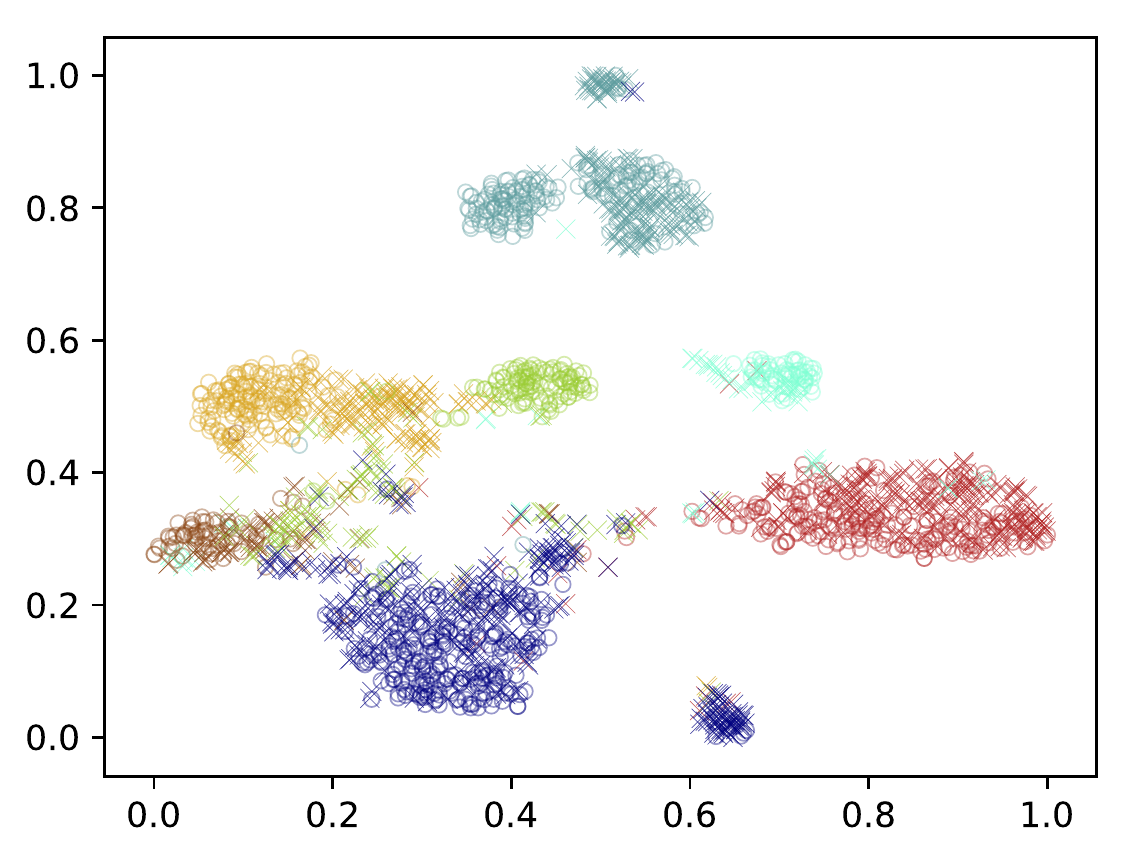}}
\caption{Illustration of the feature distributions learned by our proposed approach at epochs 0, 5, 10, 15, and 20 (from left to right) on the CK+ (upper) and SFEW2.0 (lower) datasets.}
\label{fig:visualization}
\end{figure*}

\subsubsection{Analysis of holistic-local feature co-adaptation}
\label{sec:hlf}
The core contribution of the proposed framework is the holistic-local feature co-adaptation module that jointly learns domain-invariant holistic-local features. To analyze its contribution, we remove this module while keeping the others unchanged. Thus, it merely uses holistic features for adaptation (namely, Ours HF). As shown in Table \ref{table:result-hlf}, removing this module leads to an obvious performance drop on all the datasets. Specifically, the accuracies drop from 85.27\% to 72.09\%, from 61.50\% to 52.11\%, from 56.43\% to 53.44\%, from 58.95\% to 57.61\%, and from 68.50\% to 63.15\% on the five datasets, respectively. The mean accuracy drops from 66.13\% to 59.68\%, with a decrease of 6.45\%. 

These obvious performance drops well demonstrate the contribution of the co-adaptation module for CD-FER well. It is also crucial that we introduce the two stacked GCNs for holistic-local feature co-adaptations. To verify their contribution, we remove the two GCNs and simply concatenate the holistic-local features for adaptation (namely, Ours HLFs). The results are presented in Table \ref{table:result-hlf}. We find that concatenating the local features can improve the performance, e.g., an improvement of 0.31\% in the mean accuracy. However, this approach still is outperformed by our AGRA approach on all five datasets, reducing the mean accuracy by 6.14\%.

\begin{table}[htp]
\centering
\footnotesize
\begin{tabular}{p{2.2cm}|p{0.5cm}p{0.5cm}p{0.75cm}p{0.75cm}p{0.6cm}p{0.6cm}}
\hline
\centering  Method & CK+ & JAFFE & SFEW2.0 &  FER2013 &  ExpW & Mean\\
\hline
\hline
\centering Ours intra-GCN & 77.52 & \textbf{61.97} & 55.28 & 57.95 & 66.99  & 63.94\\
\centering Ours inter-GCN & 77.52 & 57.75 & 49.77 & 55.64 & 66.00 & 61.34\\
\centering Ours single GCN & 74.42 & 56.34 & 52.06 & 57.33 & 67.30 & 61.49\\
\centering Ours  & \textbf{85.27} & 61.50 & \textbf{56.43} & \textbf{58.95}  & \textbf{68.50} & \textbf{66.13} \\
\hline
\end{tabular}
\vspace{2pt}
\caption{Accuracies of our approach using only the intra-domain GCN (Ours intra-GCN), using only the inter-domain GCN (Ours inter-GCN), using only one GCN (Ours single GCN), and our original approach (Ours) on the CK+, JAFFE, SFEW2.0, FER2013, and ExpW datasets.}
\label{table:result-gcn}
\end{table}

Note that we use two stacked GCNs, in which an intra-domain GCN propagates messages within each domain to capture holistic-local feature interactions and an inter-domain GCN transfers messages across different domains to ensure domain adaptation. To demonstrate the effectiveness of this point, we conduct an experiment that uses one single GCN for message propagation within and across the source and target domains. As shown in Table \ref{table:result-gcn}, we find dramatic performance drops on all the datasets, e.g., a decrease in the mean accuracy by 4.64\%. The reason is mainly because message propagations within each domain and across different domains are different, and using only one GCN cannot model two types of propagation well. To further analyze the actual contribution of each GCN, we conduct two more experiments. The first experiment removes the inter-domain GCN and merely performs message propagation within each domain, while the second experiment removes the intra-domain GCN, and message propagation is only carried out across different domains. We find that both experiments show obvious performance drops, i.e., a decrease in the mean accuracy by 2.19\% if the inter-domain GCN is removed and a decrease in the mean accuracy by 4.79\% if the intra-domain GCN is removed, as shown in Table \ref{table:result-gcn}.

\begin{table}[htp]
\centering
\footnotesize
\begin{tabular}{p{1.6cm}|p{0.5cm}p{0.5cm}p{0.75cm}p{0.75cm}p{0.6cm}p{0.6cm}}
\hline
\centering  Method & CK+ & JAFFE & SFEW2.0 &  FER2013 &  ExpW & Mean\\
\hline
\hline
\centering Ours mean & 82.95 & 52.58 & 55.96 & 58.45 & 65.23 & 63.03 \\
\centering Ours iter  & 82.17 & 58.28 & 52.98 & 56.40 & 68.32 & 63.63 \\
\centering Ours epoch & 80.62 & 56.81 & 53.67 & 55.58 & 66.59 & 62.65 \\
\centering Ours  & \textbf{85.27} & \textbf{61.50} & \textbf{56.43} & \textbf{58.95}  & \textbf{68.50} & \textbf{66.13} \\
\hline
\end{tabular}
\vspace{2pt}
\caption{Accuracies of our approach with the mean statistical distribution (Ours mean), our approach and updating the per-class statistical distributions every iteration (Ours iter), our approach and updating the per-class statistical distributions every ten epochs (Ours epoch) and our original approach (Ours) on the CK+, JAFFE, SFEW2.0, FER2013, and ExpW datasets.}
\label{table:result-statistical-distribution}
\end{table}

\subsubsection{Analysis of the per-class statistical distributions}
To ensure meaningful initializations for the nodes of each domain when the input image comes from the other domain, we learn the per-class statistical feature distributions. Here, we first illustrate the feature distributions of samples from the lab-controlled CK+ and in-the-wild SFEW2.0 datasets during different training stages. As shown in Figure \ref{fig:visualization}, it can be observed that the proposed model can gather the samples of the same category and from different domains together, which suggests that it can learn discriminative and domain-variant features. To quantitatively analyze its contribution, we learn the dataset-level statistical feature distributions and replace the per-class statistical feature distributions for node initialization. We find that the mean accuracy drops from 66.13\% to 63.03\%, as shown in Table \ref{table:result-statistical-distribution}.

As stated above, we learn the per-class statistical distributions by updating every iteration and reclustering every ten epochs. To analyze the effect of the updating mechanism, we conduct experiments that merely update every iteration or merely recluster every ten epochs and present the results in Table \ref{table:result-statistical-distribution}. We find that both experiments exhibit an obvious drop in performance; i.e., the mean accuracy decreases by 2.50\% if the per-class statistical distributions are updated every iteration and by 3.48\% if reclustering is performed every ten epochs.

\begin{table}[htp]
\centering
\footnotesize
\begin{tabular}{p{1.2cm}|p{0.5cm}p{0.5cm}p{0.75cm}p{0.75cm}p{0.6cm}p{0.6cm}}
\hline
\centering  Method & CK+ & JAFFE & SFEW2.0 &  FER2013 &  ExpW & Mean\\
\hline
\hline
\centering Ours RM & 68.99 & 50.70 & 54.36 & 55.47 & 67.88 & 59.48 \\
\centering Ours OM & 79.07 & 57.28 & 53.90 & 57.07 & 66.71 & 62.81 \\
\centering Ours FM & 68.99 & 47.42 & 54.13 & 53.28 & 56.25 & 56.01 \\
\centering Ours  & \textbf{85.27} & \textbf{61.50} & \textbf{56.43} & \textbf{58.95}  & \textbf{68.50} & \textbf{66.13} \\
\hline
\end{tabular}
\vspace{2pt}
\caption{Accuracies of our approach where the matrices are initialized with randomly initialized matrices (Ours RM), our approach where the matrices are initialized with all-one matrices (Ours OM),  our approach where the matrices are initialized with fixed matrices (Ours FM), and our original approach (Ours) on the CK+, JAFFE, SFEW2.0, FER2013, and ExpW datasets.}
\label{table:result-am}
\end{table}

\subsubsection{Analysis of the adjacency matrix}
We initialize the two adjacency matrices of the intra-domain and inter-domain graphs with manually defined connections, which can provide prior guidance to regularize message propagation. In this part, we replace the adjacency matrices with two randomly initialized matrices (denoted by Ours RM) and with two all-one matrices (denoted by Ours OM) to verify the effectiveness of this point. We present the results in Table \ref{table:result-am}. We observe that both experiments show severe performance degradation on all the  datasets; i.e., the mean accuracies are degraded by 6.65\% and 3.32\%, respectively. It is noteworthy that the experiment with randomly initialized matrices exhibits more obvious performance degradation than the experiment with the all-ones matrices. One possible reason is that the randomly initialized matrices may provide misleading guidance for message propagation, which further indicates the importance of the prior adjacency matrices.

To adjust the adjacency matrices to better guide message propagation, they are also jointly fine-tuned during the training process. In this part, we verify the effectiveness by fixing the prior matrix training. We present the results in Table \ref{table:result-am}. The mean accuracy drops from 66.13\% to 56.01\%, which suggests that jointly adjusting the adjacency matrices can learn dataset-specified matrices, which is crucial to promoting CD-FER.

\begin{table}[htp]
\centering
\footnotesize
\begin{tabular}{p{0.6cm}p{0.6cm}|p{0.5cm}p{0.5cm}p{0.75cm}p{0.75cm}p{0.7cm}p{0.7cm}}
\hline
\centering  $T_{intra}$ & $T_{inter}$ & CK+ & JAFFE & SFEW2.0 &  FER2013 & ExpW & Mean\\
\hline
\hline
\centering  1 & 1 & 75.19 & 52.11 & 55.28 & 57.22 & 67.32 & 61.42 \\
\centering  2 & 1 & \textbf{85.27} & \textbf{61.50} & \textbf{56.43} & \textbf{58.95}  & \textbf{68.50} & \textbf{66.13} \\
\centering  3 & 1 & 80.62 & 53.06 & 50.46 & 56.82 & 64.41 & 61.07\\
\hline
\hline
\centering  2 & 2 & 74.42 & 54.46 & 54.59 & 58.31 & 66.94 & 61.74 \\
\centering  2 & 3 & 79.07 & 49.77 & 51.61 & 56.85 & 67.14 & 60.89\\
\hline
\end{tabular}
\vspace{2pt}
\caption{Accuracies of our approach with different numbers of iterations for the intra-domain GCN and inter-domain GCN on the CK+, JAFFE, SFEW2.0, FER2013, and ExpW datasets.}
\label{table:result-gcn-layer}
\end{table}

\begin{table}[htp]
\centering
\footnotesize
\begin{tabular}{p{1.4cm}|p{0.5cm}p{0.5cm}p{0.75cm}p{0.75cm}p{0.6cm}p{0.6cm}}
\hline
\centering  Method & CK+ & JAFFE & SFEW2.0 &  FER2013 &  ExpW & Mean\\
\hline
\hline
\centering Ours TR & 75.19 & 51.64 & 54.82 & 57.33 & 66.51 & 61.10 \\
\centering Ours NT & 74.42 & 59.62 & 53.21 & 56.93 & 67.26 & 62.29 \\
\centering Ours GCN & \textbf{85.27} & \textbf{61.50} & \textbf{56.43} & \textbf{58.95} & \textbf{68.50} & \textbf{66.13} \\
\hline
\end{tabular}
\vspace{2pt}
\caption{Accuracies of our approach that uses GCN-based (Ours GCN), neural tensor (Ours NT), and transformer-based (Ours TR) fusion algorithms on the CK+, JAFFE, SFEW2.0, FER2013, and ExpW datasets.}
\label{table:result-GCN}
\end{table}

\subsubsection{Analysis of GCN}
In this work, we use GCN for message propagation. Increasing the number of layers of the GCN can promote deeper feature interaction, but it may lead to message smoothing and hurt the discriminative ability. Here, we present experimental studies to analyze the effect of the number of iterations (i.e., $T_{intra}$ and $T_{inter}$) of both GCNs on CD-FER. To this end, we first fix $T_{inter}$ as 1 and vary $T_{intra}$ from 1 to 3. As shown in Table \ref{table:result-gcn-layer}, the performance can be boosted by increasing $T_{intra}$ from 1 to 2, but the performance drops when further increasing it to 3. Thus, we set the number of layers of the intra-domain GCN to 2 and conduct an experiment that varies $T_{inter}$ from 1 to 3. We find that setting $T_2$ to 1 achieves the best performance and increasing this number results in performance degradation, as depicted in Table \ref{table:result-gcn-layer}. Thus, we set $T_{intra}$ to 2 and $T_{inter}$ to 1 for all the experiments.

On the other hand, other message propagation and fusion algorithms exist, such as neural tensor and transformer-based fusion. To verify the effectiveness of the GCN-based message propagation and fusion, we further implement two more baselines that use the neural tensor and transformer-based fusion algorithms, respectively. 1) For the neural tensor fusion based algorithm, we replace the two GCNs with two neural tensor fusion operations, in which an intra-domain neural tensor fusion operation is used to fuse the holistic and local features of the source and target domain respectively, and then an inter-domain neural tensor fusion operation is used to fuse features from both domains. 2) For the transformer-based algorithm, we similarly replace the two GCNs with two transformers, in which an intra-domain transformer is used to sequentially fuse the holistic and local features of the source and target domain respectively, and then an inter-domain transformer is used to sequentially fuse features from both domains. As shown in Table \ref{table:result-GCN}, the results obtained by using GCN-based fusion algorithm obviously outperforms those obtained by using these two algorithms.

\subsubsection{Contribution of adversarial learning}
Adversarial learning is also key to facilitating learning domain-invariant features. To validate its contribution, we remove the adversarial loss and use only the classification loss to train the whole model. As shown in Table \ref{table:result-al}, the accuracies drop to 68.22\%, 49.30\%, 52.98\%, 56.46\% and 63.93\% on the five datasets, much worse than that using adversarial learning.

\begin{table}[htp]
\centering
\footnotesize
\begin{tabular}{p{1.8cm}|p{0.5cm}p{0.5cm}p{0.75cm}p{0.75cm}p{0.6cm}p{0.6cm}}
\hline
\centering  Method & CK+ & JAFFE & SFEW2.0 &  FER2013 &  ExpW & Mean\\
\hline
\hline
\centering Ours w/o AL & 68.22 & 49.30 & 52.98 & 56.46 & 63.93 & 58.18 \\
\centering Ours w/ AL  & \textbf{85.27} & \textbf{61.50} & \textbf{56.43} & \textbf{58.95}  & \textbf{68.50} & \textbf{66.13} \\
\hline
\end{tabular}
\vspace{2pt}
\caption{Accuracies of our approach with and without adversarial learning (Ours w/ AL and Ours w/o AL, respectively) on the CK+, JAFFE, SFEW2.0, FER2013, and ExpW datasets.}
\label{table:result-al}
\end{table}

\subsubsection{Efficiency analysis}
During inference stage, the proposed framework introduces two graphs for within-domain and cross-domain message propagation. Here, we further compare the computational overhead. As suggested in Sec. \ref{Network architecture}, the intra-domain GCN consists of two layers with dimensions of $64\times 128$ and $128\times 64$, while the inter-domain GCN contains merely one layer with dimensions of $64\times 64$, which introduces negligible computational overhead. As shown in Table \ref{table:efficieny-result}, the baseline with ResNet-50 backbone has 6.7700G multiply-accumulate operations (MACs) and adding the two GCNs increases the MACs to 6.7702G, with an relative increase of 0.00295\%. For more practical analysis, we further compare the running time for our method with and without the GCNs. Specifically, the experiment is conducted with a batch size of 1 and on a desktop with a single NVIDIA GTX 1070 Ti. As shown, when using ResNet-50 as backbone, the proposed framework with the GCNs needs 2.54 ms more time than that without the GCNs.

For a comprehensive analysis on this point, we also analyze the training times of our method with and without the GCNs. Specifically, we use RAF-DB as the source dataset and CK+ as the target dataset and conduct the experiment on a desktop with a single NVIDIA GTX 1070 Ti. As described in Sec. \ref{Sec:TrainingDetail}, it consists of one training stage on the source dataset (first training stage, shorted as FTS) and another training stage on both the source and target datasets (second training stage, shorted as STS). And we present the training time for each epoch on these two stages in Table \ref{table:efficieny-result-2}. As shown, adding the GCNs increases about 5\% to 20\% training time for each epoch.

\begin{table}[h]
\centering
\scriptsize
\begin{tabular}{c|cccccc}
\hline
\centering  & \multicolumn{2}{c|}{ResNet-50} & \multicolumn{2}{c|}{ResNet-18} & \multicolumn{2}{c}{ MobileNet-v2} \\
\hline
\centering Method & MACs & Time & MACs & Time & MACs & Time\\
\hline
Our w/o GCNs & 6.7700 & 18.30 & 3.0665 & 15.88 & 2.9481 & 20.58 \\
Our w/ GCNs & 6.7702 & 20.84 & 3.0667 & 16.86 & 2.9483 & 21.87 \\
\hline
\end{tabular}
\vspace{2pt}
\caption{The multiply-accumulate operations (MACs) and running time of the proposed framework with and without the GCNs. The units are giga (G) for MACs and millisecond (ms) for running time.}
\label{table:efficieny-result}
\end{table}

\begin{table}[h]
\centering
\scriptsize
\begin{tabular}{c|cccccc}
\hline
\centering  & \multicolumn{2}{c|}{ResNet-50} & \multicolumn{2}{c|}{ResNet-18} & \multicolumn{2}{c}{ MobileNet-v2} \\
\hline
\centering Method & FTS & STS & FTS & STS & FTS & STS \\
\hline
Our w/o GCNs & 327	& 648 & 242 & 435 & 283 & 537\\
Our w/ GCNs & 386 & 713 & 277 & 468 & 334 & 579 \\
\hline
\end{tabular}
\vspace{2pt}
\caption{The training time (seconds) for each epoch of the proposed framework with and without the GCNs. We select RAF-DB and CK+ as the source and target datasets.}
\label{table:efficieny-result-2}
\end{table}

\section{Conclusion}
\label{sec:conclusion}
In this work, we first analyze the inconsistent choices of the source/target datasets and feature extractors and their performance effect on the CD-FER task. Then, we construct a unified evaluation CD-FER benchmark, in which all the competing methods are compared fairly with unified source/target datasets and feature extractors. A new AFE dataset is also built and added to the benchmark. In addition, based on the observation that current leading CD-FER methods mainly focus on learning holistic domain-invariant features but ignore local features that are more transferable and carry more detailed content, we develop a novel AGRA framework that integrates the graph propagation mechanism with adversarial learning for effective holistic-local representation co-adaptation across different domains. In the experiments, we use the unified evaluation benchmark to compare the proposed AGRA framework with current state-of-the-art methods, which demonstrates the effectiveness of the proposed framework.

\ifCLASSOPTIONcaptionsoff
\newpage
\fi

{
\bibliographystyle{IEEEtran}
\bibliography{reference}
}

\begin{IEEEbiography}[{\includegraphics[width=1in,height=1.25in,clip,keepaspectratio]{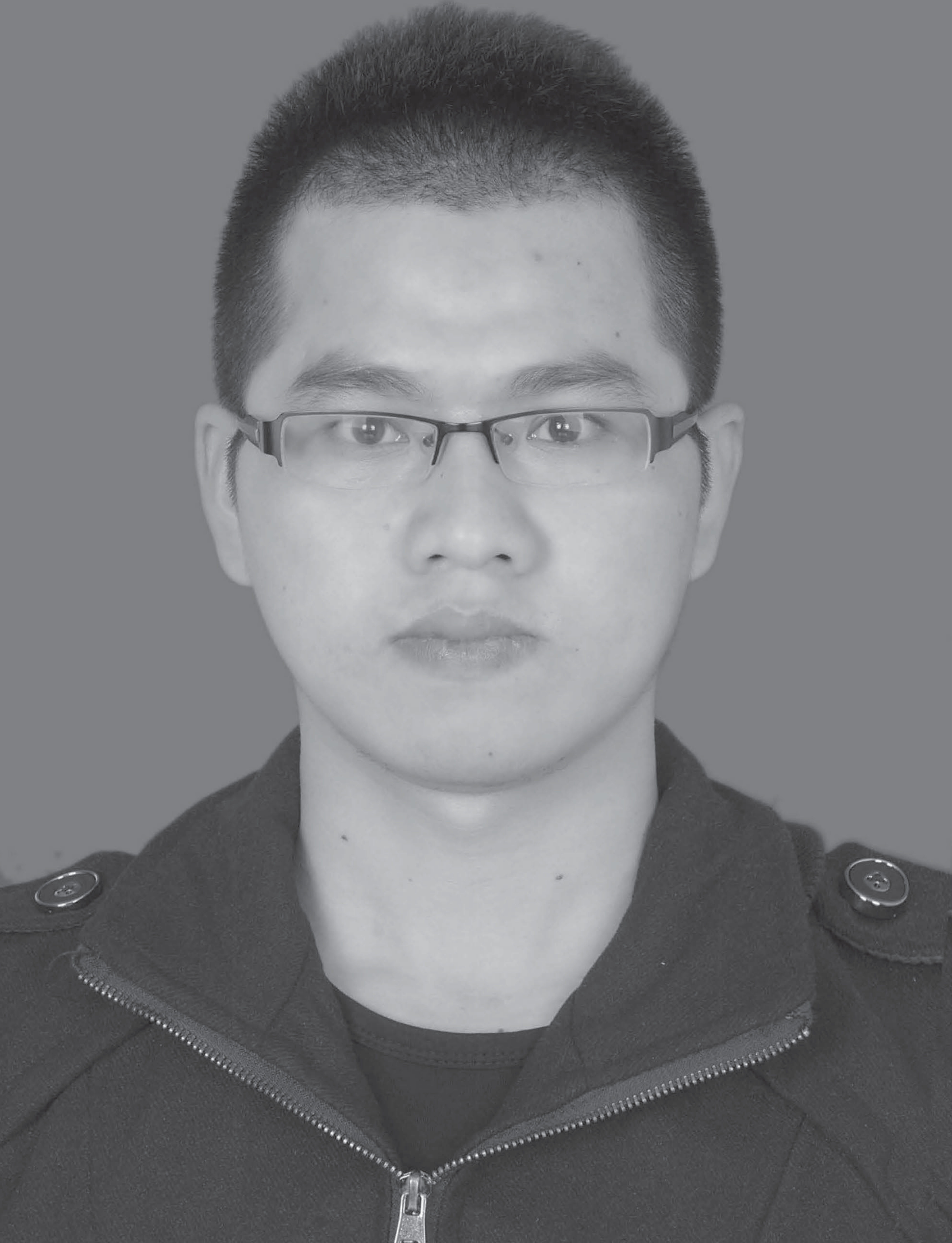}}]{Tianshui Chen} received a Ph.D. degree in computer science at the School of Data and Computer Science Sun Yat-sen University, Guangzhou, China, in 2018. Prior to earning his Ph.D, he received a B.E. degree from the School of Information and Science Technology in 2013. He is currently the lecturer in the Guangdong University of Technology. His current research interests include computer vision and machine learning. He has authored and coauthored approximately 30 papers published in top-tier academic journals and conferences, including T-PAMI, T-NNLS, T-IP, T-MM, CVPR, ICCV, AAAI, IJCAI, ACM MM, etc. He has served as a reviewer for numerous academic journals and conferences. He was the recipient of the Best Paper Diamond Award at IEEE ICME 2017.
\end{IEEEbiography}

\begin{IEEEbiography}[{\includegraphics[width=1in,height=1.25in,clip,keepaspectratio]{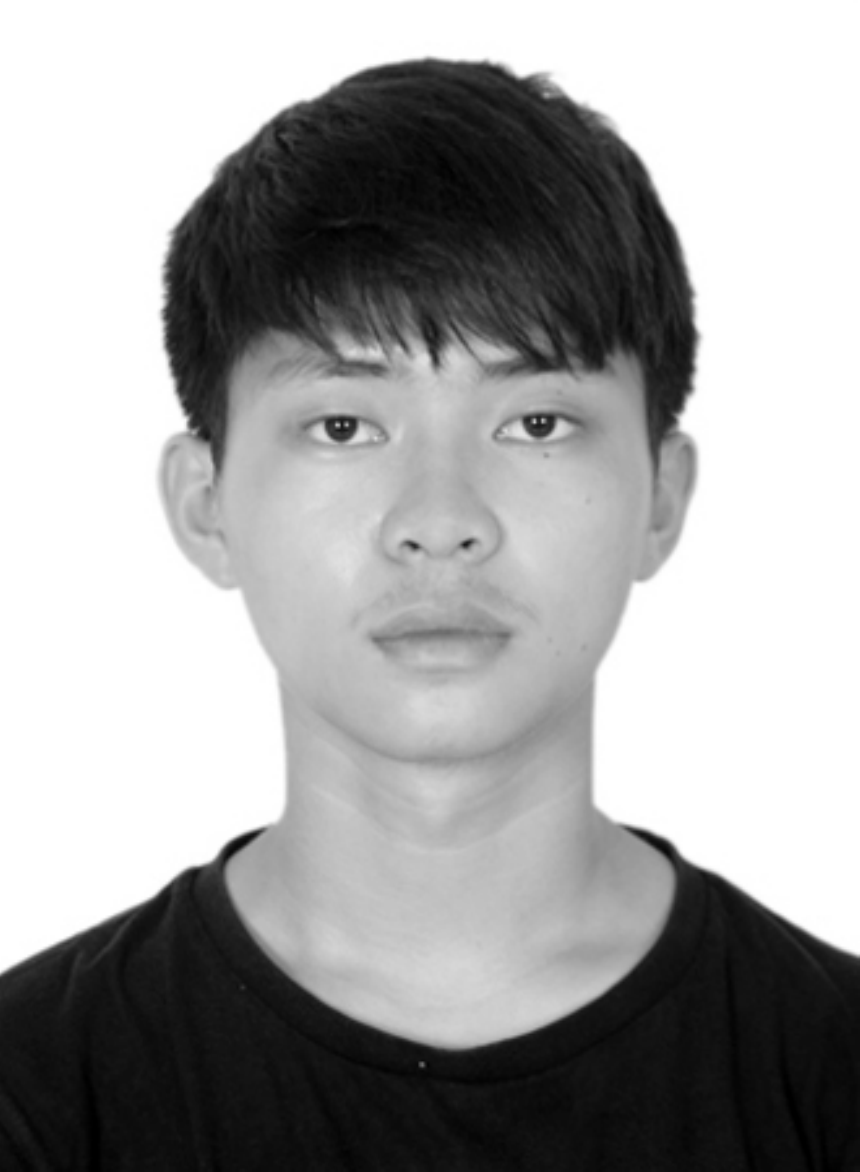}}]{Tao Pu}
received a B.E. degree from the School of Computer Science and Engineering, Sun Yat-sen University, Guangzhou, China, in 2020, where he is currently pursuing a master's degree in computer science. His current research interests include computer vision and machine learning.
\end{IEEEbiography}

\begin{IEEEbiography}[{\includegraphics[width=1in,height=1.25in,clip,keepaspectratio]{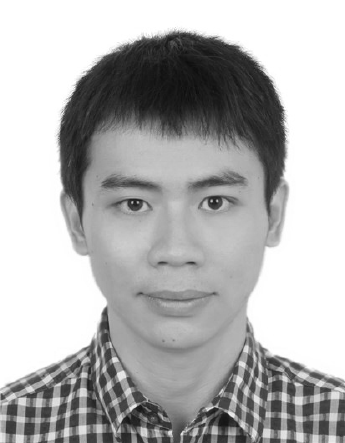}}]{Hefeng Wu} received a B.S. degree in computer science and technology and a Ph.D. degree in computer application technology from Sun Yat-sen University, China, in 2008 and 2013, respectively. He is currently a full research scientist at the School of Computer Science and Engineering, Sun Yat-sen University, China. His research interests include computer vision, multimedia, and machine learning. He has published works in and served as reviewers for many top-tier academic journals and conferences, including T-PAMI, T-IP, T-MM, CVPR, ICCV, AAAI, ACM MM, etc.
\end{IEEEbiography}

\begin{IEEEbiography}[{\includegraphics[width=1in,height=1.25in,clip,keepaspectratio]{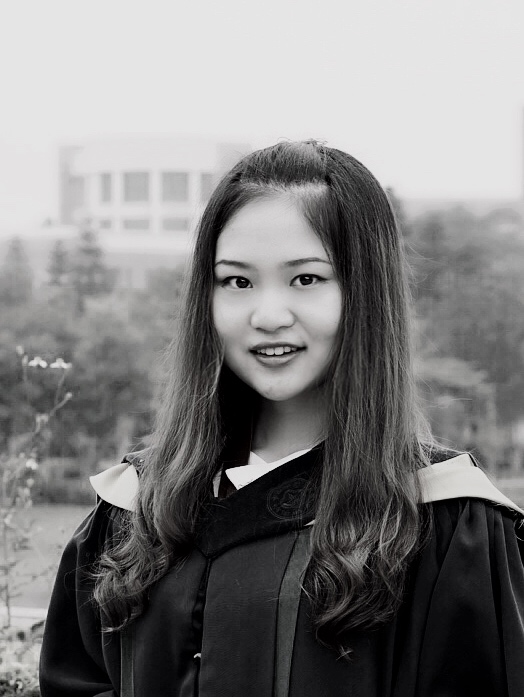}}]{Yuan Xie}
received a B.E. degree in software engineering and a master's degree in computer science and technology from the School of Data and Computer Science Sun Yat-sen University, Guangzhou, China, in 2016 and 2019, respectively. She is currently a senior researcher at DMAI. Her research interests are in computer vision and human behavior analysis and their applications to human behavior analysis, human–robot interaction, and personalized
learning.
\end{IEEEbiography}

\begin{IEEEbiography}[{\includegraphics[width=1in,height=1.25in,clip,keepaspectratio]{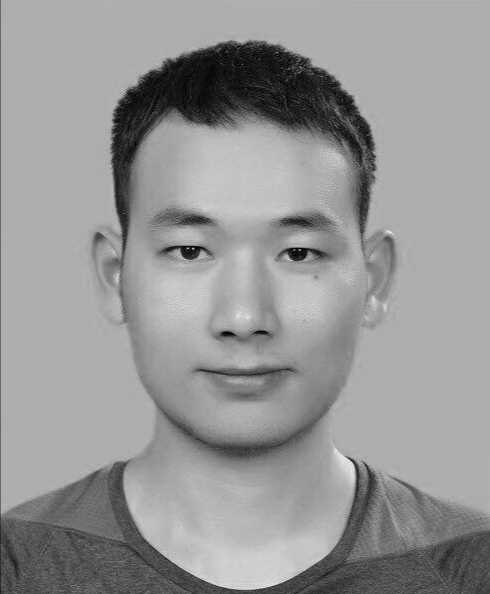}}]{Lingbo Liu} is currently a postdoctoral fellow in the Department of Computing at the Hong Kong Polytechnic University. He received his Ph.D. degree in computer science from Sun Yat-Sen University in 2020. He was a research assistant at the University of Sydney, Australia. His current research interests include machine learning and intelligent transportation systems. He has authored and coauthored more than 15 papers in top-tier academic journals and conferences.
\end{IEEEbiography}

\begin{IEEEbiography}[{\includegraphics[width=1in,clip]{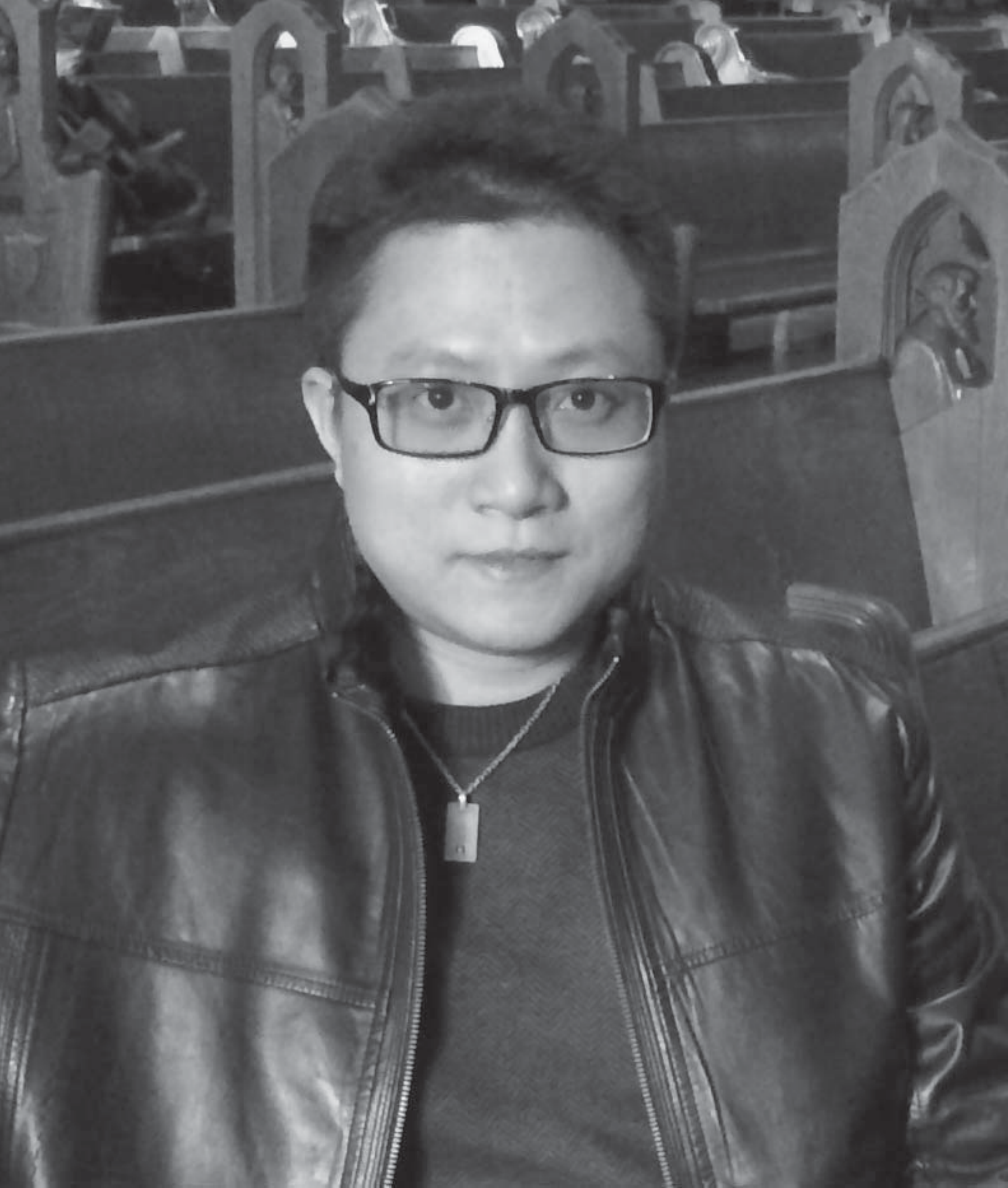}}]{Liang Lin} (M'09, SM'15) is a full professor at Sun Yat-sen University. From 2008 to 2010, he was a postdoctoral fellow at the University of California, Los Angeles. From 2016--2018, he led the SenseTime R\&D teams to develop cutting-edge and deliverable solutions for computer vision, data analysis and mining, and intelligent robotic systems. He has authored and coauthored more than 100 papers in top-tier academic journals and conferences (e.g., 15 papers in TPAMI and IJCV and 60+ papers in CVPR, ICCV, NIPS, and IJCAI). He has served as an associate editor of IEEE Trans. Human-Machine Systems, The Visual Computer, and Neurocomputing and as an area/session chair for numerous conferences, such as CVPR, ICME, ACCV, and ICMR. He was the recipient of the Annual Best Paper Award by Pattern Recognition (Elsevier) in 2018, the Best Paper Diamond Award at IEEE ICME 2017, the Best Paper Runner-Up Award at ACM NPAR 2010, Google Faculty Award in 2012, the Best Student Paper Award at IEEE ICME 2014, and the Hong Kong Scholars Award in 2014. He is an IET Fellow.
\end{IEEEbiography}

\clearpage

\section*{Supplementary Materials}
\label{sec:appendix}
In the main body of this paper, we have presented extensive and fair comparisons of the leading CD-FER algorithms using the unified evaluation benchmark, which also demonstrates the effectiveness of the proposed framework. In this part, we further present more comprehensive experiments and evaluations that we have conducted to provide supplementary analyses.

\subsection*{Credibility of Re-implementation}
To ensure fair comparison, we re-implement several state-of-the-art CD-FER algorithms that use the same backbone and source/target datasets choices, including the ICID algorithm \cite{ji2019cross}, discriminative feature adaptation (DFA) \cite{zhu2016discriminative}, the locality-preserving loss (LPL) \cite{li2017reliable}, a deep emotion transfer network (DETN) \cite{li2018deep}, a fine-tuned deep convolutional network (FTDNN) \cite{zavarez2017cross}, and an ECAN \cite{li2020deeper}. To convincingly show that the re-implementations do justice to the original published works, we also conduct experiments that follow the settings of the original paper to re-produce the results  and compare these results to the those reported in the original paper. As shown in Table \ref{table:ICID-reimplementation-result}, \ref{table:DFA-reimplementation-result}, \ref{table:LPL-reimplementation-result}, \ref{table:DETN-reimplementation-result}, \ref{table:FTDNN-reimplementation-result}, \ref{table:ECAN-reimplementation-result}, our implementations achieve comparable or even better results than those reported in the original works.
The details are presented in the following:

\begin{table}[h]
\centering
\begin{tabular}{c|c|c|cc}
\hline
\centering  Method & Backbone & Source set & CK+ & SFEW2.0\\
\hline
Original Paper & ResNet-101 & CK+ & 94.3 & 30.3 \\
Our Implementation & ResNet-101 & CK+ & 92.26 & 33.55 \\
\hline
\end{tabular}
\vspace{2pt}
\caption{Accuracies of ICID on the CK+ and SFEW2.0 datasets. It includes the reported results by original paper and the results of our implementation.}
\label{table:ICID-reimplementation-result}
\end{table}

\noindent\textbf{ICID} \cite{ji2019cross} originally uses CK+ as the source dataset and ResNet-101 as the backbone. Here, we also adopt the same source and backbone to conduct an experiment and compare with the results reported in the original paper to verify the justice of the re-implementation. As shown in Table \ref{table:result-statistical-distribution}, the overall performances are comparable across the two target datasets. Specifically, the performance of the re-implementation is slightly worse on CK+ while it is better on SFEW2.0.

\begin{table}[htp]
\centering
\begin{tabular}{c|c|c|c}
\hline
\centering  Method & Backbone & Source set & JAFFE \\
\hline
Original Paper & Gabor filter bank & CK+ & 63.38 \\
Our Implementation & Gabor fifter bank & CK+ & 63.49 \\
\hline
\end{tabular}
\vspace{2pt}
\caption{Accuracies of DFA on the JAFFE dataset. It includes the reported results by original paper and the results of our implementation.}
\label{table:DFA-reimplementation-result}
\end{table}

\noindent\textbf{DFA} \cite{zhu2016discriminative} uses Gabor filter bank to extract image features instead of deep neural networks. To ensure fair comparisons, we also use the same feature extractor for re-implementation and follow this work to conduct experiments with CK+ and JAFFE as the source and target datasets, respectively. The comparison results are presented in Table \ref{table:DFA-reimplementation-result}. We observe that the re-implementation performs slightly better than that reported in the original paper.

\begin{table}[htp]
\centering
\begin{tabular}{c|c|c|c}
\hline
\centering  Method & Backbone & Source set & RAF-DB \\
\hline
Original Paper & DLP-CNN & RAF-DB & 76.20\\
Our Implementation & DLP-CNN & RAF-DB & 76.43 \\
\hline
\end{tabular}
\vspace{2pt}
\caption{Accuracies of LPL on the RAF-DB dataset. It includes the reported result by original paper and the results of our implementation.}
\label{table:LPL-reimplementation-result}
\end{table}

\begin{table}[htp]
\centering
\begin{tabular}{c|c|c|c}
\hline
Layer Type & Kernel & Output & Stride  \\
\hline
Conv & 3 & 64 & 1  \\
ReLU & - & - & 1  \\
MaxPooling & 2 & - & 2  \\
Conv & 3 & 96 & 1  \\
ReLU & - & - & 1  \\
MaxPooling & 2 & - & 2  \\
Conv & 3 & 128 & 1  \\
ReLU & - & - & 1  \\
MaxPooling & 2 & - & 2  \\
Conv & 3 & 256 & 1  \\
ReLU & - & - & 1  \\
Conv & 3 & 256 & 1  \\
ReLU & - & - & 1  \\
FC &- & 2000 & - \\
ReLU & - & - & 1  \\
\hline
\end{tabular}
\vspace{2pt}
\caption{The network details of the DLP-CNN \cite{li2017reliable}.}
\label{table:DLP-net}
\end{table}

\noindent\textbf{LPL} \cite{li2017reliable} manually designs a new DLP-CNN for feature extractions as presented in Table \ref{table:DLP-net}. Here, we follow this approach to conduct an experiment that adopts this DLP-CNN as the backbone to present the re-implementation results in Table \ref{table:LPL-reimplementation-result}. This work merely presents the results of a within-dataset setting, in which the source and target datasets are both RAF-DB, and thus, we also report the results under this setting. As shown, the re-implementation achieves slightly better performance. 

\begin{table}[h]
\centering
\begin{tabular}{c|c|c|c}
\hline
\centering  Method & Backbone & Source set & FER2013 \\
\hline
Original Paper & DETN & RAF-DB & 52.37 \\
Our Implementation & DETN & RAF-DB & 53.10 \\
\hline
\end{tabular}
\vspace{2pt}
\caption{Accuracies of DETN on the FER2013 dataset. It includes the reported result by original paper and the results of our implementation.}
\label{table:DETN-reimplementation-result}
\end{table}

\begin{table}[htp]
\centering
\begin{tabular}{c|c|c|c}
\hline
Layer Type & Kernel & Output & Stride  \\
\hline
Conv & 3 & 64 & 1  \\
ReLU & - & - & 1  \\
MaxPooling & 2 & - & 2  \\
Conv & 3 & 96 & 1  \\
ReLU & - & - & 1  \\
MaxPooling & 2 & - & 2  \\
Conv & 3 & 128 & 1  \\
ReLU & - & - & 1  \\
MaxPooling & 2 & - & 2  \\
Conv & 3 & 128 & 1  \\
ReLU & - & - & 1  \\
MaxPooling & 2 & - & 2  \\
Conv & 3 & 256 & 1  \\
ReLU & - & - & 1  \\
Conv & 3 & 256 & 1  \\
ReLU & - & - & 1  \\
\hline
\end{tabular}
\vspace{2pt}
\caption{The network details of the DETN-CNN \cite{li2017reliable}.}
\label{table:DETN-net}
\end{table}

\noindent\textbf{DETN} \cite{li2018deep} also designs a new DETN backbone for feature extraction and utilize RAF and FER2013 as the source and target datasets, respectively. The network details are presented in Table \ref{table:DETN-net}. In addition, we adopt the DETN backbone and use the same source and target datasets for comparisons. The results are presented in Table \ref{table:DETN-reimplementation-result}. The re-implementation results is slightly better than those reported in the original paper. 

\begin{table}[h]
\centering
\begin{tabular}{c|c|c|c}
\hline
\centering  Method & Backbone & Source set & JAFFE \\
\hline
Original Paper & VGG-Face& Six datasets & 44.32 \\
Our Implementation & VGG-Face & Two datasets & 45.07 \\
\hline
\end{tabular}
\vspace{2pt}
\caption{Accuracies of FTDNN on the JAFFE dataset. The six datasets include CK+, MMI, RaFD, KDEF, BU3DFE, and ARFace, while the two datasets contains CK+, MMI.}
\label{table:FTDNN-reimplementation-result}
\end{table}

\noindent\textbf{FTDNN} \cite{zavarez2017cross} adopts six datasets (i.e., CK+, MMI, RaFD, KDEF, BU3DFE, and ARFace) as the source. However, we cannot obtain the RaFD, KDEF, BU3DFE, and ARFace datasets. Thus, we merely select the CK+ and MMI as the source datasets. For fair comparisons, we follow the work to use VGG-Face as the backbone. As shown in Table \ref{table:FTDNN-reimplementation-result}, the re-implementation achieves better performance, although it uses only two datasets.

\begin{table}[h]
\centering
\begin{tabular}{c|c|c|cc}
\hline
\centering  Method & Backbone & Source set & SFEW2.0 & FER2013 \\
\hline
Original Paper & VGG-Face & RAF2.0 & 54.34 & 58.21 \\
Our Implementation & VGG-Face & RAF-DB* & 52.44 & 56.73 \\
\hline
\end{tabular}
\vspace{2pt}
\caption{Accuracies of ECAN on the SFEW2.0 and FER2013 datasets. It includes the reported results by the original paper and the results of our implementation. RAF-DB* indicates that we used RAF and part of ExpW, which follow the image distribution in original paper to the train model.}
\label{table:ECAN-reimplementation-result}
\end{table}

\noindent\textbf{ECAN} \cite{li2020deeper}
ECAN builds a new RAF-DB 2.0 dataset \cite{li2020deeper} that supplements the RAF-DB dataset with 642 images with anger label, images with disgust label and 1,010 images with fear label to address the imbalanced problem. As suggested in work \cite{li2020deeper}, The algorithms using RAF-DB 2.0 as the source dataset can perform much better than those using the RAF-DB. However, this dataset is not available online, and we have asked the authors for this dataset but received no reply. Thus, we can merely use the RAF-DB as the source to conduct experiments and compare with the reported results that use the RAF-DB 2.0 in the original paper. As shown in Table \ref{table:ECAN-reimplementation-result}, the re-implementation with RAF-DB as source dataset achieves slightly worse than the reported results with RAF-DB 2.0 as source dataset. These comparisons verify the justice of the implementation.

\subsection*{Analysis of Different Optimizers}
In this work, we use SGD to train the proposed AGRA framework. As known, training a model with different optimizers may have some influences on the performance. To analyze these influences, we replace SGD with another widely-used Adam optimizer \cite{kingma2014adam} and train the framework following the same process as described in Sec. \ref{Sec:TrainingDetail}. Here, we use ResNet-50 as the backbone network and RAF-DB as the source dataset. As shown in Table \ref{table:result-solver}, replacing the optimizer decreases the mean accuracy from 66.13\% to 65.60\%, a minor decrease of 0.53\%.

\begin{table}[htp]
\centering
\footnotesize
\begin{tabular}{p{1.8cm}|p{0.5cm}p{0.5cm}p{0.75cm}p{0.75cm}p{0.6cm}p{0.6cm}}
\hline
\centering  Method & CK+ & JAFFE & SFEW2.0 &  FER2013 &  ExpW & Mean\\
\hline
\hline
\centering Ours  SGD & 84.91 & 60.97 & 55.45 & 58.34 & 68.33 & 65.60 \\
\centering Ours  Adam  & \textbf{85.27} & \textbf{61.50} & \textbf{56.43} & \textbf{58.95}  & \textbf{68.50} & \textbf{66.13} \\
\hline
\end{tabular}
\vspace{2pt}
\caption{Accuracies of our approach with SGD and Adam optimizers (Ours Adam and Ours SGD, respectively) on the CK+, JAFFE, SFEW2.0, FER2013, and ExpW datasets.}
\label{table:result-solver}
\end{table}

\subsection*{Contribution of Pretraining}
As described in Sec. \ref{Sec:TrainingDetail}, we use the MS-Celeb-1M \cite{guo2016ms} to pretrain the backbone network. To analyze the contribution of this pretraining stage, we conduct an experiment that removes this stage for comparison. Similarly, we also use ResNet-50 as the backbone network and RAF-DB as the source dataset. As shown in Table \ref{table:result-pre}, removing this stage leads to an obvious drop, i.e., a mean accuracy drop of 5.30\%. This comparison well verifies the effectiveness of the pretraining stage.

\begin{table}[htp]
\centering
\footnotesize
\begin{tabular}{p{1.8cm}|p{0.5cm}p{0.5cm}p{0.75cm}p{0.75cm}p{0.6cm}p{0.6cm}}
\hline
\centering  Method & CK+ & JAFFE & SFEW2.0 &  FER2013 &  ExpW & Mean\\
\hline
\hline
\centering Ours w/o pre & 78.22	& 55.07 & 51.97 & 55.38 & 63.51 & 60.83 \\
\centering Ours w/ pre & \textbf{85.27} & \textbf{61.50} & \textbf{56.43} & \textbf{58.95}  & \textbf{68.50} & \textbf{66.13} \\
\hline
\end{tabular}
\vspace{2pt}
\caption{Accuracies of our approach with and without MS-Celeb-1M pretraining (Ours w/ pre and Ours w/o pre) on the CK+, JAFFE, SFEW2.0, FER2013, and ExpW datasets.}
\label{table:result-pre}
\end{table}

\subsection*{Effectiveness of using AFE source dataset on JAFFE}
Both AFE and JAFFE mainly cover Asian faces, while RAF-DB, CK+, SFEW2.0, FER2013, ExpW feature Western individuals. Intuitively, using AFE as the source dataset should obtain better performance on JAFFE compared with those using current RAF-DB, CK+, SFEW2.0, FER2013, ExpW as the source datasets. Here, we experiment to validate this point. In order to evaluate the performance on JAFFE using different source datasets, we average the accuracies of all competing algorithms to compute an algorithm-mean accuracy. As shown in Table \ref{table:result-jaffe}, the algorithm-mean accuracies on JAFFE are 53.76\%, 41.97\%, 31.09\%, 45.70\%, 44.06\% using the RAF-DB, CK+, SFEW2.0, FER2013, ExpW as the source datasets, respectively. These results suggest that the performance is quite poor using current RAF-DB, CK+, SFEW2.0, FER2013, ExpW as the source datasets, which mainly feature Western individuals and are quite different from JAFFE. On the other hand, AFE also mainly covers Asian faces that are more similar to JAFFE. Thus, using this dataset leads to obvious performance improvement, i.e., the algorithm-mean accuracy improvements of 3.42\%, 15.21\%, 26.09\%, 11.48\%, 13.12\%, respectively.

\begin{table}[htp]
\centering
\footnotesize
\begin{tabular}{ccccccc}
\hline
\centering  AFE & RAF-DB & CK+  & SFEW2.0 &  FER2013 &  ExpW \\
\hline
\hline
\centering \textbf{57.18} & 53.76 & 41.97 & 31.09 & 45.70 & 44.06\\
\hline
\end{tabular}
\vspace{2pt}
\caption{The algorithm-mean accuracies of JAFFE using the AFE, RAF-DB, CK+, SFEW2.0, FER2013, and ExpW source datasets. Algorithm-mean accuracy is computed by averaging the accuracies of all the algorithms using ResNet-50 as the backbone network.}
\label{table:result-jaffe}
\end{table}

\subsection*{Performance Analyses Using Different Source Datasets}
In the main body of the paper, we mainly present the comparison results using the RAF-DB and newly-built AFE are the source datasets. For more comprehensive evaluations, we further conduct experiments that select one of the CK+, JAFFE, SFEW2.0, FEER2013, and ExpW datasets as the source and use the rest unselected ones as the target datasets for evaluation. As shown in Table \ref{table:fair-evaluation-results-appendix-resnet50}, \ref{table:fair-evaluation-results-appendix-resnet18}, \ref{table:fair-evaluation-results-appendix-mobilenetv2}, the proposed AGRA framework achieves the best mean accuracies using different source datasets with different backbone networks. With the ResNet-18 backbone, it achieves the averege accuracies of 40.01\%, 41.89\%, 51.63\%, 63.13\%, and 62.76\% when using CK+, JAFFE, SFEW2.0, FER2013, ExpW as source, outperforming the previous-best algorithms by 4.67\%, 6.19\%, 2.06\%, 4.68\%, 4.58\%, respectively.

\begin{table*}[!t]
\centering
\begin{tabular}{c|c|c|cccccccc}
\hline
\centering  Method & Source set & Backbone & CK+ & JAFFE & SFEW2.0 & FER2013 & ExpW & RAF-DB & AFE & Mean\\
\hline

\hline
DT & CK+ & ResNet-50 & - & 45.07 & 19.50 & 29.56 & 16.38 & 23.12 & 29.02 & 27.11 \\ 
PLFT & CK+ & ResNet-50 & - & 47.42 & 22.02 & 34.12 & 18.10 & 26.34 & 33.00 & 30.17 \\ 
ICID \cite{ji2019cross} & CK+ & ResNet-50 & - & 35.21 & 26.83 & 31.31 & 38.36 & 35.41 & 51.47 & 36.43 \\ 
DFA \cite{zhu2016discriminative} & CK+ & ResNet-50 & - & 34.21 & 22.16 & 27.95 & 33.23 & 27.27 & 45.43 & 31.71 \\ 
LPL \cite{li2017reliable} & CK+ & ResNet-50 & - & 46.01 & 20.18 & 28.10 & 10.00 & 15.23 & 23.17 & 23.78 \\ 
DETN \cite{li2018deep} & CK+ & ResNet-50 & - & 35.09 & 18.24 & 23.45 & 15.23 & 17.37 & 16.81 & 21.03 \\ 
FTDNN \cite{zavarez2017cross} & CK+ & ResNet-50 & - & 31.46 & 24.08 & 32.12 & 27.80 & 37.20 & 36.34 & 31.50 \\ 
ECAN \cite{li2020deeper} & CK+ & ResNet-50 & - & 39.83 & 19.33 & 25.34 & 16.75 & 18.74 & 16.62 & 22.77 \\ 
CADA \cite{long2018conditional} & CK+ & ResNet-50 & - & 46.48 & 26.38 & 31.48 & 38.69 & 38.38 & 14.69 & 32.68 \\ 
SAFN \cite{xu2019larger} & CK+ & ResNet-50 & - & 44.13 & 19.27 & 29.90 & 17.07 & 24.00 & 29.75 & 27.35 \\ 
SWD \cite{lee2019sliced} & CK+ & ResNet-50 & - & 46.95 & 27.79 & 32.66 & 25.63 & 29.51 & 53.91 & 36.08 \\ 
JUMBOT \cite{fatras2021unbalanced} & CK+ & ResNet-50 & - & 43.56 & 28.10 & 29.51 & \textbf{41.10} & 34.10 & 54.95 & 38.53 \\
ETD \cite{li2020enhanced} & CK+ & ResNet-50 & - & 44.29 & 24.77 & 31.82 & 38.76 & 32.06 & 47.03 & 36.44 \\
\hline
\textbf{Ours} & CK+ & ResNet-50 & - & \textbf{47.89} & \textbf{30.05} & \textbf{37.79} & 39.49 & \textbf{39.55} & \textbf{59.18} & \textbf{42.33} \\ 
\hline

\hline
DT & JAFFE & ResNet-50 & 46.51 & - & 10.32 & 22.39 & 24.67 & 30.16 & 39.45 & 28.92 \\ 
PLFT & JAFFE & ResNet-50 & 51.16 & - & 20.41 & 25.54 & 27.91 & 33.97 & 44.58 & 33.93 \\ 
ICID \cite{ji2019cross} & JAFFE & ResNet-50 & 38.76  & - & 16.74 & 18.37 & 14.94 & 18.78 & 20.60 & 21.37 \\ 
DFA \cite{zhu2016discriminative} & JAFFE & ResNet-50 & 51.16 & - & 22.94 & 27.71 & 30.21 & 34.72 & 37.21 & 33.99 \\ 
LPL \cite{li2017reliable} & JAFFE & ResNet-50 & 41.85 & - & 18.35 & 16.57 & 29.84 & 33.32 & 46.32 & 31.04 \\ 
DETN \cite{li2018deep} & JAFFE & ResNet-50 & 51.16 & - & 22.71 & 28.97 & 33.45 & 22.17 & 35.34 & 32.30 \\ 
FTDNN \cite{zavarez2017cross} & JAFFE & ResNet-50 & 44.96 & - & 6.42 & 13.59 & 11.04 & 17.57 & 27.22 & 20.13 \\ 
ECAN \cite{li2020deeper} & JAFFE & ResNet-50 & 53.41 & - & 23.05 & 29.03 & \textbf{35.59} & 24.53 & 36.40 & 33.67 \\ 
CADA \cite{long2018conditional} & JAFFE & ResNet-50 & 73.64 & - & \textbf{25.46} & 31.39 & 24.68 & 36.39 & \textbf{59.80} & 41.90 \\ 
SAFN \cite{xu2019larger} & JAFFE & ResNet-50 & 44.96 & - & 10.32 & 23.80 & 25.68 & 31.20 & 40.43 & 29.40 \\ 
SWD \cite{lee2019sliced} & JAFFE & ResNet-50 & 55.81 & - & 16.51 & 25.12 & 28.23 & 31.14 & 45.36 & 33.70 \\ 
JUMBOT \cite{fatras2021unbalanced} & JAFFE & ResNet-50 & 51.94 & - & 14.45 & 25.31 & 26.36 & 33.32 & 52.44 & 33.97 \\
ETD \cite{li2020enhanced} & JAFFE & ResNet-50 & 56.67 & - & 10.64 & 23.49 & 24.85 & 35.49 & 51.14 & 33.71 \\
\hline
\textbf{Ours} & JAFFE & ResNet-50 & \textbf{78.99} & - & 18.81 & \textbf{32.21} & 29.30 & \textbf{38.15} & 53.99 & \textbf{41.91} \\ 
\hline

\hline
DT & SFEW2.0 & ResNet-50 & 48.06 & 27.70 & - & 36.06 & 42.80 & 45.71 & 60.37 & 43.45 \\ 
PLFT & SFEW2.0 & ResNet-50 & 55.04 & 28.64 & - & 40.45 & 45.26 & 49.85 & 65.53 & 47.46 \\ 
ICID \cite{ji2019cross} & SFEW2.0 & ResNet-50 & 35.74 & 23.94 & - & 36.34 & 48.00 & 48.16 & 61.09 & 42.11 \\ 
DFA \cite{zhu2016discriminative} & SFEW2.0 & ResNet-50 & 15.50 & 22.54 & - & 33.42 & 44.87 & 41.41 & 6.45 & 27.37 \\ 
LPL \cite{li2017reliable} & SFEW2.0 & ResNet-50 & 51.94 & 28.17 & - & 37.38 & 37.93 & 50.18 & 66.67 & 45.38 \\ 
DETN \cite{li2018deep} & SFEW2.0 & ResNet-50 & 44.81 & 25.87 & - & 36.17 & 43.19 & 38.25 & 36.12 & 37.40 \\ 
FTDNN \cite{zavarez2017cross} & SFEW2.0 & ResNet-50 & 63.09 & 34.94 & - & 47.34 & 57.65 & 54.40 & 61.20 & 53.10 \\ 
ECAN \cite{li2020deeper} & SFEW2.0 & ResNet-50 & 45.00 & 27.45 & - & 43.05 & 45.20 & 38.02 & 36.05 & 39.13 \\ 
CADA \cite{long2018conditional} & SFEW2.0 & ResNet-50 & \textbf{65.12} & \textbf{41.31} & - & 40.68 & 47.35 & 44.93 & 55.15 & 49.09 \\ 
SAFN \cite{xu2019larger} & SFEW2.0 & ResNet-50 & 48.06 & 27.70 & - & 35.86 & 42.43 & 44.93 & 60.54 & 43.25 \\ 
SWD \cite{lee2019sliced} & SFEW2.0 & ResNet-50 & 59.69 & 38.03 & - & 43.74 & 48.20 & \textbf{51.12} & \textbf{65.49} & 51.05 \\ 
JUMBOT \cite{fatras2021unbalanced} & SFEW2.0 & ResNet-50 & 61.94 & 35.21 & - & 46.33 & 49.32 & 47.90 & 60.08 & 50.13 \\
ETD \cite{li2020enhanced} & SFEW2.0 & ResNet-50 & 59.52 & 34.38 & - & 44.00 & 49.54 & 45.40 & 61.05 & 48.98 \\
\hline
\textbf{Ours} & SFEW2.0 & ResNet-50 & 63.57 & 39.44 & - & \textbf{48.53} & \textbf{59.48} & 50.70 & 58.69 & \textbf{53.40} \\ 
\hline

\hline
DT & FER2013 & ResNet-50 & 68.99 & 44.13 & 42.43 & - & 54.17 & 63.84 & 70.45 & 57.34 \\ 
PLFT & FER2013 & ResNet-50 & 79.07 & 46.48 & 42.20 & - & 54.96 & 67.92 & 73.11 & 60.62 \\ 
ICID \cite{ji2019cross} & FER2013 & ResNet-50 & 63.57 & 44.60 & 43.58 & - & 54.01 & 62.08 & 66.09 & 55.66 \\ 
DFA \cite{zhu2016discriminative} & FER2013 & ResNet-50 & 55.81 & 42.25 & 34.86 & - & 44.55 & 48.84 & 63.67 & 48.33 \\ 
LPL \cite{li2017reliable} & FER2013 & ResNet-50 & 68.22 & 42.72 & 44.27 & - & 52.45 & 64.23 & 68.47 & 56.73 \\ 
DETN \cite{li2018deep} & FER2013 & ResNet-50 & 65.89 & 37.89 & 37.39 & - & 52.15 & 50.51 & 63.90 & 51.29 \\ 
FTDNN \cite{zavarez2017cross} & FER2013 & ResNet-50 & 72.09 & \textbf{53.99} & 45.64 & - & 54.67 & 64.40 & 70.20 & 60.17 \\ 
ECAN \cite{li2020deeper} & FER2013 & ResNet-50 & 60.47 & 41.76 & 46.01 & - & 48.88 & 53.41 & 63.72 & 52.38 \\ 
CADA \cite{long2018conditional} & FER2013 & ResNet-50 & 81.40 & 45.07 & 46.33 & - & 54.84 & 65.96 & 62.78 & 59.40 \\ 
SAFN \cite{xu2019larger} & FER2013 & ResNet-50 & 68.99 & 45.07 & 38.07 & - & 53.91 & 62.80 & 69.49 & 56.39 \\ 
SWD \cite{lee2019sliced} & FER2013 & ResNet-50 & 65.89 & 49.30 & 45.64 & - & 56.05 & 65.28 & 68.39 & 58.43 \\ 
JUMBOT \cite{fatras2021unbalanced} & FER2013 & ResNet-50 & 75.76 & 49.69 & 44.33 & - & 52.37 & 63.08 & 67.55 & 58.80 \\
ETD \cite{li2020enhanced} & FER2013 & ResNet-50 & 77.42 & 44.17 & 39.58 & - & 49.97 & 61.18 & 65.68 & 56.33 \\
\hline
\textbf{Ours} & FER2013 & ResNet-50 & \textbf{85.69} & 52.74 & \textbf{49.31} & - & \textbf{60.23} & \textbf{67.62} & \textbf{76.11} & \textbf{65.28} \\ 
\hline

\hline
DT & ExpW & ResNet-50 & 45.74 & 38.97 & 40.37 & 46.69 & - & 70.62 & 69.56 & 51.99 \\ 
PLFT & ExpW & ResNet-50 & 61.24 & 43.19 & 44.04 & 48.58 & - & 72.94 & 72.69 & 57.11 \\ 
ICID \cite{ji2019cross} & ExpW & ResNet-50 & 58.91 & 38.50 & 43.81 & 45.91 & - & 69.38 & 70.52 & 54.51 \\ 
DFA \cite{zhu2016discriminative} & ExpW & ResNet-50 & 63.33 & 34.74 & 33.62 & 39.87 & - & 63.84 & 65.83 & 50.21 \\ 
LPL \cite{li2017reliable} & ExpW & ResNet-50 & 69.77 & 42.25 & 41.74 & 45.74 & - & 67.33 & 71.79 & 56.44 \\ 
DETN \cite{li2018deep} & ExpW & ResNet-50 & 42.11 & 42.39 & 34.54 & 31.70 & - & 44.93 & 58.03 & 42.28 \\ 
FTDNN \cite{zavarez2017cross} & ExpW & ResNet-50 & 62.79 & 52.58 & 45.41 & 49.51 & - & 73.07 & 78.00 & 60.23 \\ 
ECAN \cite{li2020deeper} & ExpW & ResNet-50 & 43.56 & \textbf{53.87} & 37.75 & 41.84 & - & 44.54 & 58.77 & 46.72 \\ 
CADA \cite{long2018conditional} & ExpW & ResNet-50 & 72.09 & 52.11 & 47.71 & 50.41 & - & 73.85 & 80.15 & 62.72 \\ 
SAFN \cite{xu2019larger} & ExpW & ResNet-50 & 58.91 & 42.72 & 37.84 & 47.65 & - & 70.13 & 73.97 & 55.20 \\ 
SWD \cite{lee2019sliced} & ExpW & ResNet-50 & 72.87 & 41.31 & 43.58 & 48.44 & - & 66.81 & 69.84 & 57.14 \\ 
JUMBOT \cite{fatras2021unbalanced} & ExpW & ResNet-50 & 70.33 & 44.05 & 39.92 & 46.32 & - & 69.78 & 71.36 & 56.96 \\
ETD \cite{li2020enhanced} & ExpW & ResNet-50 & 69.52 & 40.65 & 43.34 & 43.21 & - & 67.46 & 68.35 & 55.42 \\
\hline
\textbf{Ours} & ExpW & ResNet-50 & \textbf{79.77} & 49.44 & \textbf{48.95} & \textbf{52.12} & - & \textbf{76.88} & \textbf{79.08} & \textbf{64.37} \\ 
\hline
\end{tabular}
\vspace{2pt}
\caption{Accuracies of our proposed framework with current leading methods on the CK+, JAFFE, SFEW2.0, FER2013, ExpW, RAF-DB and AFE datasets. The results are generated by our implementation with exactly the same source dataset and backbone network.}
\label{table:fair-evaluation-results-appendix-resnet50}
\end{table*}

\begin{table*}[!t]
\centering
\begin{tabular}{c|c|c|cccccccc}
\hline
\centering  Method & Source set & Backbone & CK+ & JAFFE & SFEW2.0 & FER2013 & ExpW & RAF-DB & AFE & Mean\\
\hline

\hline
DT & CK+ & ResNet-18 & - & 44.13 & 16.74 & 22.59 & 9.53 & 14.97 & 31.46 & 23.24 \\ 
PLFT & CK+ & ResNet-18 & - & 43.66 & 19.50 & 22.95 & 11.82 & 11.74 & 39.31 & 24.83 \\ 
ICID \cite{ji2019cross} & CK+ & ResNet-18 & - & 27.23 & 17.20 & 23.77 & 14.90 & 19.73 & 23.13 & 20.99 \\ 
DFA \cite{zhu2016discriminative} & CK+ & ResNet-18 & - & 21.60 & 11.01 & 15.33 & 15.36 & 9.78 & 25.34 & 16.40 \\ 
LPL \cite{li2017reliable} & CK+ & ResNet-18 & - & 46.48 & 20.64 & 26.75 & 20.46 & 23.35 & 36.58 & 29.04 \\ 
DETN \cite{li2018deep} & CK+ & ResNet-18 & - & 16.54 & 11.93 & 18.74 & 16.78 & 23.26 & 17.67 & 17.49 \\ 
FTDNN \cite{zavarez2017cross} & CK+ & ResNet-18 & - & 34.27 & 20.41 & 25.74 & 21.41 & 22.11 & 38.65 & 27.10 \\ 
ECAN \cite{li2020deeper} & CK+ & ResNet-18 & - & 24.24 & 12.34 & 19.33 & 23.26 & 26.04 & 26.32 & 21.92 \\ 
CADA \cite{long2018conditional} & CK+ & ResNet-18 & - & 48.58 & 27.98 & 28.33 & 36.36 & 32.74 & 38.03 & 35.34 \\ 
SAFN \cite{xu2019larger} & CK+ & ResNet-18 & - & 41.31 & 20.18 & 21.69 & 16.02 & 16.92 & 36.36 & 25.41 \\ 
SWD \cite{lee2019sliced} & CK+ & ResNet-18 & - & 39.44 & 25.23 & \textbf{33.95} & 27.48 & 32.02 & 46.95 & 34.18 \\ 
JUMBOT \cite{fatras2021unbalanced} & CK+ & ResNet-18 & - & 41.67 & 25.83 & 25.71 & 25.65 & 27.44 & 42.09 & 31.40 \\
ETD \cite{li2020enhanced} & CK+ & ResNet-18 & - & 38.03 & 20.36 & 23.15 & 29.84 & 24.47 & 41.96 & 29.64 \\
\hline
\textbf{Ours} & CK+ & ResNet-18 & - & \textbf{49.30} & \textbf{28.44} & 30.89 & \textbf{37.52} & \textbf{35.44} & \textbf{58.44} & \textbf{40.01} \\ 
\hline

\hline
DT & JAFFE & ResNet-18 & 32.56 & - & 6.24 & 8.27 & 7.19 & 9.23 & 15.18 & 13.11 \\ 
PLFT & JAFFE & ResNet-18 & 32.56 & - & 6.88 & 8.55 & 8.73 & 10.07 & 17.07 & 13.98 \\ 
ICID \cite{ji2019cross} & JAFFE & ResNet-18 & 24.03 & - & 16.74 & 18.54 & 11.61 & 16.27 & 8.83 & 16.00 \\ 
DFA \cite{zhu2016discriminative} & JAFFE & ResNet-18 & 48.84 & - & 11.93 & 11.20 & 8.21 & 13.24 & 15.35 & 18.13 \\ 
LPL \cite{li2017reliable} & JAFFE & ResNet-18 & 41.86 & - & 15.83 & 23.40 & 24.01 & 24.56 & 38.13 & 27.97 \\ 
DETN \cite{li2018deep} & JAFFE & ResNet-18 & 31.01 & - & 21.56 & 27.03 & \textbf{32.81} & 26.77 & 29.40 & 28.10 \\ 
FTDNN \cite{zavarez2017cross} & JAFFE & ResNet-18 & 31.78 & - & 17.43 & 15.84 & 11.87 & 17.84 & 10.52 & 17.55 \\ 
ECAN \cite{li2020deeper} & JAFFE & ResNet-18 & 40.31 & - & 22.47 & 27.06 & 31.97 & 27.13 & 32.34 & 30.21 \\ 
CADA \cite{long2018conditional} & JAFFE & ResNet-18 & 73.64 & - & 25.23 & 26.79 & 28.28 & 29.61 & 30.62 & 35.70 \\ 
SAFN \cite{xu2019larger} & JAFFE & ResNet-18 & 41.86 & - & 9.63 & 17.13 & 20.61 & 24.75 & 17.59 & 21.93 \\ 
SWD \cite{lee2019sliced} & JAFFE & ResNet-18 & 51.94 & - & 17.66 & 26.95 & 29.90 & 31.79 & \textbf{44.71} & 33.83 \\ 
JUMBOT \cite{fatras2021unbalanced} & JAFFE & ResNet-18 & 54.55 & - & 27.70 & 20.18 & 26.27 & 36.85 & 34.14 & 33.28 \\
ETD \cite{li2020enhanced} & JAFFE & ResNet-18 & 53.74 & - & 23.53 & 21.35 & 21.57 & 30.28 & 34.78 & 30.88 \\
\hline
\textbf{Ours} & JAFFE & ResNet-18 & \textbf{82.17} & - & \textbf{29.36} & \textbf{28.44} & 29.86 & \textbf{39.71} & 41.81 & \textbf{41.89} \\ 
\hline

\hline
DT & SFEW2.0 & ResNet-18 & 40.31 & 21.13 & - & 31.81 & 39.77 & 41.60 & 58.96 & 38.93 \\ 
PLFT & SFEW2.0 & ResNet-18 & 40.31 & 23.47 & - & 36.01 & 41.93 & 42.81 & 60.98 & 40.92 \\ 
ICID \cite{ji2019cross} & SFEW2.0 & ResNet-18 & 39.53 & 29.58 & - & 36.93 & 41.87 & 43.23 & 57.44 & 41.43 \\ 
DFA \cite{zhu2016discriminative} & SFEW2.0 & ResNet-18 & 35.71 & 26.67 & - & 41.54 & 34.15 & 38.44 & 44.19 & 36.78 \\ 
LPL \cite{li2017reliable} & SFEW2.0 & ResNet-18 & 57.36 & 30.05 & - & 38.40 & 48.01 & \textbf{48.73} & 56.25 & 46.47 \\ 
DETN \cite{li2018deep} & SFEW2.0 & ResNet-18 & 36.28 & 22.61 & - & 34.09 & 37.75 & 25.86 & 37.24 & 32.31 \\ 
FTDNN \cite{zavarez2017cross} & SFEW2.0 & ResNet-18 & 47.87 & 30.77 & - & 37.15 & 35.09 & 35.82 & 53.44 & 40.02 \\ 
ECAN \cite{li2020deeper} & SFEW2.0 & ResNet-18 & 34.73 & 25.65 & - & 33.81 & 36.66 & 28.05 & 37.21 & 32.69 \\ 
CADA \cite{long2018conditional} & SFEW2.0 & ResNet-18 & 55.81 & 40.38 & - & 38.45 & 35.00 & 44.31 & 48.32 & 43.71 \\ 
SAFN \cite{xu2019larger} & SFEW2.0 & ResNet-18 & 41.86 & 21.13 & - & 32.10 & 48.07 & 46.63 & 57.72 & 41.25 \\ 
SWD \cite{lee2019sliced} & SFEW2.0 & ResNet-18 & 57.36 & 36.62 & - & \textbf{44.05} & \textbf{49.83} & 46.21 & 63.34 & 49.57 \\ 
JUMBOT \cite{fatras2021unbalanced} & SFEW2.0 & ResNet-18 & 51.37 & 33.62 & - & 35.67 & 40.41 & 43.19 & 60.19 & 44.08 \\
ETD \cite{li2020enhanced} & SFEW2.0 & ResNet-18 & 49.18 & 35.96 & - & 36.50 & 39.68 & 40.89 & 58.05 & 43.38 \\
\hline
\textbf{Ours} & SFEW2.0 & ResNet-18 & \textbf{60.47} & \textbf{46.01} & - & 42.67 & 45.16 & 47.38 & \textbf{68.08} & \textbf{51.63} \\ 
\hline

\hline
DT & FER2013 & ResNet-18 & 56.59 & 40.38 & 41.28 & - & 52.53 & 57.87 & 66.56 & 52.54 \\ 
PLFT & FER2013 & ResNet-18 & 72.87 & 39.44 & 41.28 & - & 54.43 & 60.09 & 71.09 & 56.53 \\ 
ICID \cite{ji2019cross} & FER2013 & ResNet-18 & 68.22 & 38.50 & 41.51 & - & 53.07 & 60.71 & 68.82 & 55.14 \\ 
DFA \cite{zhu2016discriminative} & FER2013 & ResNet-18 & 61.24 & 45.54 & 41.89 & - & 47.10 & 49.56 & 71.00 & 52.72 \\ 
LPL \cite{li2017reliable} & FER2013 & ResNet-18 & 65.89 & 45.07 & 43.35 & - & 55.28 & 63.42 & 68.82 & 56.97 \\ 
DETN \cite{li2018deep} & FER2013 & ResNet-18 & 72.09 & 33.65 & 16.97 & - & 33.96 & 57.14 & 64.38 & 46.37 \\ 
FTDNN \cite{zavarez2017cross} & FER2013 & ResNet-18 & 63.57 & 43.66 & 44.04 & - & 54.12 & 61.17 & 70.35 & 56.15 \\ 
ECAN \cite{li2020deeper} & FER2013 & ResNet-18 & 65.89 & 37.86 & 31.65 & - & 50.36 & 59.35 & 65.24 & 51.73 \\ 
CADA \cite{long2018conditional} & FER2013 & ResNet-18 & 70.54 & 46.48 & 45.18 & - & 53.15 & 60.65 & 61.28 & 56.21 \\ 
SAFN \cite{xu2019larger} & FER2013 & ResNet-18 & 64.34 & 46.01 & 42.89 & - & 53.01 & 58.62 & 69.20 & 55.68 \\ 
SWD \cite{lee2019sliced} & FER2013 & ResNet-18 & 72.87 & 46.48 & 42.43 & - & 54.94 & 62.80 & 71.18 & 58.45 \\ 
JUMBOT \cite{fatras2021unbalanced} & FER2013 & ResNet-18 & 75.00 & 41.46 & 39.72 & - & 44.72 & 58.71 & 67.15 & 54.46 \\
ETD \cite{li2020enhanced} & FER2013 & ResNet-18 & 71.54 & 40.35 & 36.67 & - & 42.25 & 54.89 & 66.71 & 52.07 \\
\hline
\textbf{Ours} & FER2013 & ResNet-18 & \textbf{80.95} & \textbf{48.16} & \textbf{49.54} & - & \textbf{57.27} & \textbf{65.12} & \textbf{77.72} & \textbf{63.13} \\ 
\hline

\hline
DT & ExpW & ResNet-18 & 64.34 & 35.68 & 38.30 & 46.27 & - & 65.50 & 71.14 & 53.54 \\ 
PLFT & ExpW & ResNet-18 & 72.09 & 38.50 & 37.39 & 47.23 & - & 68.08 & 74.21 & 56.25 \\ 
ICID \cite{ji2019cross} & ExpW & ResNet-18 & 61.24 & 40.85 & 40.83 & 45.77 & - & 69.97 & 73.08 & 55.29 \\ 
DFA \cite{zhu2016discriminative} & ExpW & ResNet-18 & 37.21 & 34.74 & 26.38 & 32.63 & - & 44.51 & 61.12 & 39.43 \\ 
LPL \cite{li2017reliable} & ExpW & ResNet-18 & 58.14 & 40.38 & 41.06 & 47.45 & - & 69.35 & 76.10 & 55.41 \\ 
DETN \cite{li2018deep} & ExpW & ResNet-18 & 32.56 & 25.56 & 29.13 & 29.03 & - & 44.15 & 57.07 & 36.25 \\ 
FTDNN \cite{zavarez2017cross} & ExpW & ResNet-18 & 52.71 & 38.97 & 42.66 & 47.96 & - & 62.80 & 74.80 & 53.32 \\ 
ECAN \cite{li2020deeper} & ExpW & ResNet-18 & 31.78 & 26.59 & 29.77 & 30.01 & - & 43.07 & 56.94 & 36.36 \\ 
CADA \cite{long2018conditional} & ExpW & ResNet-18 & 65.12 & 45.54 & 43.35 & 48.30 & - & 69.64 & 77.13 & 58.18 \\ 
SAFN \cite{xu2019larger} & ExpW & ResNet-18 & 48.06 & 35.68 & 43.12 & 45.91 & - & 67.17 & 74.76 & 52.45 \\ 
SWD \cite{lee2019sliced} & ExpW & ResNet-18 & 63.57 & 44.13 & 41.51 & 49.09 & - & \textbf{71.83} & 72.74 & 57.15 \\ 
JUMBOT \cite{fatras2021unbalanced} & ExpW & ResNet-18 & 61.01 & 36.29 & 44.52 & 45.08 & - & 67.45 & 71.05 & 54.23 \\
ETD \cite{li2020enhanced} & ExpW & ResNet-18 & 66.36 & 35.79 & 40.48 & 47.53 & - & 63.44 & 75.38 & 54.83 \\
\hline
\textbf{Ours} & ExpW & ResNet-18 & \textbf{83.33} & \textbf{46.01} & \textbf{46.79} & \textbf{51.26} & - & 71.14 & \textbf{78.00} & \textbf{62.76} \\ 
\hline
\end{tabular}
\vspace{2pt}
\caption{Accuracies of our proposed framework with current leading methods on the CK+, JAFFE, SFEW2.0, FER2013, ExpW, RAF-DB and AFE datasets. The results are generated by our implementation with exactly the same source dataset and backbone network.}
\label{table:fair-evaluation-results-appendix-resnet18}
\end{table*}

\begin{table*}[!t]
\centering
\begin{tabular}{c|c|c|cccccccc}
\hline
\centering  Method & Source set & Backbone & CK+ & JAFFE & SFEW2.0 & FER2013 & ExpW & RAF-DB & AFE & Mean\\
\hline

\hline
DT & CK+ & MobileNet-v2 & - & 30.52 & 14.45 & 25.96 & 14.76 & 17.44 & 20.80 & 20.66 \\ 
PLFT & CK+ & MobileNet-v2 & - & 32.39 & 16.97 & 25.60 & 13.19 & 15.26 & 24.31 & 21.29 \\ 
ICID \cite{ji2019cross} & CK+ & MobileNet-v2 & - & 34.27 & 19.95 & 21.91 & 24.45 & 19.66 & 22.45 & 23.78 \\ 
DFA \cite{zhu2016discriminative} & CK+ & MobileNet-v2 & - & 14.08 & 13.53 & 13.19 & 20.61 & 19.77 & 12.07 & 15.54 \\ 
LPL \cite{li2017reliable} & CK+ & MobileNet-v2 & - & 26.29 & 20.18 & 24.87 & 12.14 & 15.36 & 13.24 & 18.68 \\ 
DETN \cite{li2018deep} & CK+ & MobileNet-v2 & - & 20.84 & 18.73 & 19.29 & 15.09 & 17.27 & 16.58 & 17.97 \\ 
FTDNN \cite{zavarez2017cross} & CK+ & MobileNet-v2 & - & 32.39 & 15.83 & 23.38 & 13.00 & 16.34 & 12.16 & 18.85 \\ 
ECAN \cite{li2020deeper} & CK+ & MobileNet-v2 & - & 26.23 & 20.22 & 22.58 & 19.38 & 17.71 & 17.09 & 20.54 \\ 
CADA \cite{long2018conditional} & CK+ & MobileNet-v2 & - & 38.97 & 21.33 & 26.67 & 25.80 & 19.04 & 24.05 & 25.98 \\ 
SAFN \cite{xu2019larger} & CK+ & MobileNet-v2 & - & 25.82 & 15.83 & 23.26 & 21.97 & 14.28 & 16.47 & 19.61 \\ 
SWD \cite{lee2019sliced} & CK+ & MobileNet-v2 & - & 29.58 & 15.83 & 27.17 & 26.99 & \textbf{26.54} & \textbf{32.47} & 26.43 \\ 
JUMBOT \cite{fatras2021unbalanced} & CK+ & MobileNet-v2 & - & 31.46 & 17.43 & 25.32 & 27.18 & 20.61 & 32.29 & 26.05 \\
ETD \cite{li2020enhanced} & CK+ & MobileNet-v2 & - & 31.32 & 19.83 & 28.05 & 24.36 & 21.62 & 25.57 & 25.13 \\
\hline
\textbf{Ours} & CK+ & MobileNet-v2 & - & \textbf{39.33} & \textbf{29.72} & \textbf{29.42} & \textbf{29.79} & 22.91 & 30.52 & \textbf{30.28} \\ 
\hline

\hline
DT & JAFFE & MobileNet-v2 & 37.98 & - & 12.84 & 8.13 & 6.51 & 8.90 & 6.92 & 13.55 \\ 
PLFT & JAFFE & MobileNet-v2 & 39.53 & - & 13.99 & 6.75 & 6.44 & 9.06 & 7.06 & 13.81 \\ 
ICID \cite{ji2019cross} & JAFFE & MobileNet-v2 & 34.88 & - & 13.76 & 14.60 & 8.40 & 12.26 & 8.12 & 15.34 \\ 
DFA \cite{zhu2016discriminative} & JAFFE & MobileNet-v2 & 27.74 & - & 8.47 & 16.25 & 13.48 & 11.29 & 12.87 & 15.02 \\ 
LPL \cite{li2017reliable} & JAFFE & MobileNet-v2 & 31.01 & - & 11.01 & 14.74 & 8.24 & 12.19 & 9.76 & 14.49 \\ 
DETN \cite{li2018deep} & JAFFE & MobileNet-v2 & 26.23 & - & 9.46 & 8.70 & 9.43 & 14.33 & 14.34 & 13.75 \\ 
FTDNN \cite{zavarez2017cross} & JAFFE & MobileNet-v2 & 31.01 & - & 13.53 & 11.67 & 8.73 & 10.89 & 11.03 & 14.48 \\ 
ECAN \cite{li2020deeper} & JAFFE & MobileNet-v2 & 33.81 & - & 12.38 & 12.33 & 13.87 & 15.29 & 13.24 & 16.82 \\ 
CADA \cite{long2018conditional} & JAFFE & MobileNet-v2 & 37.21 & - & 21.10 & 17.83 & 15.11 & 16.99 & 19.41 & 21.28 \\ 
SAFN \cite{xu2019larger} & JAFFE & MobileNet-v2 & 29.46 & - & 18.58 & 10.07 & 7.20 & 9.03 & 6.90 & 13.54 \\ 
SWD \cite{lee2019sliced} & JAFFE & MobileNet-v2 & 31.01 & - & 13.07 & 19.77 & 13.64 & 15.42 & 20.64 & 18.93 \\ 
JUMBOT \cite{fatras2021unbalanced} & JAFFE & MobileNet-v2 & 41.03 & - & 18.35 & 19.35 & 11.39 & 14.74 & 19.23 & 20.68 \\
ETD \cite{li2020enhanced} & JAFFE & MobileNet-v2 & 34.29 & - & 19.67 & 21.26 & 9.69 & 12.94 & 21.18 & 19.84 \\
\hline
\textbf{Ours} & JAFFE & MobileNet-v2 & \textbf{78.40} & - & \textbf{27.20} & \textbf{28.34} & \textbf{27.71} & \textbf{21.85} & \textbf{24.16} & \textbf{34.61} \\ 
\hline

\hline
DT & SFEW2.0 & MobileNet-v2 & 26.36 & 17.37 & - & 19.07 & 37.44 & 31.53 & 43.22 & 29.17 \\ 
PLFT & SFEW2.0 & MobileNet-v2 & 27.13 & 23.00 & - & 19.27 & 41.53 & 34.14 & 42.23 & 31.22 \\ 
ICID \cite{ji2019cross} & SFEW2.0 & MobileNet-v2 & 17.05 & 16.90 & - & 25.26 & 37.59 & 28.07 & 37.72 & 27.1 \\ 
DFA \cite{zhu2016discriminative} & SFEW2.0 & MobileNet-v2 & 22.03 & 14.79 & - & 13.54 & 25.36 & 28.92 & 33.86 & 23.08 \\ 
LPL \cite{li2017reliable} & SFEW2.0 & MobileNet-v2 & 28.68 & 15.49 & - & 23.84 & 33.48 & 33.29 & 45.20 & 30.00 \\ 
DETN \cite{li2018deep} & SFEW2.0 & MobileNet-v2 & 17.48 & 23.21 & - & 24.18 & 34.73 & 27.18 & 28.31 & 25.85 \\ 
FTDNN \cite{zavarez2017cross} & SFEW2.0 & MobileNet-v2 & 21.21 & 25.61 & - & 27.50 & 42.11 & 36.23 & 35.42 & 31.35 \\ 
ECAN \cite{li2020deeper} & SFEW2.0 & MobileNet-v2 & 19.81 & 26.57 & - & 26.98 & 37.58 & 29.70 & 32.26 & 28.82 \\ 
CADA \cite{long2018conditional} & SFEW2.0 & MobileNet-v2 & 31.01 & 25.82 & - & 28.75 & 36.64 & 31.56 & 36.34 & 31.69 \\ 
SAFN \cite{xu2019larger} & SFEW2.0 & MobileNet-v2 & 18.60 & 13.62 & - & 27.03 & 38.99 & 25.79 & 41.94 & 27.66 \\ 
SWD \cite{lee2019sliced} & SFEW2.0 & MobileNet-v2 & 36.43 & 23.94 & - & \textbf{31.70} & 43.63 & \textbf{40.37} & \textbf{51.42} & 37.92 \\ 
JUMBOT \cite{fatras2021unbalanced} & SFEW2.0 & MobileNet-v2 & 33.18 & 29.25 & - & 30.15 & 44.68 & 35.75 & 44.94 & 36.33 \\
ETD \cite{li2020enhanced} & SFEW2.0 & MobileNet-v2 & 29.63 & 24.31 & - & 27.52 & 45.41 & 30.77 & 42.86 & 33.42 \\
\hline
\textbf{Ours} & SFEW2.0 & MobileNet-v2 & \textbf{41.86} & \textbf{32.43} & - & 30.69 & \textbf{48.57} & 37.65 & 45.74 & \textbf{39.49} \\ 
\hline

\hline
DT & FER2013 & MobileNet-v2 & 62.02 & 39.44 & 30.96 & - & 47.05 & 40.95 & 40.74 & 43.53 \\ 
PLFT & FER2013 & MobileNet-v2 & 61.24 & 44.60 & 28.67 & - & 53.63 & 40.30 & 35.67 & 44.02 \\ 
ICID \cite{ji2019cross} & FER2013 & MobileNet-v2 & 55.81 & 39.44 & 31.42 & - & 41.50 & 41.21 & 42.48 & 41.98 \\ 
DFA \cite{zhu2016discriminative} & FER2013 & MobileNet-v2 & 55.81 & 36.15 & 27.78 & - & 43.83 & 34.18 & 44.89 & 40.44 \\ 
LPL \cite{li2017reliable} & FER2013 & MobileNet-v2 & 60.47 & 37.56 & 31.88 & - & 49.83 & 43.92 & 48.88 & 45.42 \\ 
DETN \cite{li2018deep} & FER2013 & MobileNet-v2 & 48.09 & 42.31 & 27.54 & - & 39.14 & 40.53 & 39.18 & 39.47 \\ 
FTDNN \cite{zavarez2017cross} & FER2013 & MobileNet-v2 & 59.69 & 45.54 & 39.68 & - & 49.87 & 52.43 & 59.72 & 51.16 \\ 
ECAN \cite{li2020deeper} & FER2013 & MobileNet-v2 & 55.65 & 44.12 & 28.46 & - & 41.53 & 42.31 & 46.91 & 43.16 \\ 
CADA \cite{long2018conditional} & FER2013 & MobileNet-v2 & 66.67 & \textbf{50.23} & 41.28 & - & 51.84 & \textbf{53.15} & 61.53 & 54.12 \\ 
SAFN \cite{xu2019larger} & FER2013 & MobileNet-v2 & 66.67 & 37.56 & 35.78 & - & 45.56 & 38.73 & 50.55 & 45.81 \\ 
SWD \cite{lee2019sliced} & FER2013 & MobileNet-v2 & 53.49 & 48.36 & 35.78 & - & 50.02 & 47.44 & 53.57 & 48.11 \\ 
JUMBOT \cite{fatras2021unbalanced} & FER2013 & MobileNet-v2 & 51.16 & 41.54 & 36.06 & - & 49.30 & 44.93 & 54.32 & 46.22 \\
ETD \cite{li2020enhanced} & FER2013 & MobileNet-v2 & 54.55 & 40.32 & 30.77 & - & 45.91 & 50.54 & 57.74 & 46.64 \\
\hline
\textbf{Ours} & FER2013 & MobileNet-v2 & \textbf{67.44} & 47.89 & \textbf{41.74} & - & \textbf{59.41} & 52.27 & \textbf{64.00} & \textbf{55.46} \\ 
\hline

\hline
DT & ExpW & MobileNet-v2 & 56.59 & 38.50 & 37.39 & 44.87 & - & 63.45 & 72.78 & 52.26 \\ 
PLFT & ExpW & MobileNet-v2 & 65.12 & 41.31 & 37.84 & 46.33 & - & 63.84 & 74.57 & 54.84 \\ 
ICID \cite{ji2019cross} & ExpW & MobileNet-v2 & 44.96 & 29.58 & 33.26 & 39.80 & - & 57.45 & 67.37 & 45.40 \\ 
DFA \cite{zhu2016discriminative} & ExpW & MobileNet-v2 & 31.01 & 23.59 & 22.48 & 28.19 & - & 39.68 & 59.85 & 34.13 \\ 
LPL \cite{li2017reliable} & ExpW & MobileNet-v2 & 60.47 & 33.80 & 36.47 & 44.25 & - & 64.53 & 72.34 & 51.98 \\ 
DETN \cite{li2018deep} & ExpW & MobileNet-v2 & 40.55 & 26.92 & 29.12 & 31.10 & - & 55.29 & 56.64 & 39.94 \\ 
FTDNN \cite{zavarez2017cross} & ExpW & MobileNet-v2 & 63.57 & 39.44 & 39.22 & 45.26 & - & 68.57 & 73.72 & 54.96 \\ 
ECAN \cite{li2020deeper} & ExpW & MobileNet-v2 & 47.29 & 28.68 & 31.39 & 36.29 & - & 57.66 & 60.35 & 43.61 \\ 
CADA \cite{long2018conditional} & ExpW & MobileNet-v2 & 62.02 & 44.13 & 41.97 & 46.84 & - & 64.10 & 74.46 & 55.59 \\ 
SAFN \cite{xu2019larger} & ExpW & MobileNet-v2 & 57.36 & 33.33 & 35.32 & 45.94 & - & 64.00 & 71.54 & 51.25 \\ 
SWD \cite{lee2019sliced} & ExpW & MobileNet-v2 & 58.91 & 32.39 & 36.01 & 45.34 & - & 64.98 & 70.53 & 51.36 \\ 
JUMBOT \cite{fatras2021unbalanced} & ExpW & MobileNet-v2 & 71.93 & 41.53 & 32.04 & 46.54 & - & \textbf{70.13} & 73.44 & 55.94 \\
ETD \cite{li2020enhanced} & ExpW & MobileNet-v2 & 75.00 & 36.78 & 34.33 & 46.56 & - & 67.08 & 66.72 & 54.41 \\
\hline
\textbf{Ours} & ExpW & MobileNet-v2 & \textbf{85.71} & \textbf{49.30} & \textbf{44.92} & \textbf{53.22} & - & 67.85 & \textbf{76.35} & \textbf{62.89} \\ 
\hline
\end{tabular}
\vspace{2pt}
\caption{Accuracies of our proposed framework with current leading methods on the CK+, JAFFE, SFEW2.0, FER2013, ExpW, RAF-DB and AFE datasets. The results are generated by our implementation with exactly the same source dataset and backbone network.}
\label{table:fair-evaluation-results-appendix-mobilenetv2}
\end{table*}

\end{document}